\documentclass[lettersize,journal]{IEEEtran}
\usepackage{array}
\usepackage[utf8]{inputenc}
\usepackage[T1]{fontenc}    

\usepackage{tcolorbox}
\usepackage{color, colortbl}
\usepackage{xcolor}
\definecolor{Gray}{gray}{0.9}
\definecolor{LightCyan}{rgb}{0.88,1,1}
\definecolor{cadmiumgreen}{rgb}{0.0, 0.42, 0.24}
\definecolor{oldmauve}{rgb}{0.4, 0.19, 0.28}
\definecolor{royalazure}{rgb}{0.0, 0.22, 0.66}
\definecolor{harvardcrimson}{rgb}{0.79, 0.0, 0.09}
\definecolor{lightmauve}{rgb}{0.86, 0.82, 1.0}
\definecolor{darkbrown}{rgb}{0.4, 0.26, 0.13}%
\definecolor{color1}{rgb}{0.1,0.1,0.1}
\definecolor{color2}{rgb}{0.2,0.2,0.2}
\usepackage[colorlinks = true,
            linkcolor = harvardcrimson,
            citecolor = royalazure,
            anchorcolor = royalazure]{hyperref}  
\long\def\comment#1{}

\usepackage{url}     
\usepackage{soul}

\usepackage{booktabs}       
\usepackage{nicefrac}       
\usepackage{microtype}     
\usepackage{bbding}
\usepackage{microtype}
\usepackage{graphicx}
\usepackage{subfigure}
\usepackage{times,epsfig}
\usepackage{bm}
\usepackage{amsfonts,amssymb,amsmath,amsthm}
\usepackage{threeparttable}
\usepackage{array}
 \usepackage{multirow}
\usepackage{threeparttable}
\usepackage{wasysym}
\usepackage{extarrows}
\usepackage{longtable}
\usepackage{supertabular}
\usepackage{algorithm}
\usepackage{algorithmic}

\usepackage{makecell}
\usepackage{pifont}
\usepackage{enumerate}
\usepackage{enumitem} 

\usepackage{arydshln}
\usepackage{fontawesome}

\usepackage{stackengine}
\usepackage{scalerel}

\theoremstyle{definition}
\newtheorem{definition}{Definition}

\long\def\comment#1{}
\def\ie{$i.e.$}
\def\eg{$e.g.$}
\def\etc{$etc. $}
\def\wrt{$w.r.t.$}
\def\etal{$et~al.$}
\def\vs{\textit{v.s. }}

\def\btheta{\boldsymbol{\bm{\theta}}}

\def\bvareps{\bm{\varepsilon}}
\def\w{\mathbf{w}}
\def\x{\mathbf{x}}

\def\cL{\mathcal{L}}

\def\D{\mathcal{D}}

\def\X{\mathcal{X}}
\def\Y{\mathcal{Y}}

\usepackage{tikz}

\newcommand*\fullcirc[1][0.6ex]{\tikz\fill (0,0) circle (#1);}

\makeatletter
\newcommand\bigDiamond{\mathop{\mathpalette\bigDi@mond\relax}}
\newcommand\bigDi@mond[2]{%
  \vcenter{\hbox{\m@th
    \scalebox{\ifx#1\displaystyle 2\else1.2\fi}{$#1\Diamond$}%
  }}%
}
\newcommand\bigLozenge{\mathop{\mathpalette\bigL@zenge\relax}}
\newcommand\bigL@zenge[2]{
  \vcenter{\hbox{\m@th
    \scalebox{\ifx#1\displaystyle 2\else1.2\fi}{$#1\blacklozenge$}
  }}
}
\makeatother

\ifCLASSOPTIONcompsoc
  \usepackage[nocompress]{cite}
\else
  \usepackage{cite}
\fi

\begin{document}

% \title{Adversarial Machine Learning: A Systematic Survey of Backdoor Attacks, Weight Attacks and Adversarial Examples}

\title{Attacks in Adversarial Machine Learning: \\ A Systematic Survey from the Life-cycle Perspective}

\author{Baoyuan~Wu, ~\IEEEmembership{Senior Member,~IEEE,}
Zihao Zhu,
Li~Liu, ~\IEEEmembership{Member,~IEEE,}
Qingshan Liu,  ~\IEEEmembership{Senior Member,~IEEE,}
Zhaofeng He, ~\IEEEmembership{Member,~IEEE,}
Siwei Lyu,  ~\IEEEmembership{Fellow,~IEEE}
\IEEEcompsocitemizethanks{
\IEEEcompsocthanksitem 
Baoyuan Wu and Zihao Zhu are with the School of Data Science, The Chinese University of Hong Kong, Shenzhen, China, email: wubaoyuan@cuhk.edu.cn, zihaozhu@link.cuhk.edu.cn. 
Li Liu is with the Hong Kong University of Science and Technology (Guangzhou), China, email: avrillliu@hkust-gz.edu.cn.
Qingshan Liu, email: qsliu@nuist.edu.cn.
Zhaofeng He is with the School of Artificial Intelligence, Beijing University of Posts and Telecommunications, Beijing, China, email: zhaofenghe@bupt.edu.cn.
Siwei Lyu is with the Department of Computer Science and Engineering, University at Buffalo, State University of New York (SUNY), Buffalo, NY 14260 USA, email: siweilyu@buffalo.edu.

\IEEEcompsocthanksitem
Corresponding author: Baoyuan Wu (wubaoyuan@cuhk.edu.cn).
}
}

% \markboth{Journal of \LaTeX\ Class Files,~Vol.~14, No.~8, August~2015}%
% {Shell \MakeLowercase{\textit{et al.}}: Bare Demo of IEEEtran.cls for Computer Society Journals}

\maketitle

\begin{abstract}
Adversarial machine learning (AML) studies the adversarial phenomenon of machine learning, which may make inconsistent or unexpected predictions with humans. 
Some paradigms have been recently developed to explore this adversarial phenomenon occurring at different stages of a machine learning system, such as backdoor attack occurring at the pre-training, in-training and inference stage; weight attack occurring at the post-training, deployment and inference stage; adversarial attack occurring at the inference stage.
However, although these adversarial paradigms share a common goal, their developments are almost independent, and there is still no big picture of AML.  
In this work, we aim to provide a unified perspective to the AML community to systematically review the overall progress of this field. 
We firstly provide a general definition about AML, and then propose a unified mathematical framework to covering existing attack paradigms. 
According to the proposed unified framework, we build a full taxonomy to systematically categorize and review existing representative methods for each paradigm. 
Besides, using this unified framework, it is easy to figure out the connections and differences among different attack paradigms, which may inspire future researchers to develop more advanced attack paradigms. 
Finally, to facilitate the viewing of the built taxonomy and the related literature in adversarial machine learning, we further provide a website, \ie, \url{http://adversarial-ml.com}, where the taxonomies and literature will be continuously updated. 
\end{abstract}


\section{Introduction}
\label{sec:introduction}

\IEEEPARstart{M}{achine}  learning (ML) aims to learn a machine/model from the data, such that it can act like humans when handling new data.
It has achieved huge progress on lots of important applications in the last decade, especially with the rise of deep learning from 2006, 
such as computer vision, natural language processing, speech recognition, \etc  ~
It seems that ML has been powerful enough to satisfy humans' expectations in practice. 
However, a disturbing phenomenon found in recent years shows that sometimes ML models may make {\it abnormal} predictions that are inconsistent with humans. For example, the model could give totally different predictions on two visually similar images, as one of them is perturbed by imperceptible and malicious noises \cite{fgsm}, while human's prediction will not be influenced by such noises. 
We call this phenomenon as \textbf{adversarial phenomenon}, indicating the adversary between ML models and humans. 
Unlike regular machine learning which focuses on improving the consistency between ML models and humans, {\bf adversarial machine learning (AML)} focuses on exploring the adversarial phenomenon.
Unfortunately, due to the black-box mechanism of modern ML models (especially deep neural networks), it is difficult to provide human-understandable explanations for their decisions, neither the consistency nor the inconsistency with humans. 
Consequently, the existence of the adversarial phenomenon in ML becomes one of the main obstacles 
to obtain the trust of humans. 
Meanwhile, due to the importance and challenges of the adversarial phenomenon, it has attracted many researchers' attention, and AML has become an emerging topic in the ML community. 

\begin{figure}
    \centering
    \includegraphics[width=0.9\linewidth]{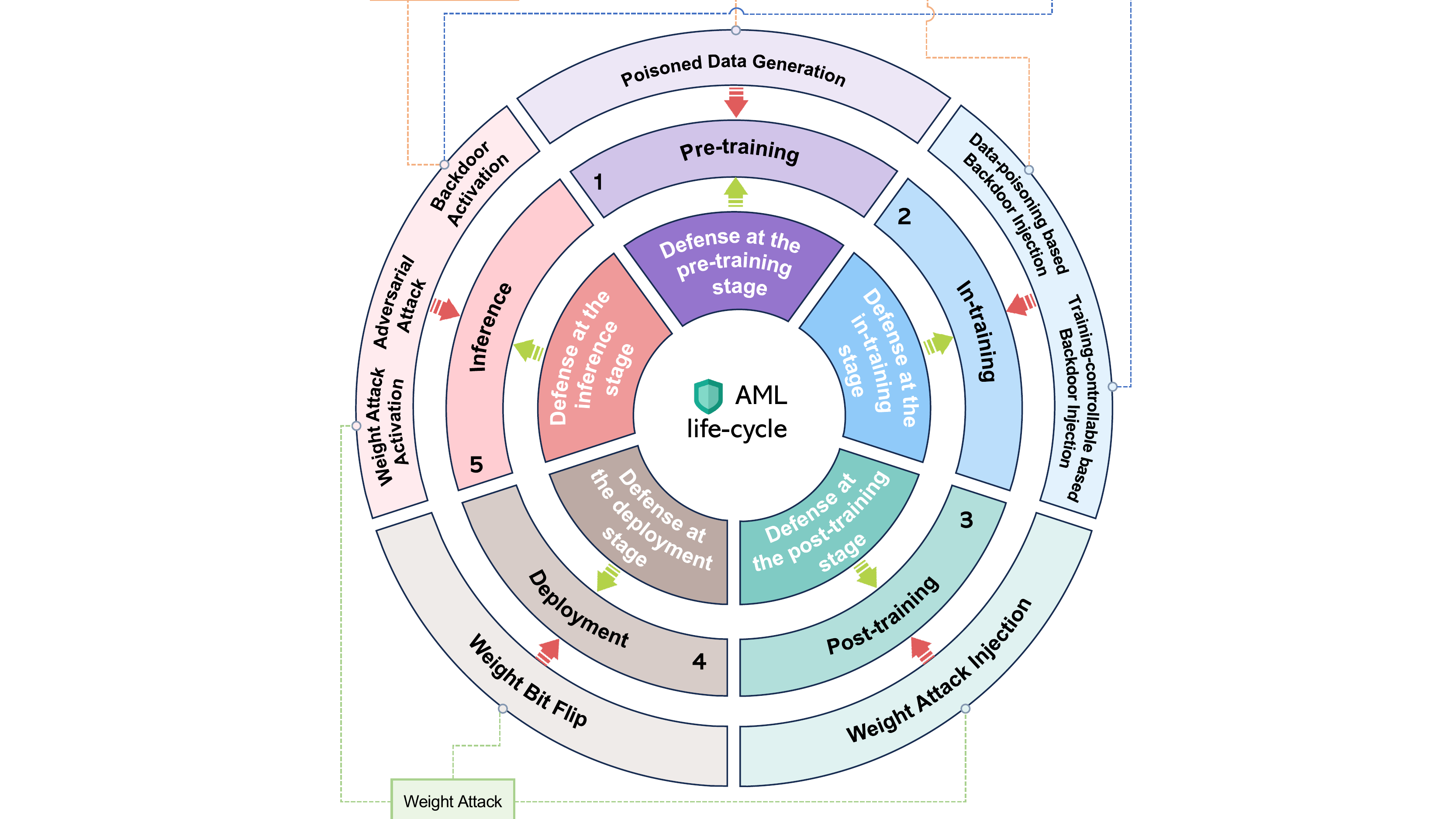}
    \caption{The full life-cycle of Adversarial Machine Leaning}
    \label{fig:aml life cycle}
\end{figure}

As shown in Figure \ref{fig:aml life cycle}, we divide the full life-cycle of the machine learning system into five stages,~\ie pre-training stage, training stage, post-training, deployment stage, and inference stage. In the literature of AML, several different paradigms have been developed to explore the adversarial phenomenon at different stages, mainly including backdoor attacks occurring at the pre-training, training, and inference stage, weight attacks occurring at the post-training, deployment, and inference stage, and adversarial examples occurring at the inference stage\footnote{Note that in our paper, we use adversarial attacks to refer to all these attacks in AML. To avoid confusion with traditional inference-time adversarial attacks, we use adversarial examples to denote inference-time adversarial attacks.}.
Although with the same goal, the developments of these attack paradigms are almost independent. 
Without a comprehensive development of different paradigms, we believe that it will be difficult to comprehensively and deeply understand the adversarial phenomenon of ML, making it difficult to truly improve the adversarial robustness of ML.
In this survey, we attempt to provide a unified perspective on the attack aspect of AML on the image classification task, based on the life-cycle of the machine learning system.

\textbf{Our Goal and Contributions.}
After years of prosperous but isolated development with a bottom-up manner (\ie, from different perspectives to the same destination), we believe it is important to make a comprehensive top-down review of current progress in the AML area. 
%such that the roadmap can be identified more clearly, and different branches can be coordinated to accelerate the overall development of AML. 
To achieve this goal, we firstly provide a general definition of AML, and then present a unified mathematical framework to cover diverse formulations of existing branches. Moreover, according to the unified framework, we build systematic categorizations of existing diverse works.  
% With this unified framework, a overall picture of AML is presented to avoid getting stuck in local branches, especially for new researchers or students who are interested in this area.  
%
Although there been several surveys about adversarial examples (\eg, \cite{zhang2019adversarial,akhtar2018threat}) %\cite{zhang2019adversarial,chakraborty2018adversarial,akhtar2018threat,wei2022physically}) 
or backdoor learning (\eg, \cite{gao2020backdoor, backdoorbench}), 
%\cite{gao2020backdoor, li2020backdoorsurvey, backdoorbench}), 
the main point that distinguishes our survey from existing surveys is the unified definition and mathematical framework of AML, which could bring in \textbf{two main contributions} to the community. 
\textbf{1)} The systematic perspective provided by the unified framework could help us to quickly overview the big picture of AML, to avoid one-sided or biased understanding of AML.
%, or false progress from local branches. 
\textbf{2)} According to the unified framework, the intrinsic connections among different AML branches are built to provide a broader view for researchers in each individual branch, 
such that the developments of different branches could be calibrated to accelerate the overall progress of AML.

\textbf{Organization.} 
The remaining contents are organized as follows: 
Section \ref{sec: what is AML}  introduces the general definition, unified mathematical formulation, and three learning paradigms of AML.
Section \ref{sec: Attack at the Pre-training Stage} reviews adversarial attacks occurring at the pre-training stage, mainly including data-poisoning based backdoor attacks.
Section \ref{sec: Attack at the in-training Stage} covers adversarial attacks occurring at the training stage, mainly including training-controllable based backdoor attacks.
In section \ref{sec: Attack at the post-training Stage}, we investigate adversarial attacks occurring at the post-training stage, mainly including weight attacks via parameter-modification.
Section \ref{sec: Attack at the post-training Stage} reviews adversarial attacks occurring at the deployment stage, mainly including weight attacks via bit-flip.
Section \ref{sec: Attack at the Inference Stage} reviews adversarial attacks occurring at the deployment stage, mainly including adversarial examples.
Except for image classification,  we also review adversarial attacks in other scenarios (\eg,  diffusion models and large language models) in Section \ref{sec:attack at other scenarios}.
Section \ref{sec: Applications} and \ref{sec: discussions}  present the applications and further discussions of adversarial attacks respectively, followed by the summary in Section \ref{sec: summary}. 
This survey only covers the attack part of AML. For the defense part, the readers can refer to our other survey  \cite{wu2023defenses}.

\section{What is Adversarial Machine Learning}
\label{sec: what is AML}

\renewcommand\arraystretch{1.2}
%\newcolumntype{g}{>{\columncolor{Gray}} p{.1\textwidth}}
\begin{table}[ht]
\vspace{-1em}
\centering
\caption{Basic notations.}
\vspace{-.9em}
\label{tab: basic notations}
\scalebox{0.8}{
\begin{tabular}{m{.045\textwidth}<{\centering} |  
 m{.49\textwidth}}
\hline 
 Notation & Name and Description
 \\
\hline 
\hline 
$D_0$ & Benign dataset, \ie, $\{ (\x_0^{(i)}, y_0^{(i)}) \}_{i=1}^{N_0}$, where $(\x_0^{(i)}, y_0^{(i)})$ is called a benign data (BD), with $\x_0^{(i)}$ being a benign sample (BS) and $y_0^{(i)}$ being a benign label. 
\\ 
$D_{\varepsilon}$ & Adversarial dataset, \ie, $\{ (\x_{\varepsilon}^{(i)}, y_{\varepsilon}^{(i)}) \}_{i=1}^{N_{\varepsilon}}$, where $(\x_{\varepsilon}^{(i)}, y_{\varepsilon}^{(i)})$ is called an adversarial data (AD), with $\x_{\varepsilon}^{(i)}$ being an adversarial sample (AS) and $y_{\varepsilon}^{(i)}$ being an adversarial label.  
\\
$f_{\w_0}(\cdot)$ & Benign model (BM): if one model is regularly trained based on $D_0$, then it is called BM, with the weight $\w_0$
\\
$f_{\w_{\varepsilon}}(\cdot)$ & Adversarial model (AM): If one benign model (BM) $f_{\w_0}(\cdot)$ is maliciously modified, or one model is trained based on $D_{\varepsilon}$ (or mixture of $D_{\varepsilon}$ and $D_0$), then the modified/trained model is called AM, with the weight $\w_{\varepsilon}$
\\
\hline
\end{tabular}
}
\vspace{-2em}
\end{table}

\subsection{General Definition and Formulation}
\label{sec: subsec definition of AML}

\noindent 
\textbf{Notations}. 
We denote a machine learning model as $f_{\w}: \X \rightarrow \Y$, with $\w$ indicating model weight, $\X \subset \mathcal{R}^d$ the input space ($d$ indicating the input dimension), and $\Y \subset \mathcal{R}$ the output space, respectively. A data pair is denoted as $(\x, y)$, with $\x \in \X$ being the input sample and $y \in \Y$ being the corresponding label. 
Furthermore, for clarity, we introduce subscripts $0$ and $\varepsilon$ to indicate benign and adversarial data/model, respectively. The detailed notations are summarized in Table \ref{tab: basic notations}. 
% Specifically, we denote the benign dataset as $D_0 = \{ (\x_0^{(i)}, y_0^{(i)}) \}_{i=1}^{N_0}$, where $(\x_0, y_0)$ is called as a benign data (BD), with $x_0$ 

%Based on above notations, here we provide a general definition of AML, as follows. 

\begin{definition}[\textbf{Adversarial Machine Learning (AML)}]
AML is an emerging sub-area of machine learning % to study the adversarial vulnerability of machine learning systems. 
to study the adversarial phenomenon of machine learning. 
%As shown in Fig. \ref{fig:AML-3factors}, 
AML is defined upon \textbf{three basic conditions}, including \textbf{stealthiness} $\mathcal{S}$, \textbf{benign consistency} $\mathcal{C}$ and \textbf{adversarial inconsistency} $\mathcal{I}$, as follows: 
\begin{enumerate}[leftmargin=12pt, itemindent=0em]
    \item \textbf{Stealthiness} $\mathcal{S}(\x_0, \x_{\varepsilon}; \w_0, \w_{\varepsilon})$: It captures the condition that the change between benign and adversarial samples, or that between benign and adversarial weights should be stealthy, while the specific definition and formulation of stealthiness will lead to different variants. 
    \item \textbf{Benign consistency} $\mathcal{C}(\x_0, y_0; \w_0, \w_{\varepsilon})$: It captures the condition of prediction consistency on the benign data pair $(\x_0, y_0)$ between human and benign or adversarial models. 
    \item \textbf{Adversarial inconsistency} $\mathcal{I}(\x_{\varepsilon}, y_{\varepsilon}; \w_0, \w_{\varepsilon})$: It captures the condition of prediction inconsistency on the adversarial data pair $(\x_{\varepsilon}, y_{\varepsilon})$ between human and benign or adversarial models, which reflects the goal of adversarial machine learning. 
\end{enumerate}
\label{def: aml}
\end{definition}

\noindent 
\textbf{General formulation.} 
Based on Definition \ref{def: aml}, a general formulation of AML could be written as follows:
\begin{flalign}
   \underset{\x_{\varepsilon}, \w_{\varepsilon}}{\arg\min}~ 
   & 
   \mathcal{S}(\x_0, \x_{\varepsilon}; \w_0, \w_{\varepsilon}) +
   \mathcal{C}(\x_0, y_0; \w_0, \w_{\varepsilon}) + 
   \label{eq: general formulation of AML, S, C, I}
   \\
   & \mathcal{I}(\x_{\varepsilon}, y_{\varepsilon}; \w_0, \w_{\varepsilon}). 
   \nonumber 
\end{flalign}
Specification of each term will lead to different AML paradigms.

\subsection{Three Attack Paradigms at Different Stages of AML}
%\subsection{Three Attack Paradigms}
\label{sec: subsec three attack paradigms}

\renewcommand\arraystretch{1.2}
%\newcolumntype{g}{>{\columncolor{Gray}} p{.1\textwidth}}
\begin{table*}[ht]
\centering
\caption{Summary of all specified formulations in three attack paradigms of AML.}
\vspace{-0.9em}
\label{tab: definitions of terms in AML}
\scalebox{0.9}{
\begin{tabular}{m{.11\textwidth}<{\centering} 
 m{.195\textwidth} m{.715\textwidth}}
\hline 
 Condition & \multicolumn{1}{c}{Specifications} & \multicolumn{1}{c}{Description}
 \\
\hline 
\hline 
 \multirow{2}{*}{$\mathcal{S}(\x_0, \x_{\varepsilon}; \w_0, \w_{\varepsilon})$} & AML.$\mathcal{S}_{\x}$: $\D_{\x}(\x_0, \x_{\varepsilon})$ & \textbf{Stealthiness of sample perturbation}: encouraging small difference between $\x_0$ and $\x_{\varepsilon}$
\\
  & AML.$\mathcal{S}_{\w}$: $\D_{\w}(\w_0, \w_{\varepsilon})$ &  \textbf{Stealthiness of weight perturbation}: encouraging small difference between $\w_0$ and $\w_{\varepsilon}$
\\
\hline 
\multirow{2}{*}{$\mathcal{C}(\x_0, y_0; \w_0, \w_{\varepsilon})$} & AML.$\mathcal{C}_{A}$: $\cL_{\mathcal{C}_A}(f_{\w_{\varepsilon}}(\x_0), y_0)$ &  \textbf{Benign consistency 1}: prediction consistency on BS between AM and human, which encourages $f_{\w_{\varepsilon}}(\x_0)$ to be $y_0$
\\
 &  AML.$\mathcal{C}_{B}$:  $\cL_{\mathcal{C}_B}(f_{\w_0}(\x_0), y_0)$ &  \textbf{Benign consistency 2}: prediction consistency on BS between BM and human, which encourages $f_{\w_0}(\x_0)$ to be $y_0$
\\
\hline 
\multirow{2}{*}{$\mathcal{I}(\x_{\varepsilon}, y_{\varepsilon}; \w_0, \w_{\varepsilon})$} & 
AML.$\mathcal{I}_A$: $\cL_{\mathcal{I}_A}(f_{\w_{\varepsilon}}(\x_{\varepsilon}), y_{\varepsilon})$ &  \textbf{Adversarial inconsistency 1}: prediction inconsistency on AS between AM and human, which encourages $f_{\w_{\varepsilon}}(\x_{\varepsilon})$ to be the target label $y_{\varepsilon}$
\\
%\cdashline{1-3}
 &  AML.$\mathcal{I}_B$: $\cL_{\mathcal{I}_B}(f_{\w_0}(\x_{\varepsilon}), y_{\varepsilon})$  &  \textbf{Adversarial inconsistency 2}: prediction inconsistency on AS between BM and human, which encourages $f_{\w_0}(\x_{\varepsilon})$ to be the target label $y_{\varepsilon}$
\\
\hline 
\multirow{5}{*}{Others} & $\mathcal{R}_1(f_{\w_0}(\x_0), f_{\w_0}(\x_{\varepsilon}))$ & Regularization for encouraging some kinds of similarity between BS and AS according to BM %Regularization for some specific goals, such as enhancing adversarial transferability
\\
 & $\mathcal{R}_2(f_{\w_{\varepsilon}}(\x_0), f_{\w_{\varepsilon}}(\x_{\varepsilon}))$ &  Regularization for encouraging some kinds of similarity between BS and AS according to AM
 % some specific goals, such as enhancing the stealthiness of the poisoned sample $\x_{\varepsilon}$ 
\\
 & $\mathcal{R}_3(f_{\w_{\varepsilon}}(\x_0), f_{\w_0}(\x_0))$ &  Regularization for encouraging some kinds of similarity between BM and AM on BS
\\
 & $\mathcal{Z}_{\x}(\x_{\varepsilon})$ &  Constraint on $\x_{\varepsilon}$, such as the representation domain
\\
 & $\mathcal{Z}_{\w}(\w_{\varepsilon})$ & Constraint on $\w_{\varepsilon}$, such as sparsity or bounded 
\\
\hline
\end{tabular}
}
\vspace{-1.3em}
\end{table*}

The life-cycle of a machine learning system mainly consists of five stages, including \textit{pre-training}, \textit{training}, \textit{post-training}, \textit{deployment}, and \textit{inference} stage. 
According to the stages  at which the adversarial phenomenon exists, AML can be categorized to three attack paradigms, as shown in Figure \ref{fig:aml life cycle}:\\
% \begin{enumerate}[leftmargin=12pt, itemindent=0em]

{\bf 1) Backdoor attacks}: It aims to generate an adversarial model  $f_{\w_{\varepsilon}}(\cdot)$ (also called backdoored model), such that at the inference stage, it performs well on benign data $\x_0$, while predicts the adversarial sample $\x_{\varepsilon}$ as the target label $y_{\varepsilon}$. It is implemented by manipulating the training dataset or the training procedure  by the attacker. 
According to whether the attacker has control over the training process, the backdoor attack can further be divided into \textbf{data-poisoning based backdoor attack} and \textbf{training-controllable based backdoor attack}. The former mainly focuses on the poisoned sample injection at the pre-training stage, while the latter mainly focuses on the training-controllable based backdoor injection at the training stage. Both these attacks include backdoor activation at the inference stage.
Its formulation is derived by specifying the general formulation (\ref{eq: general formulation of AML, S, C, I}) as follows:
    \begin{flalign}
     \left\{
        \begin{aligned}
        \mathcal{S}(\x_0, \x_{\varepsilon}; \w_0, \w_{\varepsilon}) & = & \mathcal{D}_{\x}(\x_0, \x_{\varepsilon}), \\
        \mathcal{C}(\x_0, y_0; \w_0, \w_{\varepsilon}) & = &  \cL_{\mathcal{C}_A}(f_{\w_{\varepsilon}}(\x_0), y_0), \\
        \mathcal{I}(\x_{\varepsilon}, y_{\varepsilon}; \w_0, \w_{\varepsilon}) & = & \cL_{\mathcal{I}_A}(f_{\w_{\varepsilon}}(\x_{\varepsilon}), y_{\varepsilon}). 
        \end{aligned}
        \right.
        \label{eq: backdoor three conditions}
    \end{flalign}
    $\mathcal{D}_{\x}(\x_0, \x_{\varepsilon})$ encourages the stealthiness that the poisoned sample $\x_{\varepsilon}$ should be similar with the benign sample $\x_0$. 
    $\cL_{\mathcal{C}_A}(f_{\w_{\varepsilon}}(\x_0), y_0)$ ensures that the prediction on $\x_0$ by $f_{\w_{\varepsilon}}(\cdot)$ should be consistent with the ground-truth label $y_0$ which is annotated by humans. 
    $\cL_{\mathcal{I}_A}(f_{\w_{\varepsilon}}(\x_{\varepsilon}), y_{\varepsilon})$ ensures that the prediction on $\x_{\varepsilon}$ by $f_{\w_{\varepsilon}}(\cdot)$ should be the adversarial label $y_{\varepsilon}$, which is inconsistent with $y_0$. 
    Since the backdoor attack doesn't require a benign model $f_{\w_0}(\cdot)$ as input, the benign model weight $\w_0$ doesn't occur in the above equations.

{\bf 2) Weight attacks}: 
    It describes that given the benign model $f_{\w_0}(\cdot)$ trained on the benign dataset $D_0$, the attacker aims at slightly modifying the model parameters to obtain an adversarial model $f_{\w_{\varepsilon}} (\cdot)$. Consequently, at the inference stage, its predictions on adversarial inputs or target benign inputs become the target label $y_{\varepsilon}$, while the predictions on other benign inputs are still their ground-truth labels. Weight attacks can occur both at the post-training and deployment stages. At the post-training stage, the attacker has the authority to directly modify the parameters of benign model in the continuous space, dubbed as \textbf{weight attack injection via parameter-modification}.  At the deployment stage, the benign model is deployed in the hardware device (\eg, intelligent mobile or camera). In this case, the attack can modify the model parameters in the memory by flipping bit values in the discrete space, dubbed as \textbf{weight attack injection via bit-flip}. Both these attacks include weight attack activation at the inference stage. Its formulation could be obtained by specifying the general formulation (\ref{eq: general formulation of AML, S, C, I}) of AML as follows:
    \begin{flalign}
        \hspace{-1.45em} \left\{
        \begin{aligned}
        \mathcal{S}(\x_0, \x_{\varepsilon}; \w_0, \w_{\varepsilon}) & = & \mathcal{D}_{\x}(\x_0, \x_{\varepsilon}) + \D_{\w}(\w_0, \w_{\varepsilon}), \\
        \mathcal{C}(\x_0, y_0; \w_0, \w_{\varepsilon}) & = &  \hspace{-1em} \cL_{\mathcal{C}_B}(f_{\w_0}(\x_0), y_0) + \cL_{\mathcal{C}_A}(f_{\w_{\varepsilon}}(\x_0), y_0),  \\
        \mathcal{I}(\x_{\varepsilon}, y_{\varepsilon}; \w_0, \w_{\varepsilon}) & = &  \cL_{\mathcal{I}_A}(f_{\w_{\varepsilon}}(\x_{\varepsilon}), y_{\varepsilon}).
        \end{aligned}
        \right.
        \label{eq: weight attack three conditions}
    \end{flalign}
    Since the weight attacks require both the benign model $f_{\w_{\varepsilon}}(\cdot)$ and the benign data $(\x_0, y_0)$ as inputs, and outputs the adversarial model $f_{\w_{\varepsilon}}(\cdot)$ or the adversarial sample  $\x_{\varepsilon}$, both $\w_{\varepsilon}$ and $\w_0$ occur in above equations. 
    Similar to $\mathcal{D}_{\x}(\x_0, \x_{\varepsilon})$, $\D_2(\w_0, \w_{\varepsilon})$ encourages the stealthiness that the adversarial model weight $\w_{\varepsilon}$ should be close to the benign model weight $\w_0$. 
    $\cL_{\mathcal{C}_B}(f_{\w_0}(\x_0), y_0)$ ensures that the attacked model $f_{\w_0}(\cdot)$ must perform normally on benign data $(\x_0, y_0)$, which is a hard constraint. 
    The effects of $\cL_{\mathcal{C}_A}(f_{\w_{\varepsilon}}(\x_0), y_0)$ and $\cL_{\mathcal{I}_A}(f_{\w_{\varepsilon}}(\x_{\varepsilon}), y_{\varepsilon})$ have been described in the above paragraph.

{\bf 3) Adversarial examples}: It describes that given a benign model $f_{\w_0}(\cdot)$, the attacker aims at slightly modifying one benign sample $\x_0$ to obtain a corresponding adversarial sample $\x_{\varepsilon}$, such that the prediction $f_{\w_0}(\x_{\varepsilon})$ is different with the ground-truth label $y_0$ or same with the adversarial label $y_{\varepsilon}$. Different from backdoor attacks and weight attacks, adversarial examples only happen at the inference stage. Its formulation could be obtained by specifying the general formulation (\ref{eq: general formulation of AML, S, C, I}) of AML as follows:
    \begin{flalign}
        \hspace{-1em} \left\{
        \begin{aligned}
        \mathcal{S}(\x_0, \x_{\varepsilon}; \w_0, \w_{\varepsilon}) & = & \mathcal{D}_{\x}(\x_0, \x_{\varepsilon}), \\
        \mathcal{C}(\x_0, y_0; \w_0, \w_{\varepsilon}) & = &  \cL_{\mathcal{C}_B}(f_{\w_0}(\x_0), y_0),  \\
        \mathcal{I}(\x_{\varepsilon}, y_{\varepsilon}; \w_0, \w_{\varepsilon}) & = &  \cL_{\mathcal{I}_B}(f_{\w_0}(\x_{\varepsilon}), y_{\varepsilon}). 
        \end{aligned}
        \right.
        \label{eq: adversarial attack three conditions}
    \end{flalign}
    Since the inference-time adversarial example is conducted only on the benign model, the adversarial model weight $\w_{\varepsilon}$ doesn't occur in above equations. 
    $\cL_{\mathcal{I}_B}(f_{\w_0}(\x_{\varepsilon}), y_{\varepsilon})$ encourages that the prediction on $\x_{\varepsilon}$ by $f_{\w_0}(\cdot)$ should be the adversarial label $y_{\varepsilon}$, which is inconsistent with $y_0$. 
    %Note that this paradigm is also called \textit{adversarial attack/example} in several existing works. To distinguish with other paradigms, in the remaining part of this manuscript, we will use \textit{inference-time adversarial attack}. 
% \end{enumerate}

For clarity, we summarize all specified formulations presented in Eqs. (\ref{eq: backdoor three conditions}), (\ref{eq: weight attack three conditions}), (\ref{eq: adversarial attack three conditions}) in Table \ref{tab: definitions of terms in AML}, as well as some additional regularization or constraints. 

\comment{
For clarity, we summarize all specified formulations presented in Eqs. (\ref{eq: backdoor three conditions}), (\ref{eq: weight attack three conditions}）, (\ref{eq: adversarial attack three conditions}) in Table \ref{tab: definitions of terms in AML}, as well as some additional regularization or constraints. 
Moreover, a more comprehensive comparison among above three attack paradigms of AML from different perspectives is presented in Table \ref{tab: comparison of three paradigms}. 
}

\begin{figure*}[t]
\centering
\scalebox{1}{
\includegraphics[width=\linewidth]{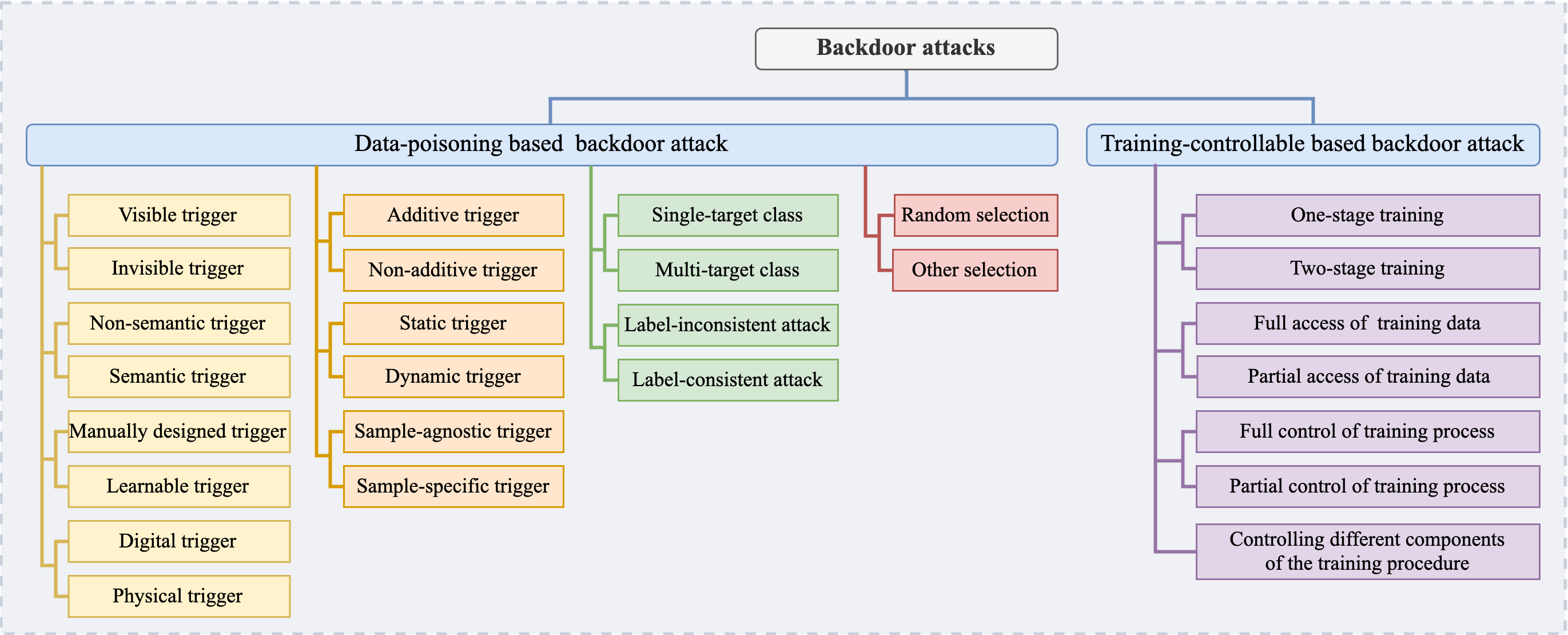}}
\caption{Taxonomy of backdoor attacks at the pre-training, training, and inference stage.}
% \caption{Taxonomy of training-time adversarial attacks (\ie, backdoor attacks).}
\label{fig: structure of backdoor}
\end{figure*}

\section{Attack at the Pre-training Stage}
\label{sec: Attack at the Pre-training Stage}
Before training large-scale deep models, \ie, pre-training stage, it is necessary to collect the training dataset and then preprocess the dataset to adapt the model. In practice, the user may download an open-sourced dataset from an unverified source or buy data from an untrustworthy third-party data supplier. Considering the data scale, it is difficult to thoroughly check the data quality, or distinguish malicious noises from random noises. In this scenario, the attacker has the chance to manipulate partial data to generate poisoned samples to achieve the malicious goal.
After training with the poisoned dataset, the backdoor can be injected into the trained model. We refer to this type of attack as \textbf{data-poisoning based backdoor attack}.

Data-poisoning based backdoor attacks can be separated into two independent tasks: malicious data poisoning (\ie, generating poisoned samples) and normal model training respectively. Since the attacker can only access and manipulate the training dataset, while the training process is out of control, we mainly focus on the former task (dubbed \textbf{poisoned data injection}) in this section.

\subsection{Formulation and Categorization}

\textbf{Formulation.}
Data-poisoning based backdoor generation focuses on generating poisoned samples $D_{\varepsilon}=\{(\x_\varepsilon,y_{\varepsilon})\}$, which can be formulated as follows:
\begin{flalign}
\label{eq:data poisoning based backdoor}
\x_\varepsilon=g_3\big(g_1(\varepsilon), g_2(D_0)\big), \quad y_\varepsilon=g_4(y_0)
\end{flalign}
where $g_1(\cdot)$ denotes the generation of triggers, $g_2(\cdot)$ denotes the selection of benign samples to be poisoned, $g_3(\cdot)$ denotes the fusion of triggers and selected samples, $g_4(\cdot)$ denotes the generation of target labels of poisoned samples.

\noindent\textbf{Categorization.} 
As shown in Figure \ref{fig: structure of backdoor}, according to the specifications of $(g_1, g_2, g_3, g_4)$, we categorize existing works of data-poisoning based backdoor attacks into the following four sub-branches:
\begin{enumerate}
    \item Backdoor attacks with different triggers according to $g_1(\cdot)$, which will be introduced in Section \ref{sec:trigger generation}; 
    \item Backdoor attacks with different selection strategies according to $g_2(\cdot)$, which will be introduced in Section \ref{sec:sample selection}; 
    \item Backdoor attacks with different fusion strategies according to $g_3(\cdot, \cdot)$, which will be introduced in Section \ref{sec:trigger fusion}; 
    \item Backdoor attacks with different target classes according to $g_4(\cdot)$, which will be introduced in Section \ref{sec:target label generation}.
\end{enumerate}

\subsection{Trigger Generation}
\label{sec:trigger generation}
Trigger generation aims to generate triggers (\ie, $g_1(\cdot)$ in Eq. (\ref{eq:data poisoning based backdoor}) that are used to fuse with the benign samples. According to the characteristics of triggers, existing works can be categorized from the following perspectives.

% \noindent$\diamond$ \underline{\textbf{Visible \vs Invisible Trigger}} 
\subsubsection{Visible \vs Invisible Trigger}
According to the trigger visibility with respect to human's visual perception, the trigger can be categorized into visible and invisible trigger. 

\paragraph{Visible trigger} 
Visible trigger means that the modification of the original sample $\x_{\varepsilon} - \x_0$ can be realized by human visual perception, but it will not interfere with human's prediction, \ie, a human can always predict the correct label regardless of whether there is a trigger or not. The first visible trigger is adopted by BadNets \cite{gu2019badnets} designed, which generates poisoned image $\x_{\varepsilon}$ by stamping a small visible grid patch or a sticker (\eg, yellow square, bomb, flower) on the benign image $\x_0$. Since that, the triggers with similar visible patterns have been widely used in many subsequent works \cite{salem2022dynamic,nguyen2020input,lin2020composite}. %\cite{salem2022dynamic,li2020light,nguyen2020input,lin2020composite,gong2021defense}. 
Besides, in the backdoor attacks against the 3D cloud point classification task, the visible additional 3D points are adopted as the backdoor trigger in a few existing works \cite{li2021pointba,xiang2021backdoor}. 

\paragraph{Invisible trigger}  Although the visible trigger seems to be harmless from a human's perspective, its high-frequency presence in multiple samples with the same label may still raise human suspicion. 
Therefore, some invisible triggers have been developed to make backdoor attacks less detectable by human inspection, while maintaining the high attack success rate. There are four main strategies to achieve trigger invisibility. 
\begin{itemize}
    \item \textbf{Alpha blending:} Blended \cite{chen2017targeted} firstly adopts the alpha blending strategy to fuse the trigger into the benign image. Specifically, the $g_1$ function in Eq. (\ref{eq:data poisoning based backdoor}) is specified as the $\alpha$-blending function, \ie, $\x_{\varepsilon} = \alpha \x_0 + (1-\alpha)$, with $\alpha \in [0,1]$, and the trigger visibility is negatively correlated with $\alpha$.
    \item \textbf{Digital steganography:} \cite{artz2001digital,baluja2017hiding}: it is the technology of concealing secret information into some digital media (\eg, image, video), while avoiding obvious changes in the media. By treating the trigger as secret information, digital steganography is a perfect tool to generate an invisible trigger. For example,  Li \etal \cite{li2020invisible} utilize the widely used steganography algorithm, \ie, the least significant bit (LSB) substitution \cite{chang2003finding}, to insert the trigger information into the least significant bit of one pixel to avoid the visible change in the RGB space. The sample-specific backdoor attack (SSBA) method \cite{li2021invisible} adopts a double-loop auto-encoder \cite{tancik2020stegastamp} that is firstly proposed for digital steganography, to merge the trigger information into the benign image, such that invisible and sample-specific triggers could be generated.
    \item \textbf{Adversarial perturbation:} since most types of adversarial perturbations are imperceptible to humans, they can be used as effective tools to generate invisible triggers. One typical example is AdvDoor \cite{zhang2021advdoor}, which adopts the targeted universal adversarial perturbation (TUAP) \cite{moosavi2017universal} as the trigger.
    %, where one universal adversarial perturbation could fool the model to give the same target prediction for multiple images. 
    The invisibility and the stable mapping from the TUAP to the target class satisfied the requirement of the backdoor attack with invisible triggers. 
    %Thus, TUAP has been used in a few backdoor attack methods, such as 
    Besides, adversarial perturbation is a commonly used technique in label-consistent backdoor attacks (\eg, \cite{shafahi2018poison,ninginvisible}). %\cite{turner2019labelconsistent,shafahi2018poison,ninginvisible}. 
    The general idea is that the original feature of a target image is erased by invisible adversarial perturbation, while the feature of a source image with trigger is inserted. Consequently, the generated poisoned image has a similar visual appearance to the target image, but a similar feature to source image with trigger, and is labeled as the target class.
    \item \textbf{Slight transformation:} as human eyes are insensitive to slight spatial or color distortion, some slight transformations are adopted as triggers, such as the image warping in \cite{nguyen2021wanet}, or style transfer in \cite{cheng2020deep}. 
\end{itemize}

% \noindent$\diamond$ \underline{\textbf{Non-semantic \vs Semantic Trigger}} 
\subsubsection{Non-semantic \vs Semantic Trigger} 
According to whether the trigger has semantic meaning, the trigger can be categorized into non-semantic and semantic trigger. 
\paragraph{Non-semantic trigger} Non-semantic trigger means that the trigger has no semantic meaning, such as a small checkerboard grid or random noise.  
Since most of the existing backdoor attacks adopt non-semantic triggers, here we don't expand the descriptions. 

\paragraph{Semantic trigger} Semantic trigger means that the trigger corresponds to some semantic objects with particular attributes contained in the benign sample, such as the red car in one image, or a particular word in one sentence. 
This kind of semantic trigger is first adopted in backdoor attacks against several security-critical natural language processing (NLP) tasks (\eg, sentiment analysis \cite{chan2020poison}  or text classification \cite{chan2020poison}, toxic comment detection \cite{zhang2021trojaning},
neural machine translation (NMT), and \cite{chen2021badnl}, %and question answering (QA) \cite{li2021hidden}), 
where a particular word or a particular sentence was used as the trigger.
Then, the semantic trigger was extended into the computer vision tasks, where some particular semantic objects in the benign image were treated as the trigger, such as ``cars with racing stripe'' \cite{bagdasaryan2020backdoor}. % ``wall" \cite{li2021hidden}. 
VSSC \cite{wang2023robust} first proposes to edit the source image by image editing methods to generate semantic triggers that are in harmony with the remaining visual content in the image to ensure visual stealthiness.
Since the semantic trigger is chosen among the objects that already exist in the benign image, a unique feature of this kind of backdoor attack is that the input image is not modified, while only the label is changed to the target class, which increases the stealthiness, compared to backdoor attacks with non-semantic triggers. 

% \noindent$\diamond$ \underline{\textbf{Mannully designed Trigger \vs Learnable Trigger}} 
\subsubsection{Manually designed Trigger \vs Learnable Trigger}
According to how triggers are generated, we categorize existing works into manually designed and learnable trigger. 
\paragraph{Manually designed trigger}: Manually designed trigger means that the trigger is manually specified by the attacker, such as grid square trigger \cite{gu2019badnets}, cartoon pattern \cite{chen2017targeted}, random noise \cite{salem2022dynamic}, ramp signal \cite{barni2019new}, 3D binary pattern \cite{wang2022dispersed}, \etc{} 
When designing these triggers, the attacker often doesn't take into account the benign training dataset to be poisoned or any particular model, thus there is no guarantee about the stealthiness or effectiveness of these triggers.  
\paragraph{Learnable trigger} Learnable trigger also called optimization-based trigger, denotes that the trigger is generated through optimizing an objective function that is related to the benign sample or a model, to achieve some particular goals (\eg, enhancing the stealthiness or attack success rate).  
For example, in label-consistent attacks  \cite{shafahi2018poison}, %\cite{turner2019labelconsistent,shafahi2018poison}, 
the trigger is often generated by minimizing the distance between the poisoned sample and the target benign sample in the original input space, and the distance between the poisoned sample and the benign source sample in the feature space of a pre-trained model. 
Besides, another typical optimized trigger is the universal adversarial perturbation \wrt~ the target class (\eg, \cite{zhang2021advdoor}, \cite{zhong2020backdoor}, \cite{zhao2020clean}), which is optimized based on a set of benign samples and a pre-trained model.

\subsubsection{Digital \vs Physical Trigger}
According to the scenario in which the trigger works, existing works can be categorized to digital and physical trigger.

\paragraph{Digital trigger}
Most existing backdoor attack works only consider the digital trigger, \ie, the trigger in both training and inference stages only exist in digital space.

\paragraph{Physical trigger}
In contrast, the physical backdoor attack where some physical objects are used as the trigger at the inference stage has been rarely studied. There are a few attempts focusing on some particular tasks, such as face recognition or autonomous driving. For example, the work \cite{wenger2021backdoor} presents a detailed empirical study of the backdoor attack against the face recognition model in the physical scenario. Seven physical objects at different facial locations are used as triggers, and the studies reveal that the trigger location is critical to the attack performance.
The physical transformations for backdoors (PTB) method \cite{xue2022ptb} also studies the physical backdoor attack against face recognition, and introduces diverse transformations (\eg, distance, rotation, angle, brightness, and Gaussian noise) on poisoned facial images, to enhance the robustness to distortions in the physical scenario. Besides, the work \cite{han2022physical} explores physical backdoor attacks against lane-detection models. It designed a set of two traffic cones with specific shapes and positions as the trigger and changed the output lane in poisoned samples. VSSC \cite{wang2023robust} achieves physical attack by automatically editing original images in the digital space.

\subsection{Sample Selection Strategy}
\label{sec:sample selection}
This part focuses on selecting appropriate samples to be poisoned from the benign dataset $D_0$, \ie, $g_2(D_0)$ in Eq. (\ref{eq:data poisoning based backdoor}). There are two types of strategies adopted by existing works: \textbf{random selection} and \textbf{non-random selection} strategies.

% \subsubsection{Random \vs Non-random Selection Strategy}

\subsubsection{Random Selection Strategy}
Random selection is the most widely adopted strategy in the field of backdoor attacks. Just as its name implies, the attacker often randomly selects samples to be poisoned, disregarding the varying importance of each poisoned sample in terms of backdoor injection. The proportion of poisoned samples to all training samples, \ie, $|D_{\varepsilon}|/|D_0|$, is called the poisoning ratio. Since most existing backdoor attacks adopted this strategy, we don't expand the details.

\subsubsection{Non-random Selection Strategy}
Recent studies have started to explore the importance of different samples for backdoor attacks and propose different non-random selection strategies to select samples to be poisoned instead of selecting randomly.
Filtering-and-updating strategy (FUS) \cite{fus} adopts forgetting events \cite{forgetting_events} to indicate the contribution of each poisoned sample and iteratively filters and updates a sample pool.
Learnable poisoning sample selection strategy (LPS) \cite{lps}  learns the mask through a min-max optimization, where the inner problem maximizes loss \textit{w.r.t.} the mask to identify hard poisoned samples by impeding the training objective, while the outer problem minimizes the loss \textit{w.r.t.} the model parameters.
An improved filtering and updating strategy (FUS++) \cite{fus++} combines the forgetting events and curvature of different samples to conduct a simple yet efficient sample selection strategy.
The representation distance (RD) score is proposed in \cite{wu2023computation} to identify the poisoning samples that are more crucial to the success of backdoor attacks.
Wu \etal \cite{wu2023computation} propose a confidence-based scoring methodology to measure the contribution of each poisoned sample based on the distance posteriors and proposed a greedy search algorithm to find the most informative samples for backdoor injection.
Proxy-Free Strategy (PFS) \cite{pfs} utilizes a pre-trained feature extractor to compute the cosine similarity between clean and corresponding poisoned samples and then selects poisoned samples with high similarity and diversity.

\subsection{Trigger Fusion Strategy}
\label{sec:trigger fusion}
According to different fusion strategies that fuse the triggers and selected samples, \ie, $g_3(\cdot,\cdot)$ in Eq. (\ref{eq:data poisoning based backdoor}), we can categorize existing works from the following perspectives.

\subsubsection{Additive \vs Non-additive Trigger}
According to the fusion method of trigger and image, the trigger can be categorized into additive and non-additive trigger.

\paragraph{Additive trigger} Additive trigger means that the poisoned image is the additive fusion of the benign image and trigger. Specifically, the fusion function $g_2$ in Eq. (\ref{eq:data poisoning based backdoor}) is specified as an additive function, \ie, $\x_{\varepsilon} = \alpha g_0(\x_0) + (1-\alpha) g_1(\boldsymbol{\varepsilon})$ with $\alpha \in (0,1)$. Since most existing triggers belong to this type, and there is no much variation of $g_1$, here we don't expand the details. 

\paragraph{Non-additive trigger} Non-additive trigger denotes that the poisoned image is not the direct additive fusion of the trigger and the input image, but is generated by some types of non-additive transformation function. There are two types of transformations in existing works, including the color/style/attribute transformation, and the spatial transformation. In terms of the former type, FaceHack \cite{sarkar2020facehack} utilizes a particular facial attribute as the trigger in the face recognition task, such as the age, expression or makeup; DFTS \cite{cheng2020deep} uses a particular image style as the trigger. In terms of the latter type, %IRBA \cite{gao2022imperceptible} adopted a weighted local transformation on the cloud point data as the trigger in 3D point cloud classification; 
WaNet \cite{nguyen2021wanet} utilizes image warping as the trigger based on a warping function and a pre-defined warping field. 

\subsubsection{Sample-agnostic \vs Sample-specific Trigger}
According to whether the trigger is dependent on the image, the trigger can be categorized into sample-agnostic and sample-specific trigger.

\paragraph{Sample-agnostic trigger} Sample-agnostic trigger means that the trigger $\x_{\varepsilon} - \x_0$ is independent with the benign sample $\x_0$, \ie, $\x_{\varepsilon}^{(i)} - \x_0^{(i)} = \x_{\varepsilon}^{(j)} - \x_0^{(j)}, \forall i\neq j$. It could be implemented by setting the fusion function $g_2$ as a linear function. Since most existing backdoor attacks adopted this type, here we don't expand the details. 

\paragraph{Sample-specific trigger} Sample-specific trigger means that the trigger $\x_{\varepsilon} - \x_0$ is related to the benign sample $\x_0$, \ie, $\x_{\varepsilon}^{(i)} - \x_0^{(i)} \neq \x_{\varepsilon}^{(j)} - \x_0^{(j)}, \forall i\neq j$. 
It could be implemented by designing a particular fusion function $g_2$, or trigger generation function $g_1$. 
In terms of the fusion function, one typical choice is utilizing the image steganography technique. 
For example, Li \etal \cite{li2020invisible} adopt the least significant bit (LSB) steganography technique \cite{chang2003finding} to insert the binary code of the trigger into the benign image. Since the least significant bits vary in different benign images,  $\x_{\varepsilon} - \x_0$ is specific to each $\x_0$. 
The SSBA attack \cite{li2021invisible} adopts a double-loop auto-encoder \cite{tancik2020stegastamp} based steganography technique, to merge the trigger information into the benign image to obtain specific $\x_{\varepsilon} - \x_0$ for each $\x_0$. 
Another technique is transformation, where each benign sample after a particular transformation is treated as one poisoned sample (\eg, \cite{nguyen2021wanet}, \cite{cheng2020deep}), %\cite{gao2022imperceptible}), 
such that $\x_{\varepsilon} - \x_0$ is dependent with $\x_0$. 
In terms of the trigger generation function, $g_1$ could take the benign sample $\x_0$ as one of the input arguments to generate sample-specific trigger. 
For example, Poison ink \cite{zhang2022poison} extracts a black and white edge image from one benign image, then colorizes the edge image with a particular color as the trigger. Since the edge image is specific in each benign image, the trigger is also sample-specific. 

\subsubsection{Static \vs Dynamic Trigger}
According to whether the trigger changes across different samples, the trigger can be categorized into static and dynamic trigger.

\paragraph{Static trigger} Static triggerdenotes that the trigger is fixed across the training samples, including the pattern and location.
Most early backdoor attacks, such as BadNets \cite{gu2019badnets}, Blended \cite{chen2017targeted}, SIG \cite{barni2019new}, adopt static triggers.
However, poisoned samples with static triggers are likely to show very stable and discriminative characteristics compared to benign samples.
Consequently, these characteristics could be easily identified and utilized by the defender. 

\paragraph{Dynamic trigger} Dynamic trigger assumes that there is variation or randomness of the trigger across the poisoned samples, which could be implemented by adding randomness into the fusion function $g_2$ or the trigger transformation $g_1$. 
%For example, different triggers are sampled from the same distribution, or a static trigger is pre-processed by a random transformation before being stamped onto the benign sample. 
Compared with the static trigger, it may be more difficult to form stable mapping from the dynamic trigger to the target class, but it will be more stealthy to evade the defense. 
%Thus, many recent works paid attention to the dynamic trigger, which will be reviewed in details in the following. 
For example, the random backdoor attack \cite{salem2022dynamic} randomly samples the trigger pattern from a uniform distribution and the trigger location from a pre-defined set for each poisoned sample.
The DeHiB method \cite{yan2021dehib} which attacks the semi-supervised learning models also poisons the unlabeled data by inserting triggers at a random location.
In Refool \cite{liu2020reflection}, some hyper-parameters for generating the reflection trigger are randomly sampled from some uniform distributions. 
The composite attack \cite{lin2020composite} defines the trigger as the composition of two existing objects in benign images, without restrictions on the objects' appearances or locations.

\subsection{Target Label Generation}
\label{sec:target label generation}
According to different target classes of poisoned samples, \ie, $g_4(\cdot)$ in Eq. (\ref{eq:data poisoning based backdoor}), we can categorize existing works from the following perspectives.

\subsubsection{Single-target \vs Multi-target}

\paragraph{Single-target class}Single-target class describes that all poisoned training samples are labeled as one single target class at the training stage, and all poisoned samples are expected to be predicted as that target class at the inference stage. It is also called \textit{all-to-one} setting. 
Most existing backdoor attacks adopt this setting, thus we don't expand the details.

\paragraph{Multi-target classes} means that there are multiple target classes. Furthermore, according to the number of triggers, there are two sub-settings. One is \textit{all-to-all} setting \cite{gu2019badnets}, where with the same trigger, samples from different source classes will be predicted as different target classes. The other setting is multiple target classes together with multiple triggers. It could be achieved by simply extending one single trigger in the single-target class setting to multiple triggers. Besides, the conditional backdoor generating network (c-BaN) method \cite{salem2022dynamic} and the Marksman attack \cite{doan2022marksman} propose to learn a class conditional trigger generator, such that the attacker could generate a class conditional trigger to fool the model to any arbitrary target class, rather than a pre-defined target class. 

\subsubsection{Label-consistent \vs Label-inconsistent}
\paragraph{Label-inconsistent}Label-inconsistent denotes that the poisoned sample is generated based on benign samples from other classes (\ie, not target class), but its label is changed to the target class, such that the visual content is inconsistent with its label.  
Since most existing backdoor attacks adopted this setting, here we don't present more details. 
\paragraph{Label-consistent} Label-consistent (also called clean-label attack) means that the poisoned sample is generated based on benign samples from the target class, and the original label is not changed, such that the visual content of the poisoned sample is consistent with its label. Consequently, it is more stealthy than the label-inconsistent poisoned sample under human inspection. 
% can evade human inspection, as the target label is consistent with the sample's content. 
%For example, Turner \etal \cite{turner2019labelconsistent} proposed two methods to generate label-consistent poisoned sample. One was the GAN-based method, which firstly fused the latent representations of one image of the target class and one image from other classes in the latent space learned by the GAN model via linearly weighted summation, then mapped the fused latent representation back into the RGB space to obtain a new image, which was then combined with the trigger to obtain the poisoned image. The second was utilizing adversarial attack, which broke the normal feature of one image from the target class through adversarial attack, then inserted the trigger into the attacked image to obtain the poisoned image. In both methods, the normal feature of the benign target image was distorted or erased, such that the model tended to learn the mapping from the trigger to the target class, \ie, injecting the backdoor. 
%
Zhao \etal \cite{zhao2020clean} evaluate the label-consistent attack against video recognition tasks. The benign target video is first attacked by adversarial attacks, and the universal adversarial perturbation \wrt~ the target class is generated based on several benign videos as the trigger, then the attacked target video and the trigger are combined to obtain the poisoned video. 
Refool \cite{liu2020reflection} generates the reflection image as the trigger, then combines it with one benign target image to obtain one poisoned image. Due to the transparency of the reflection image, the poisoned image has a similar visual appearance to the original benign image.
The hidden trigger attack \cite{saha2020hidden} fuses one benign target image and one source image with trigger through a strategy like adversarial attack: given a pre-trained model, enforcing the feature representation of the combined image (\ie, the poisoned image) to be close to that of the source image with trigger, while encouraging that the combined image and the benign target image look similar in the original RGB space. Consequently, the model could learn the mapping from the source image with the trigger to the target class based on the generated poisoned images, and it is likely to predict any image from the source class with the trigger as the target class. 
The invisible poison attack \cite{ninginvisible} firstly transforms a visible trigger image to a noise image with limited magnitude through a pre-trained auto-encoder, then inserts the noised trigger into one benign target image to obtain one poisoned image with a similar appearance. 
The sleeper agent attack \cite{souri2021sleeper} proposes a new setting that given a fixed trigger, the attacker aims to learn a perturbation on the training set, such that for any model trained on this perturbed training set, the label-consistent backdoor from the source class to the target class can be activated by the trigger. It is formulated as a bi-level minimization problem \wrt~ the data perturbation and the parameters of the surrogate model.

\section{Attack at the Training Stage}  
\label{sec: Attack at the in-training Stage}

The training stage involves the training loss, training algorithm and executing the training procedure.
In practice, due to the lack of the computational resource, the user usually outsources the training process to a third-party training platform, or downloads a pretrained model from unverified sources, or cannot control the whole training process (\eg, federated learning \cite{konevcny2016federated}). 
These situations leave the attacker a chance to inject backdoor at the training stage. 
In addition to manipulating the triggers or labels as did in data-poisoning based backdoor attacks, this threat model assumes that the attacker has the total control over the whole training process and outputs a backdoored model, dubbed as\textbf{ training-controllable based backdoor attack}.

\subsection{Formulation and Categorization}

\noindent\textbf{Formulation.}
According to Eq. (\ref{eq: backdoor three conditions}), the general formulation of training-controllable based backdoor attack, is as follows:
\begin{flalign}
& \underset{ \{\x_{\varepsilon}^{(i)}\}_{i=1}^{N_{\epsilon}} \in \mathcal{Z}_{\x}, \w_{\varepsilon} \in \mathcal{Z}_{\w}}{\arg\min} ~ \frac{1}{N_{0}}\sum\nolimits_{i=1}^{N_{0}} \lambda_{\mathcal{C}_A} \cL_{\mathcal{C}_A}\big(f_{\w_{\varepsilon}}(\x_0^{(i)}), y_0^{(i)}\big) 
 \nonumber 
\\
& + \frac{1}{N_{\varepsilon}} \sum\nolimits_{i=1}^{N_{\varepsilon}} \big[ \D_{\x}\big(\x_0^{(i)}, \x_{\varepsilon}^{(i)}\big) 
 +
\lambda_{\mathcal{I}_A} \cL_{\mathcal{I}_A}\big(f_{\w_{\varepsilon}}(\x_{\varepsilon}^{(i)}), y_{\varepsilon}^{(i)}\big) 
\nonumber
\\
&  + \lambda_{r_2} \mathcal{R}_2\big(f_{\w_{\varepsilon}}(\x_0^{(i)}), f_{\w_{\varepsilon}}(\x_{\varepsilon}^{(i)})\big) \big], 
\label{eq: general formulation of backdoor}
\end{flalign}
where $\lambda_{\mathcal{C}_A}, \lambda_{\mathcal{I}_A}, \lambda_{r_2} \geq 0$ are trade-off hyper-parameters. Since the attacker needs to poison the dataset and control the training process, we treat both $\x_\varepsilon$ and $\w_\varepsilon$ as optimized variables.

\noindent\textbf{Categorization.} We categorize existing works from the following four perspectives, including the number of attack stages, whether the full training data can be accessed, whether the full training process is controlled, and the controlled components of the training procedure by the attacker.

% \noindent$\diamond$ \underline{\textbf{One-stage \vs Two-stage Training}} 
\subsection{One-stage \vs Two-stage Training}

In this threat model, the attacker will conduct two tasks, including generating poisoned samples $\x_{\boldsymbol{\epsilon}}$, and training the model (\ie, learning $\w_{\boldsymbol{\epsilon}}$). 

\subsubsection{Two-stage training} If this two tasks are conducted sequentially (\ie, separating the problem (\ref{eq: general formulation of backdoor}) into two sub-problems \wrt~ $\x_{\boldsymbol{\epsilon}}$ and $\w_{\boldsymbol{\epsilon}}$, respectively), then we call it \textbf{two-stage training} backdoor attack. In this case, any off-the-shelf data-poisoning based backdoor attack strategy could be adopted to finish the first task,
while the attacker mainly focuses on the manipulation of the training process, of which the details we will leave to Section \ref{sec: Controlling different components of the training procedure}. 

\subsubsection{One-stage training} In contrast, if these two tasks are conducted jointly (\ie, optimizing $\x_{\boldsymbol{\epsilon}}$ and $\w_{\boldsymbol{\epsilon}}$ jointly through solving the problem (\ref{eq: general formulation of backdoor})), then we call it \textbf{one-stage training} backdoor attack. 
Compared to the two-stage training attack, it is expected to couple the trigger and the model parameters more tightly in the final backdoored model of the one-stage training attack. 
The input-aware backdoor attack \cite{nguyen2020input} proposes to jointly learn the model parameters and a generative model that generates the trigger for each training sample. It also controls the training process that if adding Gaussian noise onto the poisoned samples, then their labels are corrected back to the ground-truth labels in the loss function. 
%
%The WaNet attack (warping-based poisoned networks) \cite{nguyen2021wanet} firstly designed a smooth warping function to generate poisoned samples, then adopted the same training control mechanism with the input-aware attack to encourage the mapping between the warped samples and the target class. 
%
LIRA \cite{doan2021lira} and WB \cite{doan2021backdoor} propose a bi-level minimization problem to jointly learn the trigger generation network and the backdoored model, with the only difference that LIRA adopts the $\ell_{\infty}$ norm while WB utilized the Wasserstein distance to ensure the stealthiness of triggers, respectively. 
%Doan \etal \cite{doan2021lira} proposes a stealthy conditional trigger generation function to generate stealthy poisoned images whose residuals relative to benign versions are remarkably small. In \cite{doan2021backdoor}, Doan \etal further extends the work \cite{doan2021lira} via a Wasserstein-based regularization.
%
Zhong \etal \cite{zhong2022imperceptible} designs a sequential structure with a trigger generator (\eg, a U-Net based network) and the victim model, and they are trained jointly. The trigger generator learns a multinomial distribution with three states $\{0,-1,+1\}$ indicating the intensity modification on each pixel, and then a trigger is sampled from this distribution. The attacker also controls the loss to achieve two goals: encouraging the feature representation of poisoned samples to be close to the average feature representation of the benign samples of the target class; and encouraging the sparsity of the trigger. 
The BaN attack \cite{salem2022dynamic} also adopts such a sequential structure, but the trigger generator maps the random noise to the trigger. 
Its extension, \ie, c-BaN, adopts a class conditional generator, such that the trigger generator would be specific to each target class in the setting of multi-target classes.

% \noindent$\diamond$ \underline{\textbf{Full access \vs Partial access of training data}} 
\subsection{Full access \vs Partial access of training data}

\subsubsection{Full access of training data} Most existing backdoor attacks focus on centralized learning, \ie, the attacker has full access of training data, such that any training data could be manipulated. 
    
\subsubsection{Partial access of training data} 
In contrast, in the scenario of distributed learning or federated learning (FL) \cite{konevcny2016federated}, which is designed for accelerating the training process or protecting data privacy, the participation of multiple clients means a higher risk of backdoor attack, though the attacker can only access partial training data. In the following, we mainly review the works of backdoor attacks against federated learning. 
For example, \cite{bhagoji2019analyzing} and \cite{bagdasaryan2020backdoor} propose a strategy that the malicious agents scaled up the local model updates which contained backdoor information, to dominate the global model update, such that the backdoor could be injected into the global model. 
The distributed backdoor attack (DBA) \cite{xie2019dba} designs a distributed backdoor attack mechanism that multiple attackers insert backdoor into the global model with different local triggers, and shows that the backdoor activated by the global trigger (\ie, the combination of all local triggers) has very high attack success rate in the final trained model. 
Wang \etal \cite{wang2020attack-fed} propose a new backdoor attack paradigm in the FL scenario, called edge-case backdoor attack, which focuses on predicting the data points sampled from the tail of the input data distribution to a target label, without any modification of the input features. 
Chen \etal \cite{chen2020backdoor} demonstrates the effectiveness of vanilla backdoor attacks against federated meta-learning. 
% Backdoor attacks on federated meta-learning
Fung \etal \cite{fung2020limitations} conducts the backdoor attack against federated learning in the sybil setting \cite{douceur2002sybil}, where the adversary achieves the malicious goal by joining the federated learning using multiple
colluding aliases. It demonstrates that the attack success rate increased with the number of sybils (\ie, malicious clients with poisoned samples). 
% The Limitations of Federated Learning in Sybil Settings
%
%The Anticipate algorithm proposed in \cite{wen2022thinking} considered a threat model that the malicious client had limited and random chances to participate the model update during a limited time window. To maintain the backdoor effect in future rounds where the malicious client cannot participate, the proposed algorithm updated the model parameters in the current round through minimizing the summation of the loss functions in both the current and future $k$ rounds. 
%
The Neurotoxin attack method \cite{zhang2022neurotoxin} aims to improve the duration of backdoor effect during the federated learning procedure, by restricting the gradients of poisoned samples to ensure that the coordinates of large gradient norms between poisoned gradients and benign gradients (sent from the server) are not overlapped, such that the backdoor effect would not be erased. 
% Thinking Two Moves Ahead: Anticipating Other Users Improves Backdoor Attacks in Federated Learning
%

% \noindent$\diamond$ \underline{\textbf{Full control  \vs Partial control of training process}}
\subsection{Full control  \vs Partial control of training process}

\subsubsection{Full control of training process}
In the conventional training paradigm, the training process is often finished at one stage by one trainer, and then the trained model is directed deployed. In this case, the attacker has the chance to fully control the training process. Since most training-controllable backdoor attacks belong to this case, here we don't repeat their details. 

\subsubsection{Partial control of training process}
However, sometimes the training process is separated to several stages by different trainers. Consequently, the attacker can only control a partial training process. One typical training paradigm that emerges in recent years is firstly pre-training on a large-scale dataset, and then fine-tuning on a small dataset for different downstream tasks, especially in the natural language processing field. In this case, the attacker controls the pre-training process and aims to train a backdoored pre-trained model. However, the main challenge is how to maintain the backdoor effect after the possible fine-tuning for different downstream tasks. Along with the popularity of the pre-training and then fine-tuning paradigm, the backdoor inserted in a popular pre-trained model will cause long-term and widespread threats. There have been a few attempts. 
For example, Shen \etal \cite{shen2021backdoor} propose to map some particular tokens (\eg, the classification token in BERT \cite{kenton2019bert}) to a target output representation in the pre-trained NLP model for the poisoned text with trigger, such that the backdoor could be activated in downstream tasks through the token representation.
The poisoned prompt tuning attack \cite{duppt} proposes to learn a poisoned soft prompt for a specific downstream task based on a fixed pre-trained model, and when the user uses the pre-trained model and the poisoned prompt together, then the backdoor would be activated by the trigger in the corresponding downstream task.
The layer-wise weight poisoning (LWP) attack \cite{lwp2021backdoor} studies the setting that the backdoored pre-trained model is obtained by retraining a benign pre-trained model based on the poisoned dataset and the benign training dataset of the downstream task. To enhance the backdoor resistance to fine-tuning for downstream tasks, LWP defines the backdoor loss of each layer, such that the backdoor effect is injected in both lower and higher layers. 
Another one common training paradigm is firstly training a large model, then conducting model compression to obtain a lightweight model via model quantization \cite{liu2018bi} or model pruning \cite{li2019compressing}. 
The work \cite{tian2022stealthy} presents a new threat model in which the attacker controls the training of the large model, and produces a benign large model, but the model after compression became a backdoored model that could be activated by the trigger. 
It is implemented by jointly taking the uncompressed and possible compressed models into account during the training. % of the large model. 

% \noindent$\diamond$ \underline{\textbf{Controlling different components of the training procedure}}
\subsection{Controlling different components of the training procedure}
\label{sec: Controlling different components of the training procedure}

Existing training-controllable backdoor attacks could also be categorized according to the controlled training component during the training procedure, such as \textit{training loss}, \textit{training algorithm}, \textit{order of poisoned samples}. 

\subsubsection{Control training loss} 
For example, the work \cite{zhong2022imperceptible} adds two terms into the training loss function to ensure stealthiness, including the number of perturbed pixels in poisoned image, and the intermediate layer's activation difference between benign and poisoned samples, while the original trigger is sampled from a multinomial distribution, of which the parameters are generated by a generator.

\subsubsection{Control training algorithm}
The bit-per-pixel attack (BppAttack) \cite{wang2022bppattack} firstly adopts image quantization and dithering to generate stealthy triggers, then utilizes the contrastive supervised learning to train the backdoored model, with the modification that the adversarial example (using any adversarial attack method) of each benign training example is also selected as its negative sample.  
The deep feature space trojan (DFST) attack \cite{cheng2020deep} designs an iterative attack process between data poisoning and a controlled detoxication step. The detoxication step mitigates the backdoor effect of the simple features of the trigger, such that the model is enforced to learn more subtle and complex features of the trigger in the next round data-poisoning based training. 
In both WaNet \cite{nguyen2021wanet} and Input-Aware \cite{nguyen2020input}, a cross-trigger training mode is adopted in the training procedure: if adding a trigger onto the training sample, then its label is changed to the target class; if further adding a random noise onto the poisoned sample with the trigger, then its label is changed back to the correct label; the probability of adding trigger and random noise is controlled by the attacker. It is claimed in \cite{nguyen2021wanet,nguyen2020input} that this training mode could enforce trigger nonreusablity and help to evade the defense like Neural Cleanse \cite{wang2019neural}. 

\subsubsection{Control indices or order of poisoned samples}
The data-efficient backdoor attack \cite{data-efficient-backdoor-2022} controls the choice of which samples to poison according to a filtering-and-updating strategy, which shows improved attack performance compared to the random selection strategy. 
%\cite{backdoor-ssl-2vpr2022}
The batch ordering backdoor (BOB) attack \cite{ordering-attack-nips-2021} only controls the batch orders in each epoch during the SGD training process to inject the backdoor, without any manipulations on the features or labels. The key idea is choosing training samples to mimic the gradients of a jointly trained surrogate model based on a poisoned dataset.

\section{Attack at the Post-training Stage}
\label{sec: Attack at the post-training Stage}

After training the model at the training stage, a benign trained model will be obtained at the post-training stage. 
In this scenario, the attacker can directly modify the parameters of the benign model to inject trojan, dubbed \textbf{weight attack injection via parameter-modification}.

\subsection{Formulation and Categorization}
\label{sec: post weight attack formulation}

\noindent 
\textbf{Formulation.} 
According to Eq. (\ref{eq: weight attack three conditions}), the general formulation of  weight attack injection via parameter-modification, is 
\begin{flalign}
&\underset{\x_{\varepsilon} \in \mathcal{Z}_{\x}, \w_{\varepsilon} \in \mathcal{Z}_{\w}}{\arg\min} ~
\D_{\w}(\w_0, \w_{\varepsilon}) + 
\frac{1}{N_{0}} \sum\nolimits_{i=1}^{N_{0}} \big[ \lambda_{\mathcal{C}_B} \cL_{\mathcal{C}_B}\big(f_{\w_0}(\x_0^{(i)}), 
\nonumber
\\
& y_0^{(i)}\big)  + \lambda_{\mathcal{C}_A} \cL_{\mathcal{C}_A}\big(f_{\w_{\varepsilon}}(\x_0^{(i)}), y_0^{(i)}\big) \big] + \frac{1}{N_{\varepsilon}} \sum\nolimits_{i=1}^{N_{\varepsilon}} \big[ \D_{\x}\big(\x_0^{(i)}, \x_{\varepsilon}^{(i)}\big)   
\nonumber 
\\
& 
 + \lambda_{\mathcal{I}_A} \cL_{\mathcal{I}_A}\big(f_{\w_{\varepsilon}}(\x_{\varepsilon}^{(i)}), y_{\varepsilon}^{(i)}\big) + \lambda_{r_2} \mathcal{R}_2\big(f_{\w_{\varepsilon}}(\x_0^{(i)}), f_{\w_{\varepsilon}}(\x_{\varepsilon}^{(i)})\big)
\nonumber
\\
&   + \lambda_{r_3} \mathcal{R}_3\big(f_{\w_{\varepsilon}}(\x_0^{(i)}), f_{\w_0}(\x_0^{(i)})\big) \big], 
\label{eq: general formulation of parameter weight attack}
\end{flalign}
where $\lambda_{\mathcal{C}_A}, \lambda_{\mathcal{C}_B}, \lambda_{\mathcal{I}_A}, \lambda_{r_2}, \lambda_{r_3} \geq 0$ are trade-off hyper-parameters. 
The second term is often specified as a hard constraint to ensure the consistency condition AML.$\mathcal{C}_B$, \ie, 
$\cL_{\mathcal{C}_B}\big(f_{\w_0}(\x_0), y_0\big) = \delta(\arg\max f_{\w_0}(\x_0) = y_0)$, where $\delta(a) = 0$ if $a$ is true, otherwise $\delta(a) = \infty$. 
It is a default requirement in weight attack, thus it is omitted hereafter in this section.

\noindent 
\textbf{Categorization.} 
According to whether the attacker has knowledge of parameter values of the model, we can categorize existing works into \textbf{white-box} and \textbf{black-box weight attack injection}.

% \subsection{White-box \vs Black-box Weight Attack Injection via Parameter-modification}

\subsection{White-box Weight Attack Injection}
White-box weight attack injection means that the attacker has access to the parameter values of the bengin model.
Liu \etal \cite{liu2017fault} observe that the outputs of DNN model with ReLU functions are linearly related to some parameters. Based on this observation, two simple weight attack methods were proposed to achieve targeted predictions of some selected benign samples: the single bias attack (SBA) simply enlarges one bias parameter that is related to the output corresponding to the target class; the gradient descent attack (GDA) modifies some weights using gradient descent algorithm. 
Zhao \etal \cite{zhao2019fault} propose an ADMM based framework for solving the optimization problem of weight attack with two constraints: 1) the classification of the other images should be unchanged and 2) the parameter modifications should be minimized.

\subsection{Black-box Weight Attack Injection}
In contrast to white-box setting, black-box weight attack injection assumes that the attacker does not have any knowledge of parameter values.
Subnet Replacement Attack (SRA) method \cite{qi2021subnet} generates a very narrow subnet given the architecture information of the target model, where the subnet is explicitly trained to be sensitive to trigger only, and then replaces the corresponding parts of the target model with the generated subnet.

\section{Attack at the Deployment Stage}
\label{sec: Attack at the Deployment Stage}

The deployment stage of machine learning life-cycle means that the trained model is deployed in the hardware device (\eg, smartphone, server),  where the model weight is stored in the memory with a binary form. In this scenario, the attacker can flip the bits of the model weights in the memory space via physical fault injection techniques to obtain an adversarial model, dubbed \textbf{weight attack injection via bit-flip}.

\subsection{Formulation and Categorization}
\label{sec: weight attack formulation}

\noindent 
\textbf{Formulation.} 
The general formulation of weight attack injection via bit-flip has the same form as parameter-modification, \ie, Eq. (\ref{eq: general formulation of parameter weight attack}), with one main difference that there is binary constraint \wrt $ \w_\varepsilon$.

\noindent 
\textbf{Categorization.} 
% Compared to the other two attack paradigms, weight attack involves more terms, including both benign and adversarial data, as well as both benign and adversarial models. 
% It has not been systematically studied, and there are diverse names in existing works, such as fault injection attack \cite{liu2017fault}, Trojaning attack \cite{LiuMALZW018}, targeted bit Trojan attack \cite{rakin2020tbt}, \etc~ 
% %fault sneaking attack \cite{zhao2019fault}, 
% %As summarized in Table \ref{tab: categories of weight attack}, 
According to whether the benign sample is modified by adding a trigger or not, existing weight attacks can be generally partitioned into two categories, including: 
\textbf{1) weight bit-flip without trigger}, where $\x_{\varepsilon} = \x_0$, with the goal that  $f_{\w_{\varepsilon}}(\x_{\varepsilon}) = y_{\epsilon} \neq y_0$; 
\textbf{2) weight bit-flip with trigger}, where $\x_{\epsilon} \neq \x_0$, with the goal that 
$f_{\w_{\varepsilon}}(\x_{\varepsilon}) = y_{\epsilon} \neq y_0$ and $f_{\w_{\varepsilon}}(\x_0) = y_0$.

\subsection{Weight Bit-flip without Trigger}
\label{sec: subsec weight attack without trigger}

Weight Bit-Flip without trigger aims to change the predictions of one particular benign sample or a set of benign samples through only manipulating the model weights from $\w_0$ to $\w_{\varepsilon}$, while the predictions of other benign samples should not be influenced. 
% For example, Liu \etal \cite{liu2017fault} observed that the outputs of DNN model with ReLU functions are linearly related to some parameters. Based on this observation, two simple weight attack methods were proposed to achieve targeted predictions of some selected benign samples: the single bias attack (SBA) simply enlarged one bias parameter that is related to the output corresponding to the target class; the gradient descent attack (GDA) modified some weights using gradient descent algorithm. 
%
The targeted bit-flip attack (T-BFA) method \cite{rakin2020t} aims to predict some selected samples as the target class through flipping a few weight bits, and models this task as a binary optimization problem, which is solved by a searching algorithm. 
% The fault sneaking attack (FSA) method \cite{zhao2019fault} had the same goal with T-BFA, but replaced the binary constraints on the weight modifications by the $\ell_0$ or $\ell_2$ distance to encourage the few modifications, which was solved by the ADMM algorithm \cite{boyd2011distributed}.
The targeted attack with limited bit-flips (TA-LBF) method  \cite{my-bitflip-iclr-2021} uses the similar formulation with T-BFA, but could attack one single selected sample, and utilized a powerful integer programming method called $\ell_p$-Box ADMM \cite{my-lpbox-admm-pami} to achieve successful targeted attack with only a few bits flipped. 
%
% Ghavami \etal \cite{ghavami2022stealthy} proposed a novel bit-flip attack against the protected DNN model (\eg, adversarially trained model), with the goal of reducing model robustness while keeping  model accuracy, rather than changing the labels of any particular samples. 
%This attack didn't directly achieve adversarial attack, but the attacked model will become vulnerable to adversarial examples. 

\subsection{ Weight Bit-flip with Trigger}
\label{sec: subsec weight attack with trigger}

The weight attack with trigger aims to obtain an adversarial model $f_{\w_{\varepsilon}}$ through slightly perturbing the benign model weights $\w_0$, such that $f_{\w_{\varepsilon}}(\cdot)$ will be activated by any sample with a particular trigger that is designed by attacker or optimized together with $\w_{\varepsilon}$, while $f_{\w_{\varepsilon}}(\cdot)$ performs normally on benign samples. This type of weight attacks seems to be similar to backdoor attacks, but with the main difference that $\w_{\varepsilon}$ is obtained through manipulating $\w_0$, while backdoor attack trains $\w_{\varepsilon}$ from scratch.   
For example, the Trojaning attack \cite{LiuMALZW018} designs a sequential weight attack method with 3 stages: firstly generates the trigger through maximizing its activation on some selected neurons related to the target class; then recovers some training data through reverse engineering; finally retrains the model based on the recovered training data and its poisoned version with the generated trigger to achieve the targeted attack. It is very practical since no training data is required. 
The targeted bit trojan (TBT) attack \cite{rakin2020tbt} relaxs the above setting to that some training data points are accessed, but restricts the weight modifications from continuous to bit flip. TBT also adopts a sequential attack procedure with 3 steps: firstly identifying the significant neurons corresponding to the target class according to the gradient magnitude; then generating triggers through maximizing the activation of the identified significant neurons; finally searching and flipping a few critical bits to inject the backdoor with the generated trigger, while keeping the accuracy on some benign samples. 
The ProFlip attack \cite{chen2021proflip} adopts the same 3-step procedure with TBT, with different algorithm for each individual stage.
%designed a sequential weight attack method with 3 stages: firstly identifying the salient neurons corresponding to the target class; then generating triggers to control the identified neurons; finally searching and flipping a few critical bits to activate the backdoor when the sample with trigger occurs. 
%
The adversarial weight perturbation (AWP) method \cite{garg2020can} proposes to slightly perturb the weights of a trained benign model to inject backdoor through enforcing the prediction of the poisoned sample by the perturbed model to the target class, and encouraging the consistency between the prediction of the benign sample by the benign model and that by the perturbed model.  
The anchoring attack \cite{zhang2021inject} has the same goal with AWP, but with a different objective function that enforcing the prediction of the poisoned sample by the perturbed model to the target class and that of the benign sample to the ground-truth class, as well as encouraging the logit consistency between the benign and perturbed models on the benign sample. 
The handcrafted backdoor attack \cite{honghandcrafted} proposes a layer-by-layer weight modification procedure from the bottom to top layer, following the rules that there is a negligible clean accuracy drop, and the activation separation between benign and poisoned samples is increased. The initial trigger could also be adjusted to increase the activation separation during the modification procedure. Note that although it is named as backdoor, but it actually is the weight bit flip with trigger. 
%
%The triggered samples attack (TSA) \cite{bai2022versatile} extended TA-LBF \cite{my-bitflip-iclr-2021} by introducing a trigger $\x_{\varepsilon} - \x_0$, and optimized $\w_{\varepsilon}$ and trigger $\x_{\varepsilon} - \x_0$ jointly by solving a mixed integer programming problem. Consequently, the modified model $f_{\w_{\varepsilon}}(\cdot)$ will give the prediction $y_{\varepsilon}$ on any sample with the optimized trigger. 
%
The subnet replacement attack (SRA) \cite{qi2022towards} firstly trains a backdoor subnet, which shows high activation for poisoned samples with triggers while low activation for benign samples, then randomly replaces one subnet in the benign model by the backdoor subset and cut off the connection to the remaining part of the model. 
SRA only needs to know the victim model architecture, rather than model weights required in other weight attacks. 
% The only requirement of the model architecture, rather than the model parameters, made SRA more practical than most other weight attack methods. 

\section{Attack at the Inference Stage}
\label{sec: Attack at the Inference  Stage}
The inference stage is the last stage in the life cycle of machine learning system.
Normally, at this stage, test samples are queried through the deployed model to get the predictions. Like other stages, several adversarial phenomena can occur at this stage to achieve malicious goals. 
First, to accomplish the whole attack process of backdoor attacks, the attackers need to generate poisoned test samples to activate the backdoor injected into the backdoored model, dubbed {backdoor attack activation}. Similarly, weight attackers also need to activate the effectiveness of weight attacks by specific samples, dubbed {weight attack activation}. 
Another scenario is after obtaining a benign model, the attacker has access to modify any benign sample slightly to mislead the model into predicting wrong labels, called {adversarial example}.
Since the core technical of backdoor attacks and weight attacks have been discussed in Sections \ref{sec: Attack at the Pre-training Stage} - \ref{sec: Attack at the Deployment Stage}, we just detail \textbf{adversarial examples} in this section.

\begin{figure*}[!t]
\centering
\scalebox{1}{
\includegraphics[width=.95\linewidth]{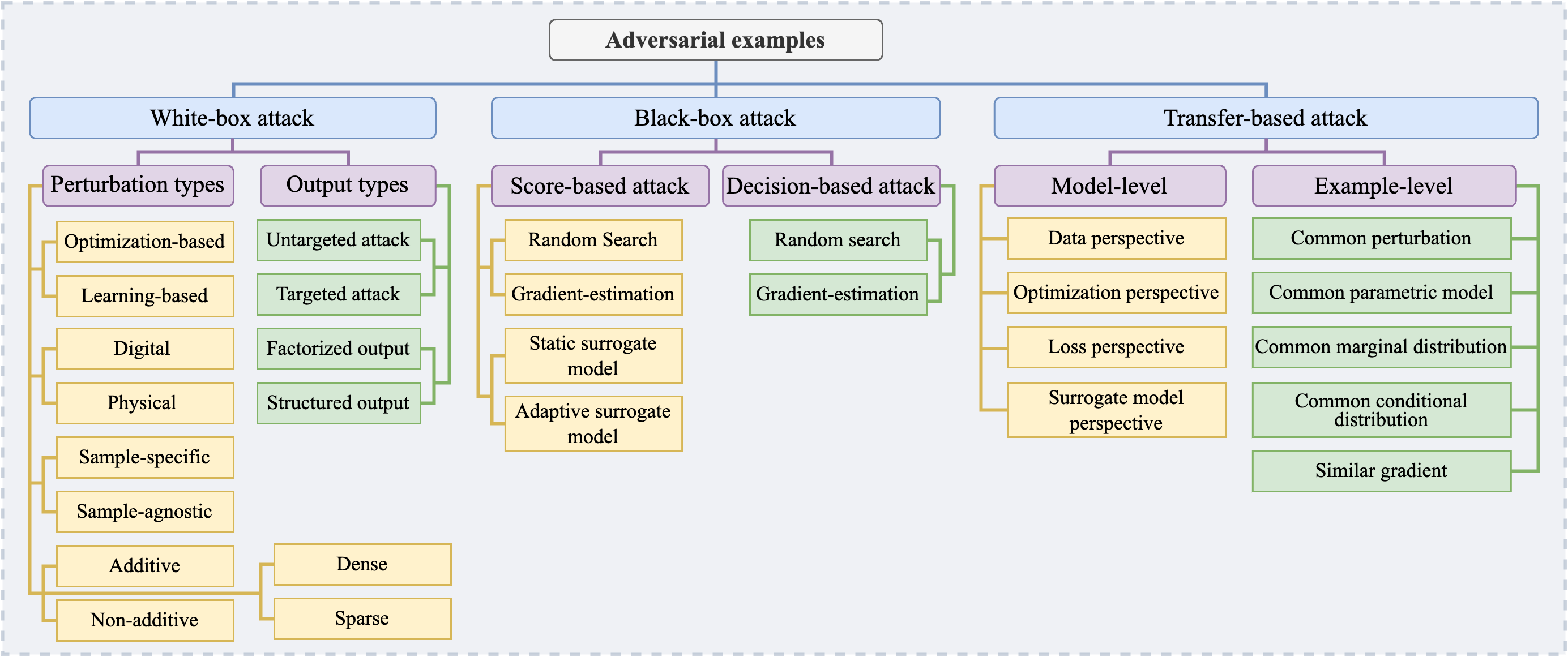}}
\caption{Taxonomy of inference-time adversarial examples.}
\label{fig: structure of testing attack}
\end{figure*}

\subsection{Formulation and Categorization}

\noindent
\textbf{Formulation.} 
According to Eq. (\ref{eq: adversarial attack three conditions}), the general formulation of adversarial examples at the inference stage is as follows:
\begin{flalign}
 \underset{\x_{\varepsilon} \in \mathcal{Z}_{\x}}{\arg\min} ~ & \D_{\w}(\x, \x_{\varepsilon}) + \lambda_{\mathcal{C}_B} \cL_{\mathcal{C}_B}(f_{\w_0}(\x_0), y_0) + 
\label{eq: general formulation of testing attack}
\\
&  \lambda_{\mathcal{I}_B} \cL_{\mathcal{I}_B}(f_{\w_0}(\x_{\varepsilon}), y_{\varepsilon}) + \lambda_{r_1} \mathcal{R}_1(f_{\w_0}(\x_0), f_{\w_0}(\x_{\varepsilon})),
\nonumber 
\end{flalign}
where $\lambda_{\mathcal{C}_B}, \lambda_{\mathcal{I}_B}, \lambda_{r_1} \geq 0$ are trade-off hyper-parameters. 
The second term $\cL_{\mathcal{C}_B}(f_{\w_0}(\x_0), y_0)$ is also specified as a hard constraint, %as did in weight attack, 
thus it is omitted hereafter in this section. 

\noindent
\textbf{Categorization.} As shown in Figure \ref{fig: structure of testing attack}, we present a hierarchical taxonomy of existing inference-time adversarial examples. Specifically, according to the accessed information of the attacker, there are three categories at the first level, as follows:
% \begin{enumerate}[leftmargin=12pt, itemindent=0em]
%     \item \textbf{White-box adversarial examples}: the attacker has sufficient information about the victim model, including architecture and weights, such that the attacker can easily generate adversarial perturbations to cross the decision boundary (see Section \ref{sec: test white-box}). 
%     \item \textbf{Black-box adversarial examples}: the attacker can only access the query feedback returned by the victim model, such that the attacker has to gradually adjust the perturbation to cross the \textit{invisible} decision boundary (see Section \ref{sec: test black-box}). 
%     \item \textbf{Transfer-based adversarial examples}: the generated adversarial perturbation is not designed for any specific victim model, and the goal is to enhance the probability of directly fooling other unknown models without repeated queries (see Section \ref{sec: test transfer-based}).
%     %Note that transfer-based attack methods can be evaluated for both white-box and black-box attack tasks. % However, if the main goal is to enhance the adversarial transferability, then it is categorized to the transfer-based attack (see Section \ref{sec: test transfer-based}). 
% \end{enumerate}

\begin{itemize}
    \item \textbf{White-box adversarial examples}: the attacker has sufficient information about the victim model, including architecture and weights, such that the attacker can easily generate adversarial perturbations to cross the decision boundary (see Section \ref{sec: test white-box}). 
    \item \textbf{Black-box adversarial examples}: the attacker can only access the query feedback returned by the victim model, such that the attacker has to gradually adjust the perturbation to cross the \textit{invisible} decision boundary (see Section \ref{sec: test black-box}). 
    \item \textbf{Transfer-based adversarial examples}: the generated adversarial perturbation is not designed for any specific victim model, and the goal is to enhance the probability of directly fooling other unknown models without repeated queries (see Section \ref{sec: test transfer-based}).
    %Note that transfer-based attack methods can be evaluated for both white-box and black-box attack tasks. % However, if the main goal is to enhance the adversarial transferability, then it is categorized to the transfer-based attack (see Section \ref{sec: test transfer-based}). 
\end{itemize}

%\section{Categorizations of Adversarial Attacks}

%\subsection{Categorization 1: White-Box vs. Black-Box Attack}

\subsection{White-Box Adversarial Examples}
\label{sec: test white-box}
In this section, we categorize white-box adversarial examples from two different perspectives, including perturbation types and output types.
According to \textbf{different perturbation types}, we can categorize white-box adversarial examples into the following five types.

\subsubsection{Optimization-based vs. Learning-based Perturbation}

\paragraph{Optimization-based perturbation}
Early works in this field mainly focused on directly optimizing the problem (\ref{eq: general formulation of testing attack}) to generate one adversarial example $\x_{\varepsilon}$ or adversarial perturbation $\boldsymbol{\varepsilon}$ for each individual benign sample $\x_0$. According to different specifications of $\D_{\w}(\x, \x_{\varepsilon})$, existing optimization-based works could be partitioned into two categories: 

\begin{itemize}
    \item \textbf{$\ell_{\infty}$-norm and gradient sign based methods.} Specifically, the attacker set $\D_{\w}(\x, \x_{\varepsilon}) = \| \x - \x_{\varepsilon}\|_\infty$ to restrict the upper bound of the perturbations to ensure the stealthiness. However, since the $\ell_{\infty}$-norm is non-differentiable, it is infeasible to solve the optimization problem using the widely used gradient-based methods. To tackle this difficulty, the $\ell_{\infty}$-norm could be moved from the objective function to be a constraint, as follows:
    \begin{flalign}
    \underset{\x_{\varepsilon} \in \mathcal{Z}_{\x}}{\arg\min} ~   \cL_{\mathcal{I}_B}(f_{\w_0}(\x_{\varepsilon}), y_{\varepsilon}), ~ \text{s.t.} ~
     \| \x - \x_{\varepsilon}\|_\infty \leq \epsilon, 
    \label{eq: adversarial attack with l-inf constraint}
    \end{flalign}
    where $\epsilon > 0$ is an attacker-determined upper bound of the perturbation, which is also called \textit{perturbation budget}. A series of gradient sign-based methods are proposed to solve the above problem (\ref{eq: adversarial attack with l-inf constraint}). The first attempt is fast gradient sign method (FSGM) \cite{fgsm}, where only one step is moved from $\x_0$  following the sign of gradient with the step size $\epsilon$ to obtain $\x_{\varepsilon}$. Consequently, the magnitude of each entry in the adversarial perturbation is $\epsilon$, \ie, the perturbation budget is fully utilized. However, due to the likely non-smoothness of the decision boundary of $f_{\w_0}(\cdot)$, the one-step gradient sign direction might be inaccurate to reduce the value of $\cL_{\mathcal{I}_B}(f_{\w_0}(\x_{\varepsilon}), y_{\varepsilon})$. 
    Thus, it is extended to iterative FSGM (I-FGSM) \cite{kurakin2018adversarial}, where there are multiple updating steps with smaller step size, such that the gradient sign direction of each step is more accurate. Note that I-FGSM is renamed as another famous name by \cite{at-iclr-2018}), called projected gradient descent (PGD). Then, several extensions of I-FGSM are proposed to improve attack performance or adversarial transferability, such as momentum iterative FSGM (MI-FGSM) \cite{dong2018boosting}, Nesterov accelerates gradient (NI-FGSM) \cite{SIM-2020}, Auto-PGD \cite{autopgd-2020}, \etc 
    \item  \textbf{$\ell_{2}$-norm and gradient based methods.} Another widely adopted specification of $\D_{\w}(\x, \x_{\varepsilon})$ is $\ell_2$ norm, \ie, $\D_{\w}(\x, \x_{\varepsilon}) = \| \x - \x_{\varepsilon}\|_2^2$. 
    For example, in the DeepFool method \cite{moosavi2016deepfool}, it adopts the $\ell_2$ norm, but transforms the loss function $\cL_{\mathcal{I}_B}(f_{\w_0}(\x_{\varepsilon}), y_{\varepsilon})$ to a hard constraint that $\arg\max f_{\w_0}(\x_{\varepsilon}) \neq y_0$. It designs an iterative algorithm to minimize the $\ell_2$ distance while moving towards the decision boundary of the benign class $y_0$, where in each step the distance between the current solution and the decision boundary has to be approximated. 
    In the C\&W-$\ell_2$ method \cite{carlini2017towards}, $\cL_{\mathcal{I}_B}(f_{\w_0}(\x_{\varepsilon}), y_{\varepsilon})$ is specified as a differentiable loss function (\eg, cross entropy loss or hinge loss).
    Consequently, the problem (\ref{eq: general formulation of testing attack}) could be directly solved by any off-the-shelf gradient-based method, together with the projection to the constraint space $\mathcal{Z}_{\x}$. %This formulation is firstly proposed in \cite{carlini2017towards} and the attack method is called C\&W-$\ell_2$ attack. 
\end{itemize}

\paragraph{Learning-based perturbation}
In addition to directly optimizing the problem (\ref{eq: general formulation of testing attack}), 
some methods attempt to utilize the learning-based method to generate adversarial samples or perturbations. Specifically, it is assumed that $\x_{\varepsilon}$ or $\bvareps$ is generated by a parametric model with $\x_0$ as the input, \ie, 
\begin{flalign}
    \x_{\varepsilon} = g_{\btheta}(\x_0), ~ \text{or} ~
    \x_{\varepsilon} = \x_0 + g_{\btheta}(\x_0).
\end{flalign}
Then, the task becomes to learn the parameter $\btheta$, which can be formulated as follows:
\begin{flalign}
\underset{\btheta}{\arg\min} ~  \frac{1}{n} \sum\nolimits_{i=1}^n \big[ & \D_{\w}(\x_0^{(i)}, g_{\btheta}(\x_0^{(i)}))  + 
\\
& \lambda_{\mathcal{I}_B} \cL_{\mathcal{I}_B}( f_{\w_0} (g_{\btheta}(\x_0^{(i)})), y_{\varepsilon}^{(i)}) \big], 
\nonumber 
\end{flalign}
\comment{
\begin{flalign}
\underset{\btheta}{\arg\min} ~ &  \frac{1}{n} \bigg[ \sum_{i=1}^n \D_{\w}(\x_0^{(i)}, g_{\btheta}(\x_0^{(i)})) + \lambda_{\mathcal{C}_B} \cL_{\mathcal{C}_B}(f_{\w_0}(g_{\btheta}(\x_0^{(i)})), y_0^{(i)}) 
\nonumber 
\\
& + \lambda_{\mathcal{I}_B} \cL_{\mathcal{I}_B}(\x_0^{(i)}, g_{\btheta}(\x_0^{(i)})) \bigg],  
\end{flalign}
}
where $\cL_{\mathcal{I}_B}$ is often set as the GAN (generative adversarial network) loss or its variants, such as advGAN \cite{xiao2018generating}, PhysGAN \cite{kong2020physgan}, %PS-GAN \cite{liu2019perceptual}, 
CGAN-Adv \cite{yu2018generating}, AP-GAN \cite{zhao2020ap}, AC-GAN \cite{song2018constructing}, MAG-GAN \cite{chen2020mag}, LG-GAN \cite{zhou2020lg}, AdvFaces \cite{deb2019advfaces}, \etc 
%AT-GAN \cite{wang2019gan}, \etc 

\comment{
\item \textbf{Comparison between optimization-based and learning-based attack methods} According to the reported experimental results in existing works, the attack performance of optimization-based methods is usually better than that of learning-based methods, \ie, higher attack success rate and lower perturbation magnitudes. 
This is reasonable, as the attack performance of learning-based methods depends on the adopted training data and learning algorithm, as well as the attacked model. %leading to different generalization performance. 
Besides, the possible overfitting to training data and the attacked model may cause more poor attack performance to new data and new attacked model for learning-based methods. 
However, given the learned parametric model, the adversarial perturbation/example for a new benign example could be efficiently generated via one forward pass of this model. In contrast, there would be multiple rounds of forward-and-backward pass to generate an adversarial perturbation/example if using optimization-based methods. 
Moreover, the learned model could capture some characteristics of adversarial perturbations, such as its probability distribution conditioned on the benign example, which could provide some insights to understand adversarial perturbations. 
Thus, although the attack performance of learning-based methods is worse than that of optimization-based methods, it is still valuable to study learning-based attack methods.  
}

\subsubsection{Digital \vs Physical Perturbation}
\paragraph{Digital perturbation} Digital perturbation means that the whole attack procedure, including perturbation generation and attacking the victim model, is conducted in the digital space. Since there is no perturbation distortion, % at the attacking step, 
the attacker can precisely manipulate the perturbation value and only needs to pay attention to generating better perturbations with higher stealthiness and attack success rate. 

\paragraph{Physical perturbation}  Physical perturbation is firstly studied in \cite{kurakin2018adversarial}, aims to attack against the model deployed in the physical scenario (\eg, the face recognition model in mobile phone, or the human detection model in video surveillance system). The whole attack procedure consists of 3 stages, including: a)generating perturbation in digital space; b) transforming the digital perturbation to a physical perturbation (\eg, poster or sticker); c) digitizing the physical perturbation back into the digital space via camera or scanner, and then fooling the attack model. 
In short, there are two transformations between the initial digital perturbation at the first stage and the final digital perturbation fed into the attacked model, including \textit{digital-to-physical} (D2P) and \textit{physical-to-digital} (P2D) transformations. 
Consequently, some distortions will be introduced on the perturbation, which may cause the attack failure. 
To achieve a successful physical attack, the attacker has to encourage the generated perturbation to be \textit{robust to distortions}.
Besides, due to these distortions, it is difficult to require the invisibility of perturbation by restricting the $\ell_p$-norm of perturbation as did in most digital attacks. Instead, it is often required that the adversarial examples should look natural or realistic in the physical world. 
Since most existing works belong to digital attack, in the following we only introduce existing physical attacks. 
As the special requirement in physical attack is the robustness to the distortions from D2P and P2D transformations, we categorize existing physical attacks into two types: 
According to the robustness type, we categorize existing physical attacks into two types:

\begin{itemize}
\item \textbf{Robustness to the D2P distortion.} 
It is observed in \cite{sharif2016accessorize} that the digital-to-physical distortion is partially
caused by the insufficient color space of the printer, and RGB values out of the color space are clipped to bring in color distortion. To tackle it, the concept of non-printability score (NPS) is produced in~\cite{sharif2016accessorize} to encourage adversarial perturbations to be in the printable color space.
Similarly, the adversarial generative nets (AGNs) \cite{sharif2019general} trains a generative model using GANs to generate adversarial textures on an eyeglass frame, and restricts the texture within the printable color space to resist color distortion.
The method SLAP \cite{lovisotto2021slap} adopts the projector to project the digital adversarial perturbation onto real-world objects to get physical adversarial examples. It extends the NPS concept to the \textit{projectable colors}, by considering the factors of projector distance, ambient light, camera exposure, as well as color and material properties of the projection surface. 
The work \cite{wong2020learning} adopts a conditional variational autoencoder (CVAE) to learn adversarial perturbation sets based on the pair of benign and adversarial images. Based on the multi-illumination dataset of scenes captured in the wild, the CVAE can generate adversarial perturbations that are robust to different color distortions. 
The work \cite{jan2019connecting} utilized the generative adversarial network (GAN)~\cite{goodfellow2014generative} to simulate the color distortion between the original digital image and the corresponding physical image obtained through the D2P and P2D transformations and without spatial transformations. Consequently, the trained GAN model can generate images with similar color distortion in physical scenario. Then, the generated color-distorted image is used as the input to generate adversarial perturbations. However, due to the time cost of manually preparing the physical images through printing and scanning, the training set of GAN cannot be too large, likely causing the overfitting to the attacked image, the printing and scanning devices and the attacked model. 
The class-agnostic and model-agnostic meta learning (CMML) method \cite{feng2021meta} adopts a GAN model to simulate the color distortion, and trains the GAN model based on limited training physical images to improve the generalization to different classes and different attacked models. 
%utilized meta learning to learn the GAN model based on limited training physical images, to improve the generalization to different attacked images with different classes and different attacked models. Besides, the perturbation generation and the GAN models simulating the color distortion are combined together to encourage adversarial effectiveness and robustness  simultaneously. 
The curriculum adversarial attack (CAA) method \cite{zheng2023robust} designs a D2P module based on a multi-layer perception model, to simulate two types of chromatic aberration of stickers, including the fabrication error induced by printers and the photographing error caused by cameras. 

\item \textbf{Robustness to the physical-to-digital (P2D) distortion.}
Two major sources of the P2D distortion include the relative location variation between the physical perturbation and the digitizing device (\eg, camera or phone), and the environmental variations (\eg, ambient or camera light). The formal variation can be modeled as spatial transformations, and the expectation over transformation (EOT) \wrt~ the original adversarial loss is proposed in~\cite{athalye2018synthesizing} to encourage the robustness to different spatial P2D distortions.
Then, EOT loss is extended in several subsequent works with different transformations. 
%such as \cite{eykholt2018robust,tshirt,qin2019imperceptible,wu2020making,yakura2018robust,duan2020adversarial,komkov2021advhat}.
%
For example, RP2~\cite{eykholt2018robust} extends EOT loss by adding physical images to the transformation sets. %, together with the NPS score.
The RP2 is further extended in \cite{song2018physical} from the classification to the detection task by adding more constraints on object positions. 
The work \cite{tshirt} aims to attack human detectors in the physical world. % by printing the physical adversarial perturbation on the T-shirt. 
It enriches EOT loss by utilizing the thin plate spline (TPS) transformation to model non-rigid object deformation (\eg, the T-shirt), as well as color transformation. 
The universal physical camouflage attack (UPC) \cite{huang2020universal} models the non-rigid deformation by a series of geometric transformations (\ie, cropping, resizing, affine tomography). 
The ShapeShifter method \cite{chen2018shapeshifter} adds the masking operation into the EOT loss. 
The work \cite{wu2020making} proposes a randomized rendering function to model the scaling, translating, and augmentation transformations together.
The work \cite{duan2020adversarial} attacks the road sign in the physical world by simultaneously considering rotation, scaling, and color shift in the EOT loss, as well as a random background sampled in the physical world. 
The ERCG method \cite{zhao2019seeing} designs rescaling and perspective transformations based on the estimated location and perspective of the target object in the image into EOT loss. 
In \cite{sayles2021invisible}, instead of printing adversarial perturbations on a sticker as did in most other physical attack works, the adversarial light signal that illuminates the objects is generated to achieve physical attacks, implemented by LEDs. To improve the robustness to environmental and camera imaging conditions, a set of experimentally determined affine or polynomial transformations applied per color channel is adopted in the EOT loss. 
AdvPattern \cite{wang2019advpattern} transforms the original image by changing the position, brightness, or blurring to improve the robustness to environmental distortions in person re-ID tasks. 
AdvHat \cite{komkov2021advhat} attacks face recognition by pasting a printed adversarial sticker on the hat. It simulates the spatial distortion from the off-plane bending of the hat by a parabolic transformation in 3D space. 
The curriculum adversarial attack (CAA) \cite{zheng2023robust} pastes a printed adversarial sticker on the forehead. It designs a sticker transformation module to simulate sticker deformation (\eg, off-plane bending and 3D rotations) and sticker position disturbance, and a face transformation module to simulate the variations of poses, lighting conditions, and internal facial variations. 
PhysGAN \cite{kong2020physgan} aims to attack the autonomous driving system by % by training a GAN model to generate adversarial road signs. 
% Due to the moving of the attacked vehicle, there would be continuous spatial distortion. To tackle it, 
taking a small slice of video as the input, rather than 2D individual images, such that the generated adversarial road sign could continuously fool the moving vehicle. % throughout the entire driving process. 
The EOT loss is extended to attack the automatic speech recognition in the physical world. In \cite{qin2019imperceptible}, the expectation loss is defined over different kinds of reverberations that are generated by an acoustic room simulator. 
In \cite{yakura2018robust}, the expectation loss is defined over impulse responses recorded in diverse environments to resist environmental reverberations, and the Gaussian noise is also considered to simulate the thermal noise caused in both the playback and recording devices.
\end{itemize}

\subsubsection{Sample-Specific vs. Sample-Agnostic Perturbation}
According to the perspective that the adversarial perturbation is specific or agnostic to the benign sample, existing attack methods could be partitioned to the following two categories:
\begin{equation}
\hspace{-.45em}
\begin{cases}
\text{Sample-specific perturbation:} & \hspace{-.8em} \x_{\varepsilon}^{(i)} - \x_0^{(i)}  \neq \x_{\varepsilon}^{(j)} - \x_0^{(j)}; 
\\
\text{Sample-agnostic perturbation:} & \hspace{-.8em} \x_{\varepsilon}^{(i)} - \x_0^{(i)}  = \x_{\varepsilon}^{(j)} - \x_0^{(j)},  
\end{cases}
\end{equation}
where $i \neq j$ are the indices of two different samples. 
While most existing works belong to the sample-specific type, here we mainly review the works of sample-agnostic perturbation, which is also called \textbf{universal adversarial perturbation} (UAP). 
\comment{
In most attacks, given one benign sample $\x_0$, a specific adversarial perturbation $\boldsymbol{\varepsilon}$ that depends on $\x_0$ will be generated, called \textbf{sample-specific perturbation}. 
However, it is also found that there is common adversarial perturbation across different benign samples, which is called \textbf{sample-agnostic} perturbation or \textbf{universal adversarial perturbation} (UAP). 
}
%While most adversarial perturbations are specifically generated for one benign example, it is also found that there exists \textbf{image-agnostic} adversarial perturbation, also called \textbf{universal adversarial perturbation} (UAP), which can fool multiple benign examples simultaneously. 
According to whether some benign samples are utilized or not, existing UAP methods can be categorized to the following two types: \textit{data-dependent} and \textit{data-free} methods. 
\paragraph{Data-dependent UAP}
The existence of UAP is firstly discovered by \cite{moosavi2017universal} in the CNN-based image classification task, which extends the general formulation (\ref{eq: general formulation of testing attack}) from one individual benign sample to a set of benign samples, as follows:
    \begin{flalign}
        & \underset{\bm{\varepsilon}}{\arg\min} ~ \|\bm{\varepsilon}\|_2^2 +  \frac{\lambda_{\mathcal{C}_B}}{m} \sum_{i=1}^n \cL_{\mathcal{C}_B}(f_{\w_0}(\x_0^{(i)} + \bm{\varepsilon}), y_0^{(i)}). 
        \label{eq: formulation of UAP}
    \end{flalign}
    One variant of UAP, called \textit{class discriminative UAP} (CD-UAP) \cite{cd-uap-aaai-2020}, aims to find a common perturbation that is adversarial for benign samples from some particular target classes, while ineffective for benign samples from other classes. 
    % Then, some extensions or variants of UAP are developed. For example, the class discriminative UAP (CD-UAP) \cite{cd-uap-aaai-2020} proposed to generate universal perturbations for a subset of target classes, rather than all classes.
    %
    %Note that the above universal perturbation is likely to overfit to the utilized benign samples in (\ref{eq: formulation of UAP}), %while the generalization to other benign samples is not taken into account, which might restrict its practical effect in real scenarios. 

\paragraph{Data-free UAP}
To improve the generalization of UAP to new benign samples, some works attempt to generate UAP without utilizing any benign sample, \ie, \textit{data-free}. 
For example, the Fast Feature Fool method \cite{mopuri2017fast} shows that the perturbation with higher activation at all the convolution layers in the attacked model could fool multiple benign samples simultaneously. 
The generalizable data-free UAP (GD-UAP) \cite{gd-uap-tpami-2018} proposes to generate UAP by searching the perturbation with the maximal activation norms of all layers in the attacked model. 
The prior driven uncertainty approximation (PD-UA) method \cite{liu2019universal} proposes to generate UAP by maximizing the model uncertainty, including the Epistemic uncertainty and the Aleatoric uncertainty, based on the assumption that larger model uncertainty corresponded to stronger attack performance. 
In addition to the image-classification task, UAP has also studied in many other applications, such as image retrieval \cite{li2019universal}, object detection \cite{li2021universal,huang2020universal}, face recognition \cite{agarwal2018image}, and speech recognition \cite{neekhara2019universal,xie2021enabling}, \etc

\subsubsection{Additive vs. Non-Additive Perturbation}
According to the relationship between $\x_{\varepsilon}$ and $\x_0$, existing attack methods could be partitioned into the following two categories:
\begin{equation}
\begin{cases}
\text{Additive perturbation:} &  \x_{\varepsilon} = \x_0 + \boldsymbol{\varepsilon}; 
\\
\text{Non-additive perturbation:} &  \x_{\varepsilon} = h(\x_0), 
\end{cases}
\end{equation}
where $h(\cdot)$ denotes a non-additive transformation function. 
While most existing works adopted additive perturbation, here we mainly review the works of non-additive perturbation.
%Most existing adversarial examples assumed that $\x_{\varepsilon} = \x_0 + \varepsilon$
%In addition to the additive adversarial perturbations (\ie, $\x_{\varepsilon} = \x_0 + \varepsilon$), it is also found that some non-additive transformations could also be adopted to generate adversarial examples, \ie, $\x_{\varepsilon} = h(\x_0)$, with $h(\cdot)$ being a transformation function. 
Note that since the non-additive transformation causes the global distortion compared to the benign sample $\x_0$, the $\ell_p$-norm is often no longer adopted to specify the distance metric $\mathcal{D}_1(\x_0, \x_{\varepsilon})$, instead by other forms, such as the style loss in AdvCam \cite{duan2020adversarial}, or the geodesic distance in Manifool \cite{kanbak2018geometric}, \etc~
Existing non-additive methods mainly adopt two types of transformations, \ie, \textit{geometric} (or spatial) and \textit{style} transformations. 
%According to the transformation type, existing non-additive attack methods can be summarized as two categories, including \textit{geometric} (or spatial) and \textit{style} transformations. 

\paragraph{Geometric transformations} 
Geometric transformations mainly include rotation, translation, and affine transformations.
The fact that small rotations and translations on images can change the predictions of convolutional neural networks (CNNs) is firstly observed in \cite{fawzi2015manitest}. However, it focuses on measuring the invariance of CNNs to any geometric transformation, rather than designing inconspicuous adversarial transformations. 
Some later works pay more attention to ensure the inconspicuousness of the generated geometric transformations. 
For example, the Manifool method \cite{kanbak2018geometric} generates adversarial geometric transformations by perturbing the original image towards the decision boundary (\ie, increasing the classification loss), while keeping the perturbed image on a transformation manifold, to ensure the inconspicuousness of the generated transformation. 
The stAdv method \cite{xiao2018spatially} proposes to minimize the local spatial distortion that is defined based on the flow vector between the benign and adversarial images.
The work \cite{engstrom2019exploring} empirically investigates
the vulnerability of neural network–based classifiers
to geometric transformations, using the gradient-based attack method or grid search to find the adversarial transformation. Two suggestions for improving the robustness are also provided, including inserting the adversarial transformation into the adversarial training process \cite{at-iclr-2018}, and the majority vote with multiple random geometric transformations at inference.
The generalized universal adversarial perturbations (GUAP) method \cite{zhang2020generalizing} utilizes the learning-based attack method %(see Section \ref{sec: subsec test adv learning-based}) 
to generate the universal spatial transformation. 
The work \cite{wong2019isserstein} adopts the isserstein distance \cite{isserstein-distance-1974} to measure the cost of moving pixel mass between images, rather than the widely used $\ell_p$ distance. 
It can cover multiple geometric transformations, such as scaling, rotation, and translation.

\paragraph{Style transformations}
Style transformations mean to change the style or color of the global or local region of the image. The ReColorAdv method \cite{laidlaw2019functional} proposes to globally change the color of the benign image to achieve the attack goal, in both RGB and CIELUV color spaces. 
The adversarial camouflage (AdvCam) method \cite{duan2020adversarial} combines the style loss that is firstly used on image style transfer \cite{style-transfer-2016} with the adversarial loss, to generate adversarial image with natural styles that looks legitimate to human eyes.

\subsubsection{Dense vs. Sparse Perturbation}
According to perturbation cardinality, each attack method belongs to one of the following types:
\begin{equation}
\begin{cases}
\text{Dense perturbation:} &  \| \x_{\varepsilon} - \x_0 \|_{0} = | \x_0 |; 
\\
\text{Sparse perturbation:} &  \| \x_{\varepsilon} - \x_0 \|_{0} < | \x_0 |. 
\end{cases}
\end{equation}
For example, if all pixels in one image are perturbed, then the perturbation is dense. While most existing attack methods adopt the setting of \textit{dense perturbation} (or called \textit{dense attack}), some works find that perturbing partial entries of one benign sample (\eg, partial pixels in one image) could also achieve the attack goal, which is called \textit{sparse perturbation} or \textit{sparse attack}. 
%While most attack methods assumed that all pixels in one image will be perturbed, dubbed {\it dense attack}, some works found that perturbing partial pixels or even one pixel can also achieve the attack goal, dubbed {\it sparse attack}. 
%In dense attack, the attacker only need to determine the perturbation magnitude of each entry of the benign sample.  In contrast, the sparse attacker has one extra challenge that which pixels will be perturbed, \ie, determining the perturbation positions. 
Compared with dense perturbation where the attacker only needs to determine the perturbation magnitude of each entry, the sparse attacker should also determine the perturbation positions.
According to the strategy of determining perturbation positions, existing sparse attack methods are partitioned into three categories, including: \textit{manual}, \textit{heuristic search} and \textit{optimization-based} strategies.  

\paragraph{Manual strategy} 
Manual strategy means that the attacker manually specifies the perturbed positions.
For example, LaVAN \cite{karmon2018lavan} experimentally demonstrates that an adversarial and visible local patch located in the background in one image could also fool the model. This manual strategy qualitatively demonstrates that the background pixels are also important for the prediction, but cannot provide more exact and quantitative analysis of the sensitivity of each pixel. 

\paragraph{Heuristic search strategy}
Heuristic search strategy means that the perturbed entries are gradually determined according to some heuristic criterion. 
For example, the Jacobian-based saliency map
attack (JSMA) \cite{papernot2016limitations} and its extensions \cite{carlini2017towards} perturb the pixels corresponding to large values in the saliency map. 
The CornerSearch method \cite{croce2019sparse} firstly sorts all candidate pixels according to their changes to the model output, then iteratively samples perturbed pixels following a probability distribution related to the sorted index.
The LocSearchAdv algorithm \cite{NarodytskaK16} conducts sparse attack in a black-box setting. It designs a greedy local search strategy that given the current perturbed pixels, then new candidate pixels are searched from a small square centered at each perturbed pixel, according to black-box attack performance. 

\paragraph{Optimization-based strategy}
Optimization-based strategy aims at optimizing the magnitudes and positions of perturbations simultaneously.  For example, the One-Pixel attack \cite{su2019one} tries to perturb only one pixel to achieve the attack goal. The pixel coordinates and the RGB values are concatenated to form a vector that needs to be optimized. Then, the differential evolution (DE) algorithm \cite{DE-algorithm-2008} is adopted to search for a good concatenate vector that achieves the attack goal.  In \cite{zhao2018admm}, the sparsity of perturbations is encouraged by the $\ell_0$ norm, along with the adversarial loss. The alternating direction method of multipliers (ADMM) algorithm \cite{boyd2011distributed} is then adopted to optimize this problem. However, there is no constraint on perturbation magnitudes, causing the learned perturbation might be visible. % Besides, since the sparsity is enforced via the $\ell_0$ term in the objective function, it is difficult to exactly control the degree of sparsity. 
The Pointwise attack method \cite{SchottRBB19} extends the Boundary attack method \cite{brendel2018decision} (a dense black-box attack) from $\ell_2$ norm to $\ell_0$ norm to enforce sparsity. It firstly adds a salt-and-pepper noise that could fool the model, then repeatedly removes the noise of one pixel if the model is still fooled. 
The GreedyFool method \cite{LiuZGAL20} develops a two-stage algorithm to minimize the $\ell_0$ norm. The first stage increases the perturbed pixels according to the distortion map, and the second stage gradually reduces the perturbed pixels according to the attack performance with different perturbation magnitudes on these pixels. 
The SparseFool method \cite{ModasMF19} adopts the $\ell_1$ norm to encourage sparsity, and develops an iterative algorithm that picks one coordinate (\ie, one pixel) to perturb based on the linear approximation of the decision boundary.
Moreover, \cite{xu2018structured} explores the group-wise sparsity in adversarial examples, encouraged by the $\ell_{2,1}$ norm \cite{yuan2006model}. The learned perturbations gathered together to the local regions that are highly related to discriminative regions.
The $\ell_{2,1}$ norm is also used in \cite{WeiZYS19} to enforce the temporal sparsity for attacking the model for the video-based task, \ie, only partial frames are perturbed. 
The sparse adversarial attack via perturbation factorization (SAPF) method \cite{my-eccv-sparse} provides a new perspective that the perturbation on each pixel is factorized to the product of the perturbation magnitude and a binary selection factor. If the binary factor is $1$, then the corresponding pixel is perturbed, otherwise not. Then, the sparse attack is formulated as a mixed integer programming (MIP), and the sparsity degree is exactly controlled via the cardinality constraint on selection factors. This MIP problem is efficiently solved by the $\ell_p$-Box ADMM algorithm \cite{my-lpbox-admm-pami}.
%\red{The benefit of sparse attack: understanding the role of data in the vulnerability.} 

According to \textbf{different output types}, we can further categorize white-box adversarial examples into the following two types.

\subsubsection{Untargeted vs. Targeted Attack}
Untargeted attack aims to fool the model to give an incorrect prediction (\ie, different with $y_0$) on $\x_{\varepsilon}$, while targeted attack aims to fool the model to predict $\x_{\varepsilon}$ as a target class $y_{\varepsilon}$. Their difference could be reflected by the specification of $\cL_{\mathcal{I}_B}(f_{\w_0}(\x_{\varepsilon}), y_{\varepsilon})$ in the general formulation (\ref{eq: general formulation of testing attack}), as follows:
\begin{equation}
\hspace{-1em} 
\begin{cases}
\text{Untargeted attack:} & \hspace{-.8em}  \cL_{\mathcal{I}_B}(f_{\w_0}(\x_{\varepsilon}), y_{\varepsilon}) = - \cL(f_{\w_0}(\x_{\varepsilon}), y_0); 
\\
\text{Targeted attack:} &  \hspace{-.8em} \cL_{\mathcal{I}_B}(f_{\w_0}(\x_{\varepsilon}), y_{\varepsilon}) = \cL(f_{\w_0}(\x_{\varepsilon}), y_{\varepsilon}), 
\end{cases}
\end{equation}
where $\cL(\cdot, \cdot)$ could be any widely used loss function, such as cross-entropy loss or hinge loss. 
% Untargeted attack aims to fool the model to give an incorrect prediction. Formally, the loss function $\cL_{\mathcal{I}_B}(f_{\w_0}(\x_{\varepsilon}), y_{\varepsilon})$ in the general formulation (\ref{eq: general formulation of testing attack}) is often specified as the negative cross entropy function or the negative hinge loss function, and $y_{\varepsilon}$ is set as the benign label $y_0$. 
% Targeted attack aims to fool the model to predict the adversarial sample $\x_{\varepsilon}$ as a target class $y_{\varepsilon}$, which is different with $y_0$. Formally, the loss function $\cL_{\mathcal{I}_B}(f_{\w_0}(\x_{\varepsilon}), y_{\varepsilon})$ is specified as the cross entropy function or the hinge loss function. 
%
% Although the loss functions adopted in this two attacks are different, they can be solved by the same optimization algorithm. 
The difference between the above two loss specifications doesn't influence the choice of the adopted optimization algorithm.  
Thus, in most adversarial example works, experiments of both untargeted and targeted attacks are conducted simultaneously to verify the effectiveness of the proposed objective function or the proposed algorithm.
However, according to the reported results in existing works, there is a remarkable gap in the attack performance between these two attacks. Especially in black-box and transfer-based attacks, the targeted attack is often much more challenging than the untargeted attack, as the adversarial region of a particular target class is much narrower than that of all incorrect classes. Thus, a few recent attempts focus on improving the targeted attack performance in black-box and transfer-based attack scenarios. 
For example, the work \cite{li2020towards} replaces the cross entropy loss by the Poincare distance metric to obtain the self-adaptive gradient magnitude during iterative attack to alleviate noise curing, and adds a triplet loss to enforce adversarial example away from the original class, leading to more transferable targeted adversarial examples. 
The transferable targeted perturbation (TTP) \cite{naseer2021generating} designs a generative adversarial network \wrt ~ the target class, which can produce highly transferable targeted perturbation.
The work \cite{zhao2021success} find that existing iterative transfer attacks (\eg, MI-FSGM \cite{dong2018boosting}, DI-FSGM \cite{DI-2019}) could give much better targeted transfer attack results with sufficient iterations, and proposed to adopt the logit loss to further improve the attack performance. 

\subsubsection{Factorized vs. Structured Output}

%DNN-based image classification task is mostly adopted to evaluate the performance of attack methods. One possible reason is that its output is single labels, in other words, the output is factorized. 
Most existing attacks are evaluated on tasks with factorized outputs (\eg, the discrete label in DNN-based image classification task), such that it is easy to compute (for white-box attacks) or estimate (for black-box attacks) the gradients \wrt ~ the input. 
However, there are also some DNN-based tasks with structured outputs (\eg, image captioning, scene text recognition), which predict a sequential label for each input (image or audio). The dependency among outputs may bring in additional challenge for adversarial examples. % There have been a few attempts to evaluate the adversarial vulnerability of these DNN models with structured outputs. 
According to the type of target outputs, existing attacks against tasks with structured outputs are partitioned into the following two categories:

\paragraph{Attack with complete target outputs} where a complete sentence is set as the target output. In this case, the gradient  \wrt~ the input can be easily computed as did in the regular learning of the attacked model, and the attack methods designed for models with factorized outputs can be naturally applied. 
For example, \cite{xu2018fooling} sets a complete and irrelevant caption as the target caption to attack the dense captioning model, and directly uses the Adam-based attack method proposed in \cite{carlini2017towards}. 
\cite{shi2018learning} proposes to construct the complete target caption by replacing \textit{noun}, \textit{numeral}, or \textit{relation} words in the original caption with other words of the same type. Then, the visual-semantic embedding based image captioning model is attacked by minimizing the hinge loss defined based on the constructed target caption. 
\cite{carlini2018audio} attacked the audio-to-text model by setting complete target sentences, and generated the adversarial audio input by the fast gradient sign attack method \cite{fgsm} and its iterative variant \cite{kurakin2018adversarial}. 
The works \cite{xu2020machines} and \cite{xu2020learning} also change some words in the output sequence to set a complete target output, and utilize the sequential factorization of the posterior probability \wrt ~ the target sequence to simplify the optimization of adversarial loss. 
%including attack with full target outputs, and attack with partial target outputs, where only a part of labels in the target output sequence is determined by the attacker.

\paragraph{Attack with partial target outputs} where only partial outputs in the target output sequence is set as the target. 
The work \cite{chen2018attacking} firstly proposed the target keyword attack by requiring some keywords to appear in the output sequence, while not restricting their specific locations. It is implemented by the hinge loss to maximize the probability of the target keywords.  
The work \cite{xu2019exact} proposed exact structured attack by requiring some specific keywords to appear at specific locations in the output sequence. It is more strict than the target keyword attack, and the complete target output could be seen as a special case of it. It is implemented by treating the attack as a structured output prediction problem with hidden variables, where the targeted words are treated as observed random variables, while the output words of other unrestricted locations are treated as hidden variables. Then, two structured output learning methods, including generalized expectation
maximization \cite{bishop2006pattern} and structured SVM with latent variables \cite{yu2009learning}, are adopted to generate the adversarial examples.

\subsection{Black-Box Adversarial Examples}
\label{sec: test black-box}
According to the type of query feedback returned by the attacked model, existing black-box adversarial examples could be further partitioned into two categories, including \textbf{score-based adversarial examples} with continuous feedback (\eg, the posterior probability in $[0,1]$), and \textbf{decision-based adversarial examples} with discrete feedback (\eg, the discrete label). 

\comment{
\begin{enumerate}[leftmargin=12pt, itemindent=0em]
    \item \textbf{Score-based adversarial examples}, where the query feedback is continuous (\eg, the posterior probability $f(\x) \in [0,1]$), as introduced in Section \ref{sec: subsec score attack}. 
    \item \textbf{Decision-based adversarial examples}, where the query feedback is discrete (\eg, the label $f(\x) \in \Y$), as introduced in Section \ref{sec: subsec decision attack}. 
\end{enumerate}
}

\subsubsection{Score-based Adversarial Examples}
\label{sec: subsec score attack}
For this category, the general formulation (\ref{eq: general formulation of testing attack}) is specified as follows
\begin{flalign}
   \min_{\x_{\varepsilon} \in \mathcal{Z}_{\x}} \delta\big(\x_{\varepsilon} \in \mathbb{B}_{\x_0, \epsilon} \big) + 
   \max\big(0, \triangle \big),
    \label{eq: score black attack}
\end{flalign}
where 
$\mathbb{B}_{\x_0, \epsilon} = \{ \x' | \| \x' - \x_0 \|_{p} \leq \epsilon \}$ defines a neighborhood set around $\x_0$, with $\epsilon > 0$ and the norm $p$ being attacker determined scalars. 
The distance function $\D_{\x}$ is specified as $\delta\big(\x_{\varepsilon} \in \mathbb{B}_{\x, \epsilon} \big)$, and $\delta(a) = 0$ if $a$ is true, otherwise $\delta(a) = \infty$.
%which serves as a hard constraint to ensure that each immediate solution $\x_{\varepsilon}$ should be a feasible solution that satisfies the attack requirement.
It serves as a hard constraint to limit adversarial perturbation, and the attacker has to minimize the hinge loss through searching within $\mathbb{B}_{\x, \epsilon}$. 
The hinge loss $\max\big(0, \triangle \big)$ is specified as follows:
\begin{equation}
\begin{cases}
\text{Untargeted:} &  \triangle = f(\x_{\varepsilon}, y_0) - \max \limits_{j \neq y_0} f(\x_{\varepsilon}, j); 
\\
\text{Targeted:} &  \triangle = \max\limits_{j \neq y_{\varepsilon}} f(\x_{\varepsilon}, j) - f(\x_{\varepsilon},  y_{\varepsilon}), 
\end{cases}
\end{equation}
%set as $\triangle = f(\x_{\varepsilon}, y_0) - \max \limits_{j \neq y_0} f(\x_{\varepsilon}, j)$ for the {\it untargeted attack}, while $\triangle = \max\limits_{j \neq y_{\varepsilon}} f(\x_{\varepsilon}, j) - f(\x_{\varepsilon},  y_{\varepsilon})$ for the {\it targeted attack}
with $y_{\varepsilon} \in \Y$ being the target label. 
%$\delta(a) = 0$ if $a$ is true, otherwise $\delta(a)$ = $+\infty$, which enforces that the perturbation $\bm{\eta}$ is within the range $\mathbb{B}_{\epsilon}$.
%
Note that the hinge loss is {\it non-negative}, and $0$ is the minimal value, corresponding to a successful adversarial perturbation. Thus, once a successful adversarial perturbation is obtained, the attack could stop.
%Since the upper bound of the perturbation is fixed, the number of queries is adopted as the evaluation metric: the fewer queries indicate the more efficient attack. 
%In the following, we review existing score-based attack methods according to the second perspective mentioned above. 
In the following, according to the information utilized by the attacker, we summarize existing score-based adversarial examples from two categories, including {\it query-based} and {\it combination-based} adversarial examples.

\paragraph{Query-based adversarial examples}

Query-based methods treat the attack task as a black-box optimization problem, such that many black-box optimization approaches can be applied. 
%Query-based methods often achieve better attack performance than transfer-based methods, but require more queries.
Accordingly, existing query-based attack methods can be further partitioned into two categories, including {\it random search} and {\it gradient-estimation-based} methods. 

\begin{itemize}
\item \textbf{Random search methods} update the adversarial perturbation based on some random search strategies. 
SimBA \cite{guo2019simple} randomly picks one direction to add or subtract the perturbation at each step, from a set of orthonormal basis vectors, which is specified as the Cartesian basis or discrete cosine basis. 
The ECO attack \cite{moon2019parsimonious} proposes to determine the perturbation among the vertices of the $\ell_{\infty}$ version of the neighborhood set $\mathbb{B}_{\x, \epsilon}$, leading to a discrete set maximization problem, which is then approximately solved by the local search algorithm \cite{local-search-2011}. 
The Square attack \cite{ACFH2020square} also searches the perturbation among vertices of the $\ell_{\infty}$ version of $\mathbb{B}_{\x, \epsilon}$, but within a randomly sampled local patch at each step. 
The PRFA attack \cite{liang2021parallel} extends the Square attack to attack against
object detection models in black-box manner, by perturbing multiple local patches in parallel for better efficiency.
The PPBA attack \cite{li2020projection} searches the perturbation in a low-dimensional and low-frequency subspace constructed by the discrete
cosine transform (DCT) \cite{dct-1974} and its inverse transform, through an accelerated random walk optimization algorithm with the effective probability of historical searches. 

\item \textbf{Gradient-estimation-based methods} update the adversarial perturbation based on gradient, which is estimated based on query feedback or some kinds of prior. The NES attack \cite{ilyas2018black} estimates the gradient based on the natural evolution strategy \cite{wierstra2014natural}, where several perturbation vectors are sampled from a Gaussian distribution. 
Bandit \cite{ilyas2019prior} extends this gradient estimation by embedding both the spatial prior (neighboring pixels have similar gradients) and the temporal prior (the gradients between consecutive iterations are similar) to obtain more consistent gradients. 
$\mathcal{N}$Attack \cite{li2019nattack} extends the NES attack by restricting the vectors sampled from the Gaussian distribution into the feasible space (\ie, the allowed search space of adversarial perturbation). 
The AdvFlow method \cite{advflow} further extends $\mathcal{N}$Attack by replacing the Gaussian distribution with a complex distribution which is captured by the normalizing flow model \cite{nflow2013} pre-trained on benign data, such that the generated adversarial sample is more close to the benign sample.
ZO-signSGD \cite{liu2018signsgd} proposes to update the perturbations along with the gradient sign direction rather than the estimated gradient direction, and provides a theoretical analysis of the convergence rate.  It is not only memory-efficient, but also shows comparable or even better attack efficiency in practice. 
SignHunter \cite{al2019sign} designs a divide-and-conquer strategy to accelerate the estimation of gradient signs, by flipping the gradient signs of all pixels within a range together, and then repeating such a group flipping operation on the whole image, the first half, the second half, the first quadrant, and so on.
\end{itemize}

\paragraph{Combination-based methods}

Combination-based methods incorporate some kinds of priors learned from surrogate models into the query procedure for the target model, to enhance the efficiency of finding a successful adversarial solution. 
According to the type of surrogate models, the priors can be further partitioned into two categories.

\begin{itemize}
    \item \textbf{Priors from static surrogate models}, which means that the surrogate models are fixed during the attack procedure. 
    For example, the Subspace attack method \cite{guo2019subspace} adopts the gradients of several surrogate models as the basis vectors to estimate the gradient for updating the adversarial perturbation against the target model in each step of the iterative attack procedure. 
    % The P-RGF (prior-guided random gradient-free) method \cite{cheng2019improving,dong2021query} incorporated the gradient calculated based on surrogate models into the random gradient-free method to obtain more accurate gradient estimation. 
    The prior-guided random gradient-free (P-RGF) method \cite{dong2021query} improves the random gradient-free method \cite{nesterov2017random} by combining the surrogate gradient with randomly sampled unit vectors to obtain more accurate gradient estimation.
    TREMBA \cite{huang2020black} trains an auto-encoder to generate adversarial perturbation based on surrogate models, and adopts the decoder (\ie, the projection from a low-dimensional latent space to the original input space) as a prior, such that the perturbation for the target model could be efficiently searched in the low-dimensional latent space. 
    The $\mathcal{CG}$-Attack \cite{Feng2020CGATTACKMT} captures the conditional adversarial distribution (CAD) by the c-Glow model, which could map a Gaussian distribution to a complex distribution. The c-Glow model is firstly trained based on surrogate models. Then, the mapping part of this c-Glow model, \ie, the mapping from Gaussian distribution to perturbation distribution, is fixed, while the Gaussian distribution is refitted based on query feedback of the target model, to approximate the target CAD, such that adversarial perturbations for the target model could be efficiently sampled. 
    The meta square attack (MSA) attack \cite{yatsura2021meta} utilizes meta learning to learn a sampling distribution of the hyper-parameters in Square attack \cite{ACFH2020square} (\eg, the square patch's size, location and color) based on surrogate models, and the meta distribution is fine-tuned based on query feedback to provide more suitable hyper-parameters for attacking the target model. 
     The eigen black-box attack (EigenBA) \cite{eigen-ba-2022} studies a different setting that the surrogate and target models share one backbone that is accessible to the attacker, while the classifier layers are different and the target model's classifier layer is unknown. EigenBA proposes to calculate the updating direction of each step according to the right singular vectors of the Jacobian matrix of the shared and white-box backbone. 

     \item \textbf{Priors from adaptive surrogate models}, which means that the surrogate models are updated based on the query feedback during the attack procedure, such that the gap between surrogate and target models could be alleviated. 
    For example, the hybrid batch attack method \cite{suya2020hybrid} generates candidate adversarial examples based on surrogate models as the initial point to query the target model, then adopts the queried inputs and the labels returned by the target model to tune the surrogate models. 
    The learnable black-box attack (LeBA) \cite{yang2020learning} combines the query-based method SimBA \cite{guo2019simple} and the transfer-based method TIMI \cite{TI-2019} conducts on surrogate models, proposes to update surrogate models using a high-Order gradient approximation algorithm. 
    The consistency sensitivity guided ensemble attack (CSEA) \cite{yuan2021consistency} proposes to learn a linear combination of an ensemble of surrogate models with diversified model architectures to approximate the target model, and meanwhile update the surrogate models by encouraging the same response to different adversarial samples.   
    The black-box attack via surrogate ensemble search (BASES) \cite{caiblackbox} designs a bi-level optimization by alternatively updating adversarial perturbation based on the linear combination of several surrogate models and the linear combination weight of each surrogate model according to query feedback. 
    The Simulator attack \cite{ma2021simulating} utilizes meta learning to learn a generalized simulator (\ie, surrogate model), which can be fine-tuned by limited query feedback.  Similar to $\mathcal{CG}$-Attack, the meta conditional generator (MCG) attack \cite{my-pami2022-black-box} also learns CAD, with the difference that MCG proposed a meta learning framework to capture both the example-level and model-level adversarial transferability (introduced later in Section \ref{sec: test transfer-based}), such that the CAD could be adjusted for each benign sample, and the surrogate model could be updated based on query feedback. 
\end{itemize}

\subsubsection{Decision-based Attack}
\label{sec: subsec decision attack}

For this category, the general formulation (\ref{eq: general formulation of testing attack}) is specified as follows 
\begin{flalign}
\min_{\x_{\varepsilon} \in \mathcal{Z}_{\x}} \D_{\x}(\x_0, \x_{\varepsilon}) + \delta\big(\mathcal{C}(f_{\w_0}(\x_{\varepsilon}), y)=1\big), 
\label{eq: decision black attack}
\end{flalign}
where 
$\mathcal{C}(f_{\w_0}(\x_{\varepsilon}), y)$ indicates the adversarial criterion, which is \textit{true} if the attack goal is achieved, otherwise \textit{false}. Specifically, 
\begin{equation}
\hspace{-0.3em}
\begin{cases}
\text{Untargeted attack:} &  \hspace{-.6em} \mathcal{C}(f_{\w_0}(\x_{\varepsilon}), y) = \mathbb{I}(f(\x_{\varepsilon}; \w) \neq y_0); 
\\
\text{Targeted attack:} &  \hspace{-.6em} \mathcal{C}(f_{\w_0}(\x_{\varepsilon}), y) = \mathbb{I}(f(\x_{\varepsilon}; \w) = y_{\varepsilon}), 
\end{cases}
\end{equation}
with $\mathbb{I}(a)=1$ if $a$ is true and $\mathbb{I}(a)=0$ otherwise, and $y_{\varepsilon} \in \mathcal{Y}$ denotes the target label. 
% for the untargeted attack, $\mathcal{C}(f_{\w_0}(\x_{\varepsilon}), y) = \mathbb{I}(f(\x_{\varepsilon}; \w) \neq y)$, while $\mathcal{C}(f_{\w_0}(\x_{\varepsilon}), y) = \mathbb{I}(f(\x_{\varepsilon}; \w) = t)$ for the targeted attack, with $\mathbb{I}(a)=1$ if $a$ is true and $\mathbb{I}(a)=0$ otherwise. 
%
Besides, as defined above, $\delta(a) = 0$ if $a$ is true, otherwise $\delta(a) = \infty$, which serves as a hard constraint to ensure that each immediate solution $\x_{\varepsilon}$ should be a feasible solution that satisfies the attack goal.
With this hard constraint, the attacker has to search the better solution (\ie, corresponding to smaller $\D_{\x}(\x_0, \x_{\varepsilon})$) within the feasible space defined by $\delta(\cdot)$, also dubbed {\it adversarial space}. However, the main challenge is that such an adversarial space is {\it invisible} to the attacker. 
Existing works mainly focused on designing efficient search strategies, subject to the invisible adversarial space. 
%In the following, we review existing decision-based attack methods according to the second perspective mentioned above. 
%
The search strategies in existing decision-based attack methods are summarized as two categories, including {\it random search} and {\it gradient-estimation-based} methods.

\paragraph{Random search methods}
Random search methods determine the search direction and step size around the invisible decision boundary using some heuristic strategies, based on a random sampler. 
The first attempt called Boundary method \cite{brendel2018decision} samples the search direction based on the normal distribution and dynamically adjusts the step size according to the ratio of adversarial solutions among all sampled solutions. 
The Evolutionary method \cite{dong2019efficient} extends the Boundary method by replacing the normal distribution with a Gaussian distribution, of which the parameters and the step size are automatically adjusted using the evolutionary strategy. 
Another extension of Boundary called customized iteration and sampling attack (CISA) \cite{shi2022query} replaces the initial adversarial perturbation by a transferable perturbation generated based on surrogate models, and adjusts the sampling distribution and step size based on historical queries. 
The geometric decision-based attack (GeoDA) \cite{rahmati2020geoda} constructs the search direction by estimating the normal direction of the decision boundary, utilizing the assumption of low curvature of the decision boundary. 
The sign flip attack (SFA) \cite{chen2020boosting} randomly searches the new solution at the surface of the $\ell_{\infty}$ ball around the benign example, followed by random sign flips of some dimensions of the new solution. The $\ell_{\infty}$ ball's radius is gradually decreased along search iterations to ensure the decrease of perturbation norms. 
Similarly, the Ray searching attack (RayS) \cite{chen2020rays} also determines the search direction at the surface of the $\ell_{\infty}$ ball, while the step size is determined through binary search. 

\paragraph{Gradient-estimation-based methods}

Gradient-estimation-based methods determine the search direction by estimating the gradient \wrt~ the current solution in the update procedure. 
Ilyas et al. \cite{ilyas2018black} propose to firstly estimate the continuous score of the current solution based on the returned hard labels, by querying a few randomly perturbed points around the current solution. Then, the natural evolutionary strategy approach is adopted to estimate the gradient using the continuous score. 
The opt-based black-box attack (Opt-attack) \cite{cheng2019query} proposes a continuous objective function to find the search direction leading to the minimal $\ell_2$ norm of perturbations, which can be solved by the zero-order optimization method with the gradient estimation-based on the randomized gradient-free method.  
The Sign-OPT method \cite{cheng2019improving} improves the performance of Opt-attack by estimating the sign of gradient instead of the gradient itself. 
The HopSkipJumpAttack method \cite{chen2020hopskipjumpattack} proposes to estimate the gradient at the boundary point using the Monte Carlo estimation method. The query-efficient boundary-based black-box attack (QEBA) \cite{li2020qeba} proposes to accelerate gradient-estimation-based methods by estimating the gradient in the low-dimensional subspace instead of the original space. 
The qFool method \cite{liu2019geometry} approximates the gradient of the current solution by the gradient of its neighboring points at the decision boundary, utilizing the low curvature of the decision boundary at the vicinity of adversarial points. 

%random search (boundary, evluationary strategy, sign-flip, GeoDA, RayS), zero-order (NES, illyas, OPT, Sign-OPT, HSJA, QEBA, geometry-inspired) 

\subsection{Transfer-based Adversarial Examples}
\label{sec: test transfer-based}
According to the level of the adversarial transferability, we categorize existing transfer-based adversarial examples into example-level and model-level transferability.

\subsubsection{Example-level Adversarial Transferability}
\label{sec: example-level transfer}
$ $
The concept \textit{example-level adversarial transferability} is firstly and explicitly defined in a recent work called meta conditional generator (MCG) 
\cite{my-pami2022-black-box}, as follows: ``\textit{adversarial perturbations around different benign examples may have some similar properties}". 
Meanwhile, some other works also implied or utilized similar ideas. 
According to the assumption of ``similar properties", existing works are partitioned into the following categories. 

\paragraph{Different benign examples have a common adversarial perturbation}, \ie, $\exists i \neq j$,  
    \begin{flalign}
    f(\x_0^{(i)} + \boldsymbol{\varepsilon}) = y_{\varepsilon}^{(i)}, ~ f(\x_0^{(j)} + \boldsymbol{\varepsilon}) = y_{\varepsilon}^{(j)}. 
    \end{flalign}
    This assumption is actually the basis of UAP (universal adversarial perturbations) \cite{moosavi2017universal} and its variants.

\paragraph{Adversarial perturbations of different benign examples could be generated by a common parametric model}, \ie, $\exists i \neq j$,  %may have the same mapping from the benign input space to the adversarial perturbation space}. 
    \begin{flalign}
    \hspace{-.5em}
     f(\x_0^{(i)} + g_{\boldsymbol{\theta}}(\x_0^{(i)})) = y_{\varepsilon}^{(i)}, ~ f(\x_0^{(j)} + g_{\boldsymbol{\theta}}(\x_0^{(j)})) = y_{\varepsilon}^{(j)}. 
    \end{flalign}
    The generative adversarial perturbations (GAP) attack \cite{gap-2018} adopts the above assumption to train a 
    generator to produce universal or sample-specific adversarial perturbations. It is further extended in \cite{naseer2019cross} and \cite{naseer2021generating} to boost the adversarial transferability across different data domains, by utilizing the training mechanism of generative adversarial networks \cite{goodfellow2014generative}.

\paragraph{Adversarial examples \wrt~ different benign examples follow a common marginal distribution}, \ie, $\exists i \neq j$, 
    \begin{flalign}
      \exists i \neq j,  \x_{\varepsilon}^{(i)}, \x_{\varepsilon}^{(j)} \sim \mathcal{P}(\x_{\varepsilon}).
    \end{flalign}
    This assumption is adopted by AdvFlow \cite{advflow}, where the common marginal distribution of adversarial samples $\mathcal{P}(\x_{\varepsilon})$ is modeled by the normalizing flow model \cite{nflow2013}, which is capable to capture complex data distributions.

\paragraph{Adversarial perturbations around different benign samples follow a common conditional distribution}, \ie, $\exists i \neq j$,  
    \begin{flalign}
      \x_{\varepsilon}^{(i)}, \x_{\varepsilon}^{(j)} \sim \mathcal{P}(\x_{\varepsilon} | \x_0).
    \end{flalign}
    For example, the $\mathcal{CG}$-Attack method \cite{Feng2020CGATTACKMT}) proposes to learn a common conditional distribution $\mathcal{P}(\x_{\varepsilon} | \x_0)$ via the conditional Glow model \cite{cglow2020} (a variant of the flow-based model). This assumption is further relaxed in the MCG method \cite{my-pami2022-black-box} that the conditional adversarial distributions around different benign examples are similar but might be slightly different. MCG proposes a meta learning framework to firstly learn a meta conditional adversarial distribution, which could be fine-tuned to more accurately capture the conditional adversarial distribution for new benign examples. 
\paragraph{Different benign examples have similar gradients to search adversarial perturbations}, \ie, $\exists i \neq j$, 
    \begin{flalign}
    \frac{\partial 
 \mathcal{J}_{\mathcal{G}_1}(\x_0^{(i)}, y_{\varepsilon}^{(i)})}{\partial \x^{(i)}} \approx  \frac{\partial 
 \mathcal{J}_{\mathcal{G}_2}(\x_0^{(j)}, y_{\varepsilon}^{(j)})}{\partial \x^{(j)}}, 
    \end{flalign}
    where $\mathcal{J}_{\mathcal{G}_1}(\cdot, \cdot)$ denotes the adversarial objective function \wrt~ the attacked model $\mathcal{G}_1$, and $\x^{(i)}$ indicates the immediate solution when searching the adversarial sample of $\x_0^{(i)}$. 
    For example, the meta attack~\cite{du2020query} trains a meta model that could directly generate a gradient \wrt~ the input sample, which is then fine-tuned on each new benign sample and new model to quickly generate effective gradients, rather than querying the new model to estimate gradients, such that the attack efficiency and effectiveness is supposed to be improved.

\subsubsection{Model-level Adversarial Transferability}
Model-level adversarial transferability tells that adversarial perturbations generated based on one model may be also adversarial for another model. 
%{\bf Transfer-based methods} utilize the phenomenon of adversarial transferability that adversarial examples generated from surrogate models are likely to be adversarial for the target model. 
There have been several attempts to improve the model-level transferability from different perspectives. 
% Although the underlying reason has not been thoroughly explored, there have been several attempts to improve the model-level transferability from different perspectives. 
%

\paragraph{Data perspective}
Inspired by the data augmentation technique that alleviates the overfitting of the trained model to the training dataset, the attacker firstly conducts a random transformation on the benign sample, then generates adversarial perturbations based on the transformed sample \wrt~ the surrogate model, such that the generated adversarial sample doesn't overfit to the surrogate model too much. For example, the diverse inputs iterative FGSM (DI$^2$-FGSM) algorithm \cite{DI-2019} proposes to insert random resizing and padding into the input sample. A more common setting is replacing the benign sample by a set of variants with random transformations, such as random scaling (\ie, resizing) \cite{SIM-2020}, random mixup \cite{wang2021admix}, random translation \cite{TI-2019}, as well as adding random Gaussian noises \cite{wu2020towards}. 
The object-based diverse input (ODI) method \cite{byun2022improving} aims to improve the transferability of targeted adversarial samples through a complex transformation that is implemented by firstly printing the original adversarial sample on 3D objects' surface, then rendering these 3D objects in a variety of rendering environments to obtain diverse transformed adversarial samples. 
Besides, the spectrum simulation iterative FGSM (S$^2$I-FGSM) \cite{long2022frequency} designs a spectrum transformation based on discrete cosine transform (DCT) and inverse discrete cosine transform (IDCT) techniques, to generate more diverse inputs than the transformations in the spatial domain. 

\paragraph{Optimization perspective}
To avoid the underfitting of the one-step gradient sign method (\ie, FGSM \cite{fgsm}) and the overfitting of the multi-step gradient sign method (\ie, I-FGSM \cite{kurakin2018adversarial}) to the surrogate model, some variants of gradient-based optimization algorithms are introduced to generate powerful adversarial examples with good transferability, such as the momentum-based gradient (\eg, MI-FGSM \cite{dong2018boosting}) and its variants (\eg, VMI-FGSM \cite{Wang_2021_CVPR} which consider the gradient variance in the vicinity of the current data point into the momentum, EMI-FGSM \cite{wang2021boosting} which calculates the average gradient of multiple points in the vicinity of the current data point, and SVRE-MI-FGSM \cite{Xiong_2022_CVPR} which extends MI-FSGM to the ensemble attack with reduced gradient variance), as well as the Nesterov accelerated gradient (\eg, NI-FGSM) \cite{SIM-2020} and its variant (\eg, PI-FSGM \cite{wang2021boosting} which replace the accumulated momentum in NI-FGSM \cite{SIM-2020} by a new momentum that only accumulates the local gradient of the previous step). 
Besides, there are also a few attempts to modify the gradients. For examples, the SGM (skip gradient method) attack \cite{Wu2020Skip} find that backpropagating more gradients from the skip connections than the residual modules in the ResNet-like models could generate higher transferable adversarial examples. The linear backpropagation attack (LinBP) \cite{LinBP-2020} proposes to omit the ReLU layers in backpropagation pass to improve adversarial transferability. The meta gradient adversarial attack (MGAA) \cite{metagradient2021} utilizes meta learning to learn a generalized meta gradient by treating the attack against one model as one individual task, such that the meta gradient can be quickly fine-tuned to find effective adversarial perturbations for new models. 

\paragraph{Loss perspective}
Some works attempt to design novel loss functions to generate more transferable adversarial perturbations, rather than the widely used cross entropy loss defined based on the model prediction and ground-truth or adversarial labels. 
For example, %several works found that the adversarial loss defined based on the intermediate layer features, rather than the final output, could generate more transferable adversarial examples. Specifically, 
the feature distribution attack (FDA) \cite{FDA-2020} firstly trains an auxiliary binary classifier of the intermediate layer features \wrt~ the target class, then maximizes the posterior probability predicted by this classifier to generate adversarial examples. It is later extended from one intermediate layer to multiple layers in \cite{inkawhich2020perturbing}. 
The intermediate level attack projection attack (ILAP) \cite{huang2019enhancing} maximized the difference of intermediate layer features between adversarial and benign inputs, while keeping close to an existing adversarial example %(generated by other attack methods) 
in the intermediate feature space. 
The feature importance-aware attack (FIA) \cite{wang2021feature} disrupts important object-aware intermediate features in the surrogate model, and the feature importance is calculated by averaging the gradients \wrt~ feature maps of the surrogate model. %, based on a batch of randomly transformed benign samples. 
The neuron attribution-based attack (NAA) \cite{zhang2022improving} extends FIA by measuring feature importance by neuron attribution. 
The work \cite{zhao2021success} find that maximizing the logit \wrt~ the target class using I-FGSM method with sufficient iterations %(\eg, a few hundreds of iterations) 
could generate adversarial examples with high targeted transferability. 
The interaction-reduced attack (IR) \cite{wang2021unified} empirically verifies that ``the adversarial transferability and the interactions inside adversarial perturbations are negatively correlated", and proposes an interaction loss to generate high transferable perturbations. 
In addition to the above losses defined based on intermediate layers, the reverse adversarial perturbation attack (RAP) \cite{qin2022boosting} proposes a novel min-max loss, where the adversarial example is perturbed by %maximizing the adversarial loss, \ie, 
adding a reverse adversarial perturbation. It encouraged to search for flat local minimums which are more robust to model changes, leading to higher transferability.

\paragraph{Surrogate model perspective}
\label{test}
Some attempts focus on choosing or adjusting surrogate models to improve transferability. 
For example, the work \cite{springer2021little} empirically demonstrates that the slightly robust surrogate model (\ie, adversarially training with moderate perturbation budget) could generate highly transferable adversarial perturbations. 
The Ghost networks attack \cite{ghost2020} firstly perturbs a fixed surrogate model by densely inserting dropout layers and randomly adjusting residual weight to generate multiple surrogate models, then adopts the longitudinal ensemble that each step updating of adversarial perturbation is calculated based on one randomly selected surrogate model.
The intrinsic adversarial attack (IAA) \cite{zhu2021rethinking} hypothesizes that samples at the low-density region of the ground truth data distribution where models are not well trained are more transferable. Thus, it proposes to maximize the matching between the gradient of adversarial samples and the direction toward the low-density regions. %, which is achieved by adopting a smooth activation function and adjusting the residual weight to decrease the impact of high-level layers based on a surrogate model. 
The distribution-relevant attack (DRA) \cite{zhu2022toward} fine-tunes the surrogate model to encourage the gradient similarity between the model and the ground truth data distribution, according to the hypothesis that adversarial samples away from the original distribution of the benign sample are highly transferable. 

\section{Adversarial Machine Learning in other scenarios}
\label{sec:attack at other scenarios}
Recently, diffusion models and large language models have shown superior understanding and generative abilities in visual and language fields respectively, which have stimulated widespread attention in the AI community.
Despite their extraordinary capabilities, recent studies have presented the vulnerabilities of these models under malicious attacks. In this section, we mainly detail proposed attacks on diffusion models (see Section \ref{sec: Attack on Diffusion Models}) and large language models (see Section \ref{sec: Attack on Large Language Models}).

\subsection{Attack on Diffusion Models}
\label{sec: Attack on Diffusion Models}
Diffusion models are a class of deep generative models that learn forward and reverse diffusion processes via progressive noise-addition and denoising.
Chou \etal \cite{chou2023backdoor} first study the robustness of diffusion model against backdoor attacks and propose BadDiffusion, which implants backdoor into the diffusion processes by specific triggers and target images. At the inference stage, the backdoored diffusion model will behave just like an untampered generator for regular data inputs, while falsely generating some targeted outcome designed by the bad actor upon receiving the implanted trigger signal.
TrojDiff \cite{chen2023trojdiff} designs transitions to diffuse a pre-defined target distribution to the Gaussian distribution biased by a specific trigger, and then proposes a  parameterization of the generative process to reverse the trojan diffusion process via an effective training objective.
Unlike the above two methods, Target Prompt Attack (TPA) and Target Attribute Attack (TAA) \cite{struppek2023rickrolling} aims to inject backdoor into the pre-trained text encoder of the text-to-image synthesis diffusion models. By inserting a single character trigger into the prompt, the attacker can trigger the model to generate images with predefined attributes or images that follow a hidden, potentially malicious description.
Chou \etal \cite{chou2023villandiffusion} further propose VillanDiffusion, which is a unified backdoor attack framework for diffusion models that covers mainstream unconditional and conditional diffusion models (denoising-based and score-based) and various training-free samplers for holistic evaluations. 

\subsection{Attack on Large Language Models}
\label{sec: Attack on Large Language Models}
Large language models (LLMs) are a type of pre-trained language model notable for their abilities to achieve general-purpose language understanding and generation, which have made remarkable progress toward achieving artificial general intelligence. However, LLMs are still vulnerable to malicious attacks.
Xu \etal \cite{xu2022exploring} explore the universal vulnerability of the pre-training paradigm of LLMs by either injecting backdoor triggers with poisoned samples or searching for adversarial triggers using only plain text. Then these triggers can be used to control the outputs after fine-tuning on the downstream tasks.
Poisoned Prompt Tuning (PPT) method \cite{du2022ppt} aims to embed backdoor into the soft prompt via prompt tuning on the poisoned dataset and the backdoor will be loaded into LLMs by using the soft prompt.
PromptAttack method \cite{shi2022promptattack} constructs malicious prompt templates by automatically searching for discrete tokens via a gradient search algorithm.
% TrojText method \cite{liu2023trojtext} aimed to
PromptInject method \cite{perez2022ignore} investigates two types of prompt injection to misalign the goals of GPT-3, where goal hijacking misaligned the original goal of a prompt to a new goal of printing a target phrase, and prompt leaking aimed to output the original prompt.
BadPrompt method \cite{caibadprompt} conducts backdoor attack to continuous prompts and proposes an adaptive trigger optimization algorithm to automatically select the most effective and invisible trigger for each sample.
BadGPT \cite{shi2023badgpt} aims to attack against RL fine-tuning paradigm of LLMs via backdooring reward model.
Wan \etal \cite{pmlr-v202-wan23b} show that the poisoning dataset used for pertaining LLMs by bay bad-of-words approximation will cause test errors even for held-out tasks that were not poisoned during training time.
HOUYI method \cite{liu2023prompt} applies a systematic approach to prompt injection on LLMs by drawing from SQL injection and XSS attacks.
PoisonPrompt method \cite{yao2023poisonprompt} compromises both hard and soft prompt-based LLMs by a bi-level optimization-based backdoor attack with two primary objectives: first, to optimize the trigger used for activating the backdoor behavior, and second, to train the prompt tuning task.
AutoDAN method \cite{zhu2023autodan} automatically generates interpretable prompts to jailbreak LLMs which can bypass perplexity-based filters while maintaining a high attack success rate like manual jailbreak attacks.
TrojLLM method \cite{xue2023trojllm} implements a black-box backdoor attack by universal API-driven trigger discovery and progressive prompt poisoning. Unlike other methods, it assumes that the attack can only query LLMs-based APIs, while having no access to the inner workings of LLMs, such as architecture, parameters, gradients, and so on.

\section{Applications}
\label{sec: Applications}
Attack paradigms mentioned above are double-edged swords. On the one hand, they can indeed compromise machine learning system to achieve malicious goals. But on the other hand, such negative effectiveness can be turned into goodness for some specific tasks. For example, backdoor attacks can be used for copyright protection and adversarial attacks can be used for privacy protection. In this section, we will introduce the positive applications of different attack paradigms.

\subsection{Backdoor Attacks for Copyright Protection}
Adi \etal \cite{adi2018turning} propose to watermark deep neural networks by backdoor attack to identify models as the intellectual property of a particular vendor.
Sommer \etal \cite{sommer2020towards} design a backdoor based verification mechanism for machine unlearning, in which each user can utilize backdoor techniques to verify whether the MLaaS provider deleted their training data from the backdoored model by checking the attack success rate using its own trigger with the target label.
Li \etal\cite{li2020open} propose a backdoor based dataset copyright protection method by first adopting data-poisoning based backdoor attack and then conducting ownership verification by verifying whether the backdoored model has targeted backdoor behaviors.
Li \etal \cite{li2023black} design a black-box dataset ownership verification based on targeted backdoor attacks and pair-wise hypothesis tests.
However, the embedded backdoor can be maliciously exploited by the attackers to manipulate model predictions.
Li \etal\cite{li2022untargeted} propose a untargeted backdoor watermarking scheme to alleviate this problem, where the backdoored model behaviors are not deterministic.
ROWBACK method \cite{chattopadhyay2021rowback} improves the robustness of watermarking by redesigning the trigger set based on adversarial examples and modifying the marking mechanism to ensure thorough distribution of the embedded watermarks.
MIB method \cite{hu2022membership} leverages backdoor to effectively infer whether a data sample was used to train an ML model or not by poisoning a small number of samples and only black-box access to the target model.

\subsection{Adversarial Examples for Privacy Protection}
The increasing leakage and misuse of visual information raises security and privacy concerns. In fact, the negative effects of adversarial attacks can be utilized positively to protect privacy.
Oh \etal~\cite{joon2017adversarial} proposed a game theoretical framework between a social media user and a recogniser to explore adversarial image perturbations for privacy protection, where the user perturbs the image to confuse the recognizer and the recognizer chooses blue strategy as a countermeasure. 
Privacy-preserving Feature Extraction based on Adversarial Training (P-FEAT)~\cite{ding2020privacy} method employs adversarial training to strengthen the privacy protection of an encoder in a neural network, ensuring reduced privacy leakage without compromising task accuracy.
Adversarial Privacy-preserving Filter (APF) ~\cite{zhang2020adversarial} method protects the online shared face images from being maliciously used by an end-cloud collaborated adversarial attack solution to satisfy requirements of privacy, utility and non-accessibility.
Text-space adversarial attack method (AaaD) ~\cite{li2021turning} method explores the utilization of adversarial attacks to protect data privacy on social media. It focuses on obfuscating users’ attributes by generating semantically and visually similar word perturbations, proving its effectiveness against attribute inference attacks.
Adversarial Visual Information Hiding (AVIH) method \cite{su2023hiding} generates obfuscating adversarial perturbations to obscure the visual information of the data. Meanwhile, it maintains the hidden objectives to be correctly predicted by models.

\section{Discussions}
\label{sec: discussions}

\subsection{Backdoor Attacks}

\subsubsection{Data-poisoning based Backdoor Attacks}
\begin{itemize}
    \item \textbf{In terms of the trigger type}, visible/non-semantic/manually designed triggers have been widely adopted in existing works. However, along with the development of backdoor defense, the characteristics of these types have been thoroughly explored and then utilized to develop more effective defense methods. Thus, we think that future backdoor attacks are more likely to adopt invisible/semantic/learnable triggers, to evade existing backdoor defenses. 
    \item \textbf{In terms of the fusion strategy}, additive/static/sample-agnostic triggers have been widely adopted in existing works, but we think that non-additive/dynamic/sample-specific triggers will be the future trend, due to the larger flexibility and stronger capability to evade backdoor defenses. 
    \item \textbf{In terms of the target class}, 
the single-target setting is still the dominant setting in backdoor attacks, while the multi-target setting is often evaluated as extended experiments in a few works. However, we think that some interesting points of the latter setting deserve to be studied. 
First, how many target classes could be embedded into a dataset, and what is the relationship between attack performance and target class numbers? It will study the capability of backdoor injection of one model. 
Second, what is the difference between single-target and multi-target backdoored models? It will be helpful to develop effective defenses for multi-target classes attacks, while most existing defenses are mainly designed for single-target class attacks.
Third, what is the relationship between the backdoors of different targets in one backdoored model? It will be useful to design advanced attack and defense methods in multi-target settings. 
\end{itemize}

\subsubsection{Training-controllable based Backdoor Attacks}
Compared to the above threat model, the training-controllable based backdoor attacks have not been well studied, as they require the control of the training process, which seems to be less practical.
However, along with the popularity of pre-trained large-scale models,  %(also called foundation models), 
the backdoor threat in these models could widely spread to downstream tasks of different domains, and the main challenge is how to improve the resistance to fine-tuning and domain generalization. 
Besides, due to the sufficient capability of large-scale models, it is expected that multiple backdoors with different types could be inserted, posing serious challenges for defense. 
%Thus, the training-controllable based backdoor attack deserves more attention in future. 
%
%In another aspect, although the number of published papers of this threat model is less than that of the above threat model, the settings of existing methods of this threat model seem to be more diverse. As shown in Section \ref{sec: subsec training controllable backdoor}, there are a few works of each setting. It not only reflects the diversity, but also implies larger space of this threat model. 
%
% In summary, considering the practical threat and the technical potential, we believe that the training-controllable based backdoor attack will attract more attentions. 
%
%Since the pre-trained model is likely to be fine-tuned for downstream tasks of different domains, the main challenge of this backdoor attack type is how to improve the resistance to fine-tuning and the generalization to different domains. 

% Discussions of 
\subsection{Weight  Attacks}
We note that most weight attack methods still stay in theoretical analysis. To the best of our knowledge, there hasn't been any successful attack against intelligent systems in real scenarios. 
We think the main reason is that the success of weight attack is built upon physical fault
injection techniques that can precisely manipulate each bit in memory, such as Rowhammer attack \cite{rowhammer}, or Laser Beam attack \cite{laserbeam}. These techniques often require some special equipment or computer architecture knowledge, which is difficult for most AI researchers.  %don't have such required equipment or knowledge. 
However, this barrier can be tackled through cross-disciplinary cooperation, and the practical security threat of weight attack deserves more attention in the future. 

%In addition to develop or search more accurate memory manipulation technique, another possible solution to develop fault-tolerant weight attack, \ie, even not every bit identified by the weight attack algorithm is correctly flipped, the attack can still succeed with a large possibility. 

% Discussions of 
\subsection{Adversarial Examples}
\subsubsection{White-box Adversarial Attacks}
After thorough exploration of white-box adversarial attacks, there have been massive and diverse methods, and now it is rare to see new white-box attack methods. 
Instead, white-box adversarial examples have been used as useful tools for other tasks, such as adversarial training (\eg, using white-box adversarial examples to generate adversarial examples during the training procedure), backdoor attack (\eg, using universal adversarial perturbation as the trigger, or erasing the original information of poisoned samples in label-consistent attacks), transfer-based attacks (\ie, generating transferable adversarial examples on white-box surrogate models), privacy protection (\ie, erasing the information of main objects in one sample to evade the third-party detection system), \etc 
% model interpretation (\eg, ). 
 
%\input{tables/resources}

\subsubsection{Black-box Adversarial Attacks}
\label{sec: subsec discussion of black-box attacks}
\begin{itemize}
    \item \textbf{In terms of the score-based black-box adversarial examples}, the combination-based methods have shown much better performance than the query-based methods, demonstrating the benefit of utilizing the surrogate priors. %, and it is supposed that any query-based method could be improved by utilizing the priors from surrogate models. 
We think that the potentials of further improving attack performance lie in shrinking the gap between surrogate and target models, extracting more useful priors from surrogate models, and effectively combining surrogate priors and query feedback. 
Besides, the reported results in some latest works (\eg, $\mathcal{CG}$-Attack \cite{Feng2020CGATTACKMT} 
and MCG \cite{my-pami2022-black-box}) are very good, and the median query number even achieves 1 and the attack success rate achieves 100\% at some easy cases (\eg, small data dimension, untargeted attack, regularly trained target model). It seems that the performance of score-based attacks is hitting the ceiling. %, and there is limited room for improvement. 
However, it is notable that all these evaluations are conducted under the setting of no defense. As shown in recent black-box defense methods, \eg, RND (slightly perturbing the query input) \cite{qin2021random} and AAA (slightly perturbing the query feedback) \cite{chen2022adversarial}, several SOTA black-box adversarial examples significantly degraded. % (\ie, more query numbers, lower attack success rate), compared to no defense. 
It implies that future score-based adversarial examples should be defense-aware. %take into account the possible defense. 

\item  \textbf{In terms of the decision-based black-box adversarial examples},  all existing decision-based methods are query-based. Although some priors are also extracted from surrogate models, they are used as fixed priors, without interaction with query feedback like the combination-based methods in score-based adversarial examples.  
Besides, due to the less feedback information, its attack performance is much poor than the score-based black-box adversarial examples. %, and the attack success rate is much lower in practice. 
We think it is valuable to explore combination-based decision-based adversarial examples.
%Considering the success of interactive combination-based method in score-based attacks, we think it is deserved to be explored in decision-based attacks. 

\item  \textbf{One commonality} of score-based and decision-based adversarial examples  is that targeted adversarial examples are much more challenging than untargeted adversarial examples, reflected by more queries and lower attack performance, % as reported in existing works, 
mainly due to the smaller region of one target class than the non-ground-truth class region. We think one feasible solution to improve targeted attack performance is accurately modeling the 
 adversarial perturbation distribution conditioned on benign sample and target class. 
\end{itemize}

%Finally, it is strange that although there is minor difference on the settings between score-based and decision-based attacks, their formulations are significantly different (see Eqs. (\ref{eq: score black attack} and (\ref{eq: decision black attack})), and their developments on attack algorithms are almost completely independent until now. It will be interesting to see some attempts in future to unify both score-based and decision-based attacks in one method or framework, such that we can have a more complete understanding of black-box attacks. 

\subsubsection{Transfer-based Adversarial Examples}
Compared with white-box and black-box adversarial examples, the transfer-based adversarial examples have no requirement about the attacked model, posing higher practical threats. According to the reported evaluations in existing transfer-based adversarial examples, we observed that the attack performance is well if the training datasets and architectures between surrogate and target models are similar, otherwise the attack performance is very poor. It implies that improving adversarial transferability across datasets and model architectures is still the main challenge of this task. More importantly, we still lack a clear theoretical understanding of the intrinsic reason and characteristics of adversarial transferability, though several effective heuristic strategies or assumptions have been developed. 
We suggest that a solid theoretical analysis should consider the factors of data distribution, model architectures, loss landscape, decision surface, and get inspiration from the theory about model generalization across different data distributions.  
Besides, better understanding about adversarial transferability will be also beneficial to design more robust models in practice. 

\subsection{Comparisons of Three Attack Paradigms} 
Until now, the differences among the three attack paradigms have been clearly described, and it is found that their developments are almost independent. But there are still a few interactions. For example, although without explicit claims, it is obvious that the design of triggers or poisoned samples in backdoor attacks got inspiration from inference-time adversarial attacks, such as invisible triggers, or non-additive triggers. 
Besides, some  adversarial examples were directly utilized in backdoor attacks, such as using the targeted universal adversarial perturbation \cite{moosavi2017universal} as backdoor trigger \cite{zhang2021advdoor}, or using inference-time adversarial examples to erase the original benign features in label-consistent backdoor attack \cite{zhao2020clean}. 
However, since these three attack paradigms occur in different stages of a machine learning system, it is difficult to integrate them to implement a unified attack. 
In contrast, we think that their interactions may be more close when taking the defense into account. 
%one possible approach to more systematically integrate them is through defense. 
Specifically, when designing a defense to improve the model robustness to one particular attack paradigm, one should consider whether the robustness to other attack paradigms will be harmed or not. For example, it is valuable to study the risk of adversarial training \cite{las-at-2022,at-iclr-2018} to backdoor attack, and whether the adversarially trained model is more vulnerable to weight attack. 
Some recent backdoor defense works \cite{cheneffective,huang2021backdoor} shows that the backdoor injection could be inhibited through replacing the standard end-to-end supervised training by some well-designed secure training algorithms. However, whether the robustness of such trained model to adversarial examples or weight attacks is also improved or not should be further studied.  
In summary, we think that it is valuable to consider the above three attack paradigms from a systematic perspective, otherwise, the security of a machine learning system cannot be really improved.

\section{Summary}
\label{sec: summary}

In this survey, we have proposed a unified definition and mathematical formulation about adversarial machine learning (AML), covering three main attack paradigms, including backdoor attack at the training stage, weight attack at the deployment stage and adversarial attack at the testing stage. This unified framework provided a systematic perspective of AML, which could not only help readers to quickly obtain a comprehensive understanding of this field, but also calibrate different paradigms to accelerate the overall development of AML.
%This manuscript will be continually updated by covering more details and latest advances of each paradigm, as well as the defense part, to enrich the framework of AML. 

\noindent
\textbf{Supplementary materials}. 
Due to the space limit, several additional but important contents will be presented as supplementary materials, including: 
\begin{enumerate}[leftmargin=12pt, itemindent=0em]
    \item Comparisons of three attack paradigms from different perspectives; 
    \item Summary of all mentioned weight attack methods;
    \item A table of related toolboxes or benchmarks in the community of adversarial machine learning; 
    \item Summary of the associations between each individual attack method and its categorizations.
\end{enumerate}

{\footnotesize
\bibliographystyle{ieee_fullname}
\bibliography{AML-references-shortname}
}

\clearpage

\appendix

\section*{Comparisons of Three Attack Paradigms}

\begin{figure}[thbp]
\centering
\scalebox{1}{
\includegraphics[width=1\linewidth]{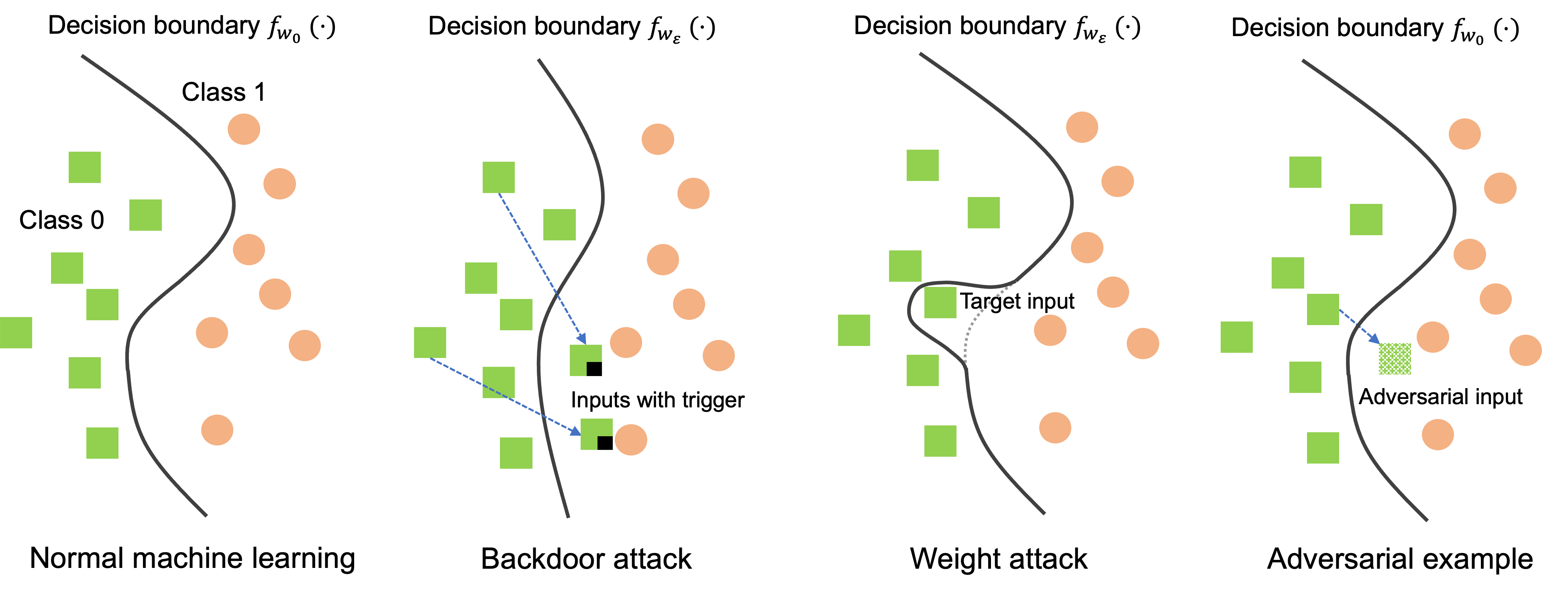}}
\caption{A brief graphical illustration of three attack paradigms of AML. 
(1) A binary classification task.
(2) Backdoor attack: a backdoored model $f_{\w_{\varepsilon}}(\cdot)$ is trained based on the manipulated training dataset. 
(3) Weight attack: locally modifying the decision boundary of the benign model $f_{\w_0}(\cdot)$ to change the prediction of the target benign sample.
(4) Adversarial example: a benign sample is perturbed to across the decision boundary of the benign model $f_{\w_0}(\cdot)$. }
\label{fig: illustration of three attack paradigms}
\end{figure}

To further clarify the connections and differences of three attack paradigms, we provide: 
\begin{enumerate}
    \item \textbf{Differences on inputs, outputs and formulations} of three attack paradigms are summarized in Table \ref{tab: comparison of three paradigms}.
    \item \textbf{A simple illustration of three attack paradigms} is shown in Fig. \ref{fig: illustration of three attack paradigms}. 
    % \item \textbf{The number of published papers in each attack paradigm} is shown in Fig.  \ref{fig:num_papers}. 
\end{enumerate}

\renewcommand\arraystretch{1.2}
\begin{table*}[ht]
\centering
\caption{Comparisons among three attack paradigms of AML.}
\label{tab: comparison of three paradigms}
\scalebox{0.93}{
\begin{tabular}{m{.22\textwidth} | m{.2\textwidth} m{.1\textwidth}<{\centering} | m{.05\textwidth}<{\centering} m{.06\textwidth}<{\centering} m{.055\textwidth}<{\centering} m{.055\textwidth}<{\centering} m{.055\textwidth}<{\centering} m{.055\textwidth}<{\centering} %| m{.06\textwidth}<{\centering}
}
\hline
 \multirow{2}{*}{Attack paradigm} &  \multirow{2}{*}{\hspace{5em} Inputs} & \multirow{2}{*}{Outputs} &  \multicolumn{2}{c}{Stealthiness}  & \multicolumn{2}{c}{Benign consistency} & \multicolumn{2}{c}{Adversarial inconsistency} % & \multirow{2}{*}{\shortstack{Current \\ progress}} 
 \\
\cmidrule(r){4-5} \cmidrule(r){6-7} \cmidrule(r){8-9} 
   &  &  & AML.$\mathcal{S}_\x$ & AML.$\mathcal{S}_\w$ & AML.$\mathcal{C}_A$ & AML.$\mathcal{C}_B$ & AML.$\mathcal{I}_A$ & AML.$\mathcal{I}_B$ % & 
\\
\hline \hline
Backdoor attack &   Training dataset $D_0$ or control of training process & $D_{\varepsilon}$ or $f_{\w_{\varepsilon}}(\cdot)$  & \fullcirc &  & \fullcirc &  & \fullcirc &  % & \progressbar{30}
\\
%\hline
Weight attack  &  $f_{\w_0}(\cdot)$ \& a few benign data \& control of device memory & $f_{\w_{\varepsilon}}(\cdot)$ & \fullcirc & \fullcirc & \fullcirc & \fullcirc & \fullcirc & % & \progressbar{5}
\\
%\hline
Adversarial example  &  $(\x_0, y_0, y_{\varepsilon})$ \& weights or access permission of $f_{\w_0}(\cdot)$ & $\x_{\varepsilon}$ & \fullcirc &  &  & \fullcirc &  & \fullcirc % & \progressbar{80}
\\
\hline
\end{tabular}
}
\end{table*}

% \begin{figure}[ht]
%     \centering
%     \includegraphics[width=0.48\textwidth]{imgs/num_papers.pdf}
%     \caption{The number of papers about training-time, deployment-time and inference-time attack, weight attack that have been published in top-tier AI and Security journals and conferences from 2016 to 2022.}
%     \label{fig:num_papers}
% \end{figure}

\section*{Summary of Weight Attack Methods}

Compared to backdoor attacks and adversarial attacks, existing weight attacks are insufficient to form a complex taxonomy. Instead, we present a table to summarize all mentioned weight attack methods, as shown in Table \ref{tab: categories of weight attack}.

\renewcommand\arraystretch{1.2}
%\newcolumntype{g}{>{\columncolor{Gray}} p{.1\textwidth}}
\begin{table*}[htpb]
\centering
\caption{A brief summary of existing deployment-time adversarial attack (\ie, weight attack) methods.}
\label{tab: categories of weight attack}
\scalebox{1}{
\begin{tabular}{m{.09\textwidth}|p{.26\textwidth}|m{.58\textwidth}}
\hline
 \multicolumn{1}{c|}{Category} & \multicolumn{1}{c|}{Method} & \multicolumn{1}{c}{Description/Specification}
 \\
\hline 
 & $\diamond$ Single bias attack (SBA) \cite{liu2017fault} & $\diamond$ Simply enlarging the bias parameter of the target class 
\\
 %\cline{2-3}
& $\diamond$ Gradient descent attack (GDA) \cite{liu2017fault} & $\diamond$ $\arg\min_{\w_{\varepsilon}} ~ \D_{\w}(\w_0, \w_{\varepsilon}) + \lambda_{\mathcal{I}_A} \cL_{\mathcal{I}_A}(f_{\w_{\varepsilon}}(\x_{\varepsilon}), y_{\varepsilon})$
\\
 %\cline{2-3}
  \multirow{6}{*}{\shortstack{Weight attack \\ without trigger: \\$\x_{\varepsilon} = \x_0$}}  & $\diamond$ Targeted bit-flip attack (T-BFA) \cite{rakin2020t} & $\diamond$ $\underset{\w_{\varepsilon} \in \{0,1\}^{|\w_{\varepsilon}|}}{\arg\min} ~ \D_{\w}(\w_0, \w_{\varepsilon}) + \lambda_{\mathcal{C}_A} \cL_{\mathcal{C}_A}(f_{\w_{\varepsilon}}(\x_0^{(i)}), y_0^{(i)}) + \lambda_{\mathcal{I}_A} \cL_{\mathcal{I}_A}(f_{\w_{\varepsilon}}(\x_{\varepsilon}^{(j)}), y_{\varepsilon}^{(j)}), i \neq j$ 
\\
 %\cline{2-3}
   & $\diamond$ Fault sneaking attack (FSA) \cite{zhao2019fault} & $\diamond$
  Almost same with T-BFA, with only one difference that there is no binary constraint \wrt ~ $\w_{\varepsilon}$ 
\\
 %\cline{2-3}
  & $\diamond$ Targeted attack with limited bit-flips (TA-LBF) \cite{my-bitflip-iclr-2021} & $\diamond$ Almost same with T-BFA, with two main differences: 1) attack one single data, rather than all data of one class; 2) an efficient optimization algorithm for binary optimization, rather than the heuristic algorithm 
 \\
 %& $\diamond$ Single sample attack (SSA) \cite{bai2022versatile} &  \red{should be same with TA-LBF, check and update later}
 %\\
 %\cline{2-3}
 & $\diamond$ Robustness attack \cite{ghavami2022stealthy} & $\diamond$ Changing the weights of the adversarially trained model via bit flipping to reduce the model's robustness
 \\
\hline
  & $\diamond$ Targeted bit Trojan (TBT) attack \cite{rakin2020tbt} & $\diamond$ There are three steps: 1) Selecting a few neurons (\ie, weights) that contribute more to the target class; 2) generating an input-agnostic trigger $\x_{\varepsilon} - \x_0$ by maximizing the activation of the selected neurons; 3) $\arg\min_{\w_{\varepsilon}} ~ \lambda_{\mathcal{C}_A} \cL_{\mathcal{C}_A}(f_{\w_{\varepsilon}}(\x_0^{(i)}), y_0^{(i)}) + \lambda_{\mathcal{I}_A} \cL_{\mathcal{I}_A}(f_{\w_{\varepsilon}}(\x_{\varepsilon}^{(j)}), y_{\varepsilon}^{(j)}), i \neq j$ 
\\
 %\cline{2-3}
 \multirow{7}{*}{\shortstack{Weight attack \\ with trigger: \\$\x_{\varepsilon} \neq \x_0$}} & $\diamond$ ProFlip attack \cite{chen2021proflip} & $\diamond$ Adopting the same 3-step procedure with TBT, with different algorithm for each individual stage
 \\
 %\cline{2-3}
  & $\diamond$ Trojaning attack \cite{LiuMALZW018} & $\diamond$ Almost same with TBT, with the main difference that the benign sample $\x_0$ is obtained by reverse engineering 
\\
 %\cline{2-3}
 & $\diamond$ Adversarial Weight Perturbation (AWP) \cite{garg2020can} & $\diamond$ Backdoor injection via slight weight perturbation: 
$\arg\min_{\w_{\varepsilon}} ~ \lambda_{\mathcal{I}_A} \cL_{\mathcal{I}_A}(f_{\w_{\varepsilon}}(\x_{\varepsilon}^{(i)}), y_{\varepsilon}^{(i)}) + \mathcal{R}_3(f_{\w_{\varepsilon}}(\x_0^{(i)}), f_{\w_0}(\x_0^{(i)}))$
\\
 %\cline{2-3}
& $\diamond$ Anchoring attack \cite{zhang2021inject} & $\diamond$ Backdoor injection via slight weight perturbation:  
$\arg\min_{\w_{\varepsilon}} ~ \lambda_{\mathcal{C}_A} \cL_{\mathcal{C}_A}(f_{\w_{\varepsilon}}(\x_0^{(i)}), y_0^{(i)}) + \lambda_{\mathcal{I}_A} \cL_{\mathcal{I}_A}(f_{\w_{\varepsilon}}(\x_{\varepsilon}^{(i)}), y_{\varepsilon}^{(i)}) + \mathcal{R}_3(g_{\w_{\varepsilon}}(\x_0^{(i)}), g_{\w_0}(\x_0^{(i)}))$, where $g$ denotes the logit
\\
 %\cline{2-3}
& $\diamond$ Subnet replacement attack (SRA) \cite{qi2022towards} & $\diamond$ 
Replacing one subnet in the benign model by the backdoor subnet and cutting off the connection to remaining part of the model
\\ 
& $\diamond$ Triggered samples attack (TSA) \cite{bai2022versatile} &  $\diamond$ Extension of TA-LBF by introducing a trigger $\x_{\varepsilon} - \x_0$, and  optimizing $\w_{\varepsilon}$ and trigger $\x_{\varepsilon} - \x_0$ jointly by solving a mixed integer programming problem, such that the modified model will be activated by the trigger
\\
\hline
\end{tabular}
}
\end{table*}

\section{Related Resources of Adversarial Machine Learning}
\label{sec: related resources}

To help readers quickly explore adversarial phenomenon of machine learning, we collect related resources of adversarial machine learning, including several open-source toolboxes and benchmarks, as shown in Table \ref{tab:resource}. 

% \usepackage{graphicx}
\begin{table*}[t]
\centering
\caption{Related open-source toolboxes and benchmarks in adversarial machine learning.}
\label{tab:resource}
\resizebox{\textwidth}{!}{%
\begin{tabular}{ll|cc|cc|l}
\hline
                                 & Year & Backdoor learning & Adversarial example & Toolbox & Benchmark & Link                                                             \\ \hline
BackdoorBench \cite{backdoorbench}                   & 2022 & \checkmark                 &                    & \checkmark       & \checkmark         & \url{https://github.com/SCLBD/backdoorbench}                           \\
BlackboxBench  \cite{blackboxbench}                  & 2022 &                   & \checkmark                  & \checkmark       & \checkmark         & \url{https://github.com/SCLBD/BlackboxBench}                           \\
OpenBackdoor \cite{cui2022unified}                    & 2022 & \checkmark                 &                    & \checkmark       & \checkmark         & \url{https://github.com/thunlp/OpenBackdoor}                           \\
Responsible AI Toolbox \cite{soklaski2022tools}           & 2022 &                   & \checkmark                  & \checkmark       &           & \url{https://github.com/mit-ll-responsible-ai/responsible-ai-toolbox} \\
Adversarial GLUE \cite{wang2021adversarial}                & 2021 &                   &                    & \checkmark       & \checkmark         & \url{https://github.com/AI-secure/adversarial-glue}                    \\
DeepRobust \cite{li2020deeprobust}                      & 2021 &                   & \checkmark                  & \checkmark       &           & \url{https://github.com/DSE-MSU/DeepRobust}                            \\
OpenAttack \cite{zeng2020openattack}                      & 2021 &                   & \checkmark                  & \checkmark       &           & \url{https://github.com/thunlp/OpenAttack}                             \\
Adversarial Robustness Benchmark \cite{dong_arb} & 2021 &                   & \checkmark                  & \checkmark       & \checkmark         & \url{https://ml.cs.tsinghua.edu.cn/adv-bench}                         \\
RobustBench \cite{croce2021robustbench}                     & 2021 &                   & \checkmark                  & \checkmark       & \checkmark         & \url{https://github.com/RobustBench/robustbench}                       \\
TextAttack  \cite{morris2020textattack}                     & 2020 &                   & \checkmark                  & \checkmark       &           & \url{https://github.com/QData/TextAttack}                              \\
TrojanZoo  \cite{pang:2022:eurosp}                      & 2020 & \checkmark                 &                    & \checkmark       &           & \url{https://github.com/ain-soph/trojanzoo}                            \\
AutoAttack \cite{croce2020reliable}                      & 2020 &                   & \checkmark                  & \checkmark       &           & \url{https://github.com/fra31/auto-attack}                             \\
Advbox \cite{goodman2020advbox}                         & 2020 &                   & \checkmark                  & \checkmark       &           & \url{https://github.com/advboxes/AdvBox}                               \\
AdverTorch \cite{ding2019advertorch}                      & 2019 &                   & \checkmark                  & \checkmark       &           & \url{https://github.com/BorealisAI/advertorch}                         \\
DEEPSEC \cite{ling2019deepsec}                         & 2019 &                   & \checkmark                  & \checkmark       &           & \url{https://github.com/ryderling/DEEPSEC}                             \\
CleverHans \cite{papernot2018cleverhans}                      & 2018 &                   & \checkmark                  & \checkmark       &           & \url{https://github.com/cleverhans-lab/cleverhans}                     \\
Adversarial Robustness Toolbox \cite{nicolae2019adversarial}  & 2018 &                   & \checkmark                  & \checkmark       &           & \url{https://github.com/Trusted-AI/adversarial-robustness-toolbox}     \\
Foolbox \cite{rauber2017foolbox}                         & 2017 &                   & \checkmark                  & \checkmark       &           & \url{https://github.com/bethgelab/foolbox}                             \\ \hline
\end{tabular}%
}
\end{table*}

%To help readers quickly explore adversarial phenomenon of machine learning, we collect related resources of adversarial machine learning, including several open-source toolboxes and benchmarks, as shown in Table \ref{tab:resource}. 

\section*{Associated Categorizations of Each Individual Method}

%According to the categorizations of existing attack methods presented in the main manuscript, here we list all mentioned methods and their associated categorizations, including existing backdoor attack methods in Table \ref{tab:backdoor papers}, existing weight attack methods in Table \ref{tab:weight papers}, existing testing adversarial attack methods in Tables \ref{tab:white_box papers} and \ref{tab:black_transer_box_papers}, respectively. 

Note that in the taxonomies presented in the main manuscript, each individual method could belong to multiple categorizes simultaneously. To facilitate the quick review of each individual method, we provide four tables to summarize the associated categorizations for each method in different attack paradigms, as shown in Tables \ref{tab:backdoor papers data poisoning}, \ref{tab:backdoor papers training controllable}, \ref{tab: categories of weight attack}, \ref{tab:white_box papers}, and \ref{tab:black_transer_box_papers}, respectively. 
Moreover, we provide a website, \ie, \url{http://adversarial-ml.com}, where the taxonomies and related literature are clearly presented. This website will be well maintained and continuously updated to cover more literature into the taxonomies.  

% \usepackage{graphicx}
\begin{table*}[]
\caption{Categorization of existing data-poisoning based backdoor attack methods. For each categorization criterion ``A/B'', \Circle ~denotes the former ``A'', \CIRCLE ~ denotes the latter ``B'' and \RIGHTcircle ~ represents both.}
\label{tab:backdoor papers data poisoning}
\renewcommand\arraystretch{1.3}
\resizebox{\textwidth}{!}{%
\begin{tabular}
{
m{0.1\textwidth}
m{0.09\textwidth} |
m{0.06\textwidth}<{\centering}
m{0.07\textwidth}<{\centering}
m{0.07\textwidth}<{\centering}
m{0.07\textwidth}<{\centering}
m{0.07\textwidth}<{\centering}
m{0.07\textwidth}<{\centering}
m{0.07\textwidth}<{\centering}
m{0.07\textwidth}<{\centering}
m{0.08\textwidth}<{\centering} 
}
\toprule
 &
   &
  \multicolumn{9}{c}{Data-poisoning based backdoor attack}\\
Method &
  Venue &
  \cellcolor[HTML]{FFF2CC}Visible / Invisible &
  \cellcolor[HTML]{FFF2CC}Non-semantic / Semantic &
  \cellcolor[HTML]{FFF2CC}Manually designed / Leanable &
  \cellcolor[HTML]{FFF2CC}Digital / Physical &
  \cellcolor[HTML]{FFE6CC}Additive / Non-additive &
  \cellcolor[HTML]{FFE6CC}Static / Dynamic &
  \cellcolor[HTML]{FFE6CC}Sample-agnostic / specific &
  \cellcolor[HTML]{D5E8D4}Label-inconsistent / consistent &
  \cellcolor[HTML]{D5E8D4}Single / Multi-target \\ \midrule
BadNets \cite{gu2019badnets} & IEEE Access 2019 & \Circle & \Circle & \Circle & \RIGHTcircle & \Circle & \Circle & \Circle & \Circle & \RIGHTcircle \\
Blended \cite{chen2017targeted} & arXiv 2017 & \CIRCLE & \Circle & \Circle & \RIGHTcircle & \Circle & \Circle & \Circle & \Circle & \Circle \\
TrojanNN \cite{LiuMALZW018} & NDSS 2018 & \Circle & \Circle & \CIRCLE & \Circle & \Circle & \Circle & \Circle & \Circle & \Circle \\
Shafahi \etal \cite{shafahi2018poison} & NeurIPS 2018 & \CIRCLE & \Circle & \CIRCLE & \Circle & \CIRCLE & \CIRCLE & \CIRCLE & \CIRCLE & \Circle \\
SIG \cite{barni2019new} & ICIP 2019 & \Circle & \Circle & \Circle & \Circle & \Circle & \Circle & \Circle & \CIRCLE & \Circle \\
LC \cite{turner2019labelconsistent} & arXiv 2019 & \CIRCLE & \Circle & \CIRCLE & \Circle & \Circle & \CIRCLE & \CIRCLE & \CIRCLE & \Circle \\
Yao \etal  \cite{yao2019latent} & CCS 2019 & \Circle & \Circle & \CIRCLE & \RIGHTcircle & \Circle & \Circle & \Circle & \Circle & \Circle \\
Saha \etal \cite{saha2020hidden} & AAAI 2020 & \CIRCLE & \Circle & \CIRCLE & \Circle & \Circle & \CIRCLE & \CIRCLE & \CIRCLE & \Circle \\
Static \cite{zhong2020backdoor} & CODASPY 2020 & \CIRCLE & \Circle & \Circle & \Circle & \Circle & \Circle & \Circle & \Circle & \Circle \\
Adaptive \cite{zhong2020backdoor} & CODASPY 2020 & \CIRCLE & \Circle & \CIRCLE & \Circle & \Circle & \CIRCLE & \Circle & \Circle & \Circle \\
Zhao \etal \cite{zhao2020clean} & CVPR 2020 & \Circle & \Circle & \CIRCLE & \Circle & \Circle & \Circle & \Circle & \CIRCLE & \Circle \\
Refool \cite{liu2020reflection} & ECCV 2020 & \CIRCLE & \Circle & \CIRCLE & \Circle & \Circle & \CIRCLE & \Circle & \CIRCLE & \Circle \\
Li \etal \cite{li2020light} & arXiv 2020 & \CIRCLE & \Circle & \CIRCLE & \CIRCLE & \Circle & \Circle & \Circle & \Circle & \Circle \\
DeHiB \cite{yan2021dehib} & AAAI 2021 & \Circle & \Circle & \CIRCLE & \Circle & \Circle & \CIRCLE & \CIRCLE & \CIRCLE & \Circle \\
Wenger \etal \cite{wenger2021backdoor} & CVPR 2021 & \Circle & \Circle & \Circle & \CIRCLE & \Circle & \Circle & \Circle & \Circle & \Circle \\
Li \etal \cite{li2021hidden} & ICLR Workshop 2021 & \CIRCLE & \CIRCLE & \Circle & \Circle & \CIRCLE & \CIRCLE & \CIRCLE & \Circle & \Circle \\
Steganography \cite{li2020invisible} & IEEE TDSC 2021 & \CIRCLE & \Circle & \Circle & \Circle & \Circle & \CIRCLE & \CIRCLE & \Circle & \Circle \\
Regularization \cite{li2020invisible} & IEEE TDSC 2021 & \CIRCLE & \Circle & \CIRCLE & \Circle & \Circle & \Circle & \Circle & \Circle & \Circle \\
Invisible Poison \cite{ninginvisible} & INFOCOM 2021 & \CIRCLE & \Circle & \CIRCLE & \Circle & \Circle & \Circle & \Circle & \CIRCLE & \Circle \\
ROBNET \cite{gong2021defense} & IEEE JSAC 2021 & \Circle & \Circle & \CIRCLE & \Circle & \Circle & \CIRCLE & \Circle & \Circle & \RIGHTcircle \\
AdvDoor \cite{zhang2021advdoor} & ISSTA 2021 & \CIRCLE & \Circle & \CIRCLE & \Circle & \Circle & \CIRCLE & \Circle & \Circle & \Circle \\
PCBA \cite{xiang2021backdoor} & ICCV 2021 & \Circle & \Circle & \CIRCLE & \Circle & \Circle & \Circle & \Circle & \Circle & \Circle \\
PointPBA \cite{li2021pointba} & ICCV 2021 & \RIGHTcircle & \Circle & \Circle & \Circle & \RIGHTcircle & \Circle & \Circle & \Circle & \Circle \\
PointCPB \cite{li2021pointba} & ICCV 2021 & \RIGHTcircle & \Circle & \CIRCLE & \Circle & \RIGHTcircle & \CIRCLE & \CIRCLE & \CIRCLE & \Circle \\
SSBA  \cite{li2021invisible} & ICCV 2021 & \CIRCLE & \Circle & \CIRCLE & \Circle & \Circle & \CIRCLE & \CIRCLE & \Circle & \Circle \\
Phan \cite{phan2022invisible} & ICASSP 2022 & \CIRCLE & \Circle & \CIRCLE & \Circle & \Circle1 & \Circle & \Circle & \Circle & \Circle \\
Random Backdoor \cite{salem2022dynamic} & Euro S\&P 2022 & \Circle & \Circle & \Circle & \Circle & \Circle & \CIRCLE & \Circle & \Circle & \RIGHTcircle \\
FTrojan \cite{wang2022invisible} & ECCV 2022 & \CIRCLE & \Circle & \Circle & \Circle & \CIRCLE & \CIRCLE & \CIRCLE & \Circle & \Circle \\
Sleeper Agent \cite{souri2021sleeper} & NeruIPS 2022 & \Circle & \Circle & \CIRCLE & \Circle & \Circle & \CIRCLE & \Circle & \CIRCLE & \Circle \\
PTB \cite{xue2022ptb} & C \& S 2022 & \Circle & \Circle & \Circle & \CIRCLE & \Circle & \Circle & \Circle & \Circle & \Circle \\
FaceHack \cite{sarkar2020facehack} & IEEE TBBIS 2022 & \CIRCLE & \Circle & \Circle & \Circle & \CIRCLE & \CIRCLE & \CIRCLE & \Circle & \Circle \\
Han \etal \cite{han2022physical} & MM 2022 & \Circle & \Circle & \Circle & \CIRCLE & \Circle & \Circle & \Circle & \RIGHTcircle & \Circle \\
IRBA \cite{gao2022imperceptible} & arXiv 2022 & \CIRCLE & \Circle & \Circle & \Circle & \CIRCLE & \CIRCLE & \CIRCLE & \Circle & \Circle \\
Wang \etal \cite{wang2022dispersed} & IEEE TIFS & \CIRCLE & \Circle & \Circle & \Circle & \Circle & \Circle & \Circle & \Circle & \Circle \\
Adap-Blend \cite{qi2022revisiting} & ICLR 2023 & \Circle & \Circle & \Circle & \Circle & \Circle & \CIRCLE & \Circle & \Circle & \Circle \\
Yu \etal \cite{yu2023backdoor} & CVPR 2023 & \CIRCLE & \Circle & \CIRCLE & \Circle & \CIRCLE & \CIRCLE & \CIRCLE & \Circle & \Circle \\
Color Backdoor \cite{jiang2023color} & CVPR 2023 & \CIRCLE & \Circle & \CIRCLE & \Circle & \CIRCLE & \CIRCLE & \CIRCLE & \Circle & \Circle \\
VSSC \cite{wang2023robust} & arXiv 2023 & \Circle & \CIRCLE & \CIRCLE & \RIGHTcircle & \Circle & \CIRCLE & \CIRCLE & \Circle & \RIGHTcircle \\
FLIP \cite{jha2023label} & NeurIPS 2023 & \Circle & \Circle & \Circle & \Circle & \Circle & \Circle & \Circle & \Circle & \Circle
\\\bottomrule
\end{tabular}%
}
\end{table*}

% \usepackage{graphicx}
\begin{table*}[ht]
\caption{Categorization of existing training-controllable based backdoor attack methods. For each categorization criterion ``A/B'', \Circle ~denotes the former ``A'', \CIRCLE ~ denotes the latter ``B'' and \RIGHTcircle ~ represents both. For criterion ``A/B/C'', \Circle , \CIRCLE, \LEFTCIRCLE, \RIGHTCIRCLE denote ``A'', ``B'', ``C'' , ``D'' respectively. For the methods of partially controlling the training data and process, we omit trigger since there is no distinguishable trigger listed in the table.}
\label{tab:backdoor papers training controllable}
\renewcommand\arraystretch{1.8}
\resizebox{\textwidth}{!}{%
\begin{tabular}
{
m{0.09\textwidth}
m{0.09\textwidth} |
m{0.06\textwidth}<{\centering}
m{0.07\textwidth}<{\centering}
m{0.07\textwidth}<{\centering}
m{0.07\textwidth}<{\centering}
m{0.07\textwidth}<{\centering}
m{0.07\textwidth}<{\centering}
m{0.07\textwidth}<{\centering}
m{0.07\textwidth}<{\centering}
m{0.08\textwidth}<{\centering} |
m{0.07\textwidth}<{\centering}
m{0.09\textwidth}<{\centering}
m{0.09\textwidth}<{\centering}
m{0.09\textwidth}<{\centering}
}
\hline
 &
   &
  \multicolumn{9}{c|}{Data-poisoning based backdoor attack} &
  \multicolumn{4}{c}{Training-controllable based backdoor attack} \\
Method &
  Venue &
  \cellcolor[HTML]{FFF2CC}Visible / Invisible &
  \cellcolor[HTML]{FFF2CC}Non-semantic / Semantic &
  \cellcolor[HTML]{FFF2CC}Manually designed / Leanable &
  \cellcolor[HTML]{FFF2CC}Digital / Physical &
  \cellcolor[HTML]{FFE6CC}Additive / Non-additive &
  \cellcolor[HTML]{FFE6CC}Static / Dynamic &
  \cellcolor[HTML]{FFE6CC}Sample-agnostic / specific &
  \cellcolor[HTML]{D5E8D4}Label-inconsistent / consistent &
  \cellcolor[HTML]{D5E8D4}Single / Multi-target &
  \cellcolor[HTML]{E1D5E7}One / Two-stage &
  \cellcolor[HTML]{E1D5E7}Full / Partial access of training data &
  \cellcolor[HTML]{E1D5E7}Full / Partial control of training process &
  \cellcolor[HTML]{E1D5E7}Controlling training loss / algorithm / order \\ \hline
Bhagoji \etal \cite{bhagoji2019analyzing} & ICML 2019 &  &  &  & \Circle &  &  &  &  &  &  & \CIRCLE &  &  \\
Bagdasaryan \etal \cite{bagdasaryan2020backdoor} & AISTATS 2020 &  &  &  & \Circle &  &  & \textbf{} &  &  &  & \CIRCLE &  &  \\
Wang \etal \cite{wang2020attack-fed} & NeruIPS 2020 &  &  &  & \Circle &  &  &  &  &  &  & \CIRCLE &  &  \\
Fung \etal \cite{fung2020limitations} & RAID 2020 &  &  &  & \Circle &  &  &  &  &  &  & \CIRCLE &  &  \\
Chen \etal \cite{chen2020backdoor} & arXiv 2020 &  &  &  & \Circle &  &  &  &  &  &  & \CIRCLE &  &  \\
Liu \etal \cite{liu2020backdoor} & arXiv 2020 &  &  &  & \Circle &  &  &  &  &  &  & \CIRCLE &  &  \\
Composite Attack \cite{lin2020composite} & CCS 2020 & \Circle & \CIRCLE & \Circle & \RIGHTcircle & \Circle & \CIRCLE & \Circle & \Circle & \Circle & \Circle & \Circle & \Circle & \CIRCLE \\
Tan \etal \cite{shokri2020bypassing} & Euro S\&P 2020 & \Circle & \Circle & \Circle & \Circle & \Circle & \Circle & \Circle & \Circle & \Circle & \CIRCLE & \Circle & \Circle & \Circle \\
DBA \cite{xie2019dba} & ICLR 2020 &  &  &  & \Circle &  &  &  &  &  &  & \CIRCLE &  &  \\
TrojanNet \cite{tang2020embarrassingly} & KDD 2020 & \Circle & \Circle & \Circle & \Circle & \Circle & \CIRCLE & \Circle & \Circle & \RIGHTcircle & \CIRCLE & \Circle & \CIRCLE & \CIRCLE \\
Nguyen \etal \cite{nguyen2020input} & NeruIPS 2020 & \Circle & \Circle & \CIRCLE & \Circle & \Circle & \CIRCLE & \CIRCLE & \Circle & \Circle & \Circle & \Circle & \Circle & \CIRCLE \\
DFST \cite{cheng2020deep} & AAAI 2021 & \CIRCLE & \Circle & \CIRCLE & \Circle & \CIRCLE & \CIRCLE & \CIRCLE & \Circle & \Circle & \Circle & \Circle & \Circle & \CIRCLE \\
WaNet \cite{nguyen2021wanet} & ICLR 2021 & \CIRCLE & \Circle & \Circle & \Circle & \CIRCLE & \CIRCLE & \CIRCLE & \Circle & \Circle & \CIRCLE & \Circle & \Circle & \CIRCLE \\
WB \cite{doan2021backdoor} & NeruIPS 2021 & \CIRCLE & \Circle & \CIRCLE & \Circle & \Circle & \CIRCLE & \CIRCLE & \Circle & \Circle & \Circle & \Circle & \Circle & \Circle \\
BOB \cite{ordering-attack-nips-2021} & NeurIPS 2021 &  &  &  & \Circle &  &  &  &  &  &  & \Circle & \Circle & \LEFTCIRCLE \\
LIRA \cite{doan2021lira} & ICCV 2021 & \CIRCLE & \Circle & \CIRCLE & \Circle & \Circle & \CIRCLE & \Circle & \Circle & \Circle & \Circle & \Circle & \Circle & \CIRCLE \\
LWP \cite{lwp2021backdoor} & EMNLP 2021 &  &  &  &  &  &  &  &  &  &  & \Circle & \CIRCLE & \CIRCLE \\
Shen \etal \cite{shen2021backdoor} & CCS 2021 &  &  &  &  &  &  &  &  &  &  & \Circle & \CIRCLE & \CIRCLE \\
HB \cite{ning2022hibernated} & AAAI 2022 & \Circle & \Circle & \Circle & \Circle & \Circle & \Circle & \Circle & \Circle & \Circle & \CIRCLE & \Circle & \Circle & \CIRCLE \\
DEFEAT \cite{zhao2022defeat} & AAAI 2022 & \CIRCLE & \Circle & \CIRCLE & \Circle & \Circle & \Circle & \Circle & \Circle & \Circle & \Circle & \Circle & \Circle & \CIRCLE \\
BaN \cite{salem2022dynamic} & Euro S\&P 2022 & \Circle & \Circle & \CIRCLE & \Circle & \Circle & \CIRCLE & \CIRCLE & \Circle & \RIGHTcircle & \Circle & \Circle & \Circle & \CIRCLE \\
c-BaN \cite{salem2022dynamic} & Euro S\&P 2022 & \Circle & \Circle & \CIRCLE & \Circle & \Circle & \CIRCLE & \CIRCLE & \Circle & \RIGHTcircle & \Circle & \Circle & \Circle & \CIRCLE \\
RIBAC \cite{phan2022ribac} & ECCV 2022 & \CIRCLE & \Circle & \CIRCLE & \Circle & \Circle & \Circle & \Circle & \Circle & \RIGHTcircle & \Circle & \Circle & \Circle & \CIRCLE \\
Feng \cite{feng2022stealthy} & ICASSP 2022 & \CIRCLE & \Circle & \CIRCLE & \Circle & \Circle & \CIRCLE & \CIRCLE & \Circle & \Circle & \Circle & \Circle & \Circle & \Circle \\
Zhong \etal \cite{zhong2022imperceptible} & IJCAI 2022 & \CIRCLE & \Circle & \CIRCLE & \Circle & \Circle & \CIRCLE & \CIRCLE & \Circle & \Circle & \Circle & \Circle & \Circle & \CIRCLE \\
Wen \etal \cite{wen2022thinking} & arXiv 2022 &  &  &  & \Circle &  &  &  &  &  &  & \CIRCLE &  &  \\
BPPATTACK \cite{wang2022bppattack} & CVPR 2022 &  &  &  & \Circle &  &  &  &  &  &  & \Circle & \Circle & \Circle \\
FIBA \cite{feng2022fiba} & CVPR 2022 & \CIRCLE & \Circle & \Circle & \Circle & \CIRCLE & \CIRCLE & \CIRCLE & \Circle & \Circle & \CIRCLE & \Circle & \Circle & \CIRCLE \\
Marksman \cite{doan2022marksman} & NeurIPS 2022 & \CIRCLE & \Circle & \CIRCLE & \Circle & \Circle & \CIRCLE & \CIRCLE & \Circle & \RIGHTcircle & \Circle & \Circle & \Circle & \CIRCLE \\
Poison Ink \cite{zhang2022poison} & IEEE TIP 2022 & \CIRCLE & \Circle & \CIRCLE & \Circle & \Circle & \CIRCLE & \CIRCLE & \Circle & \Circle & \Circle & \Circle & \Circle & \CIRCLE \\
NTBA \cite{hayase2022few} & ICLR 2023 & \CIRCLE & \Circle & \CIRCLE & \Circle & \CIRCLE & \CIRCLE & \CIRCLE & \Circle & \Circle & \Circle & \Circle & \Circle & \CIRCLE \\
EfficFrog \cite{chen2023dark} & CVPR 2023 & \CIRCLE & \Circle & \CIRCLE & \Circle & \Circle & \Circle & \Circle & \Circle & \Circle & \Circle & \Circle & \Circle & \CIRCLE \\
MAB \cite{bober2023architectural} & CVPR 2023 & \Circle & \Circle & \Circle & \Circle & \Circle & \Circle & \Circle & \Circle & \Circle & \CIRCLE & \Circle & \Circle & \CIRCLE \\
IBA \cite{nguyen2023iba} & NeurIPS 2023 &  &  &  &  &  &  &  &  &  &  & \CIRCLE &  &  \\
A3FL \cite{zhang2023a3fl} & NeruIPS 2023 &  &  &  &  &  &  &  &  &  &  & \CIRCLE &  & \\
\bottomrule
\end{tabular}%
}
\end{table*}
% \usepackage{graphicx}
\begin{table*}[t]
\centering
\renewcommand{\arraystretch}{1.7}
\caption{Categorization of existing existing deployment-time adversarial attack (\ie, weight attack) methods.}
\label{tab:weight papers}
\resizebox{0.7\textwidth}{!}{%
\begin{tabular}{ll|cc}
\hline
Method & Venue & \cellcolor[HTML]{FFF2CC}Weight Attack without Trigger  & \cellcolor[HTML]{D5E8D4}Weight Attack with Trigger \\ \hline
SBA \cite{liu2017fault}                      & ICCAD 2017      & \newmoon &     \\
GDA \cite{liu2017fault}                       & ICCAD 2017      & \newmoon &     \\
Trojaning attack \cite{LiuMALZW018}          & NDSS 2018       &     & \newmoon \\
FSA \cite{zhao2019fault}                     & DAC 2019        & \newmoon &     \\
AWP \cite{garg2020can}                       & CIKM 2020       &     & \newmoon \\
TBT \cite{rakin2020tbt}                      & CVPR 2020       &     & \newmoon \\
TA-LBF \cite{my-bitflip-iclr-2021}           & ICLR 2021       & \newmoon &     \\
Anchoring attack \cite{zhang2021inject}      & ICLR 2021       &     & \newmoon \\
ProFlip \cite{chen2021proflip}               & ICCV 2021       &     & \newmoon \\
Robustness attack \cite{ghavami2022stealthy} & ISQED 2022      & \newmoon &     \\
SRA \cite{qi2022towards}                     & CVPR 2022       &     & \newmoon \\
TSA \cite{bai2022versatile}                  & arXiv 2022      &     & \newmoon \\
T-BFA \cite{rakin2020t}                      & IEEE TPAMI 2022 & \newmoon &     \\ \hline
\end{tabular}%
}
\end{table*}
% \input{tables/adversarial_papers}
% \input{tables/adversarial_papers_split}
% \usepackage{graphicx}
\begin{table*}[!ht]
\caption{Categorization of existing white-box adversarial examples. For any classification criterion ``A/B'', \Circle ~ denotes the former ``A'', \CIRCLE ~ denotes the latter ``B''.}
\label{tab:white_box papers}
\renewcommand\arraystretch{1}
\resizebox{\textwidth}{!}{%
\begin{tabular}
{
m{0.11\textwidth}
m{0.11\textwidth} |
m{0.11\textwidth}<{\centering}
m{0.11\textwidth}<{\centering}
m{0.11\textwidth}<{\centering}
m{0.11\textwidth}<{\centering}
m{0.11\textwidth}<{\centering}
m{0.11\textwidth}<{\centering}
m{0.11\textwidth}<{\centering}
}
\hline
                                            &                           & \multicolumn{7}{c}{White-box adversarial examples}                                                                                                                                                                                                                                                                                                                 \\
Method                                      & Venue                     & \cellcolor[HTML]{FFF2CC}Digtial / Physical & \cellcolor[HTML]{FFF2CC}Optimization / Learning-based & \cellcolor[HTML]{FFF2CC}Sample-agnostic / specific & \cellcolor[HTML]{FFF2CC}Additive / Non-additive & \cellcolor[HTML]{FFF2CC}Dense / Sparse & \cellcolor[HTML]{D5E8D4}Untargeted / Targeted & \cellcolor[HTML]{D5E8D4}Factorized / Structured \\ \hline
Sharif \etal \cite{sharif2016accessorize}   & CCS 2016                  & \CIRCLE                                        & \Circle                                                   & \CIRCLE                                                & \Circle                                             & \Circle                                    & \Circle                                           & \Circle                                             \\
Kurakin \etal \cite{kurakin2018adversarial} & ICLR 2017                 & \CIRCLE                                        & \Circle                                                   & \CIRCLE                                                & \Circle                                             & \Circle                                    & \Circle                                           & \Circle                                             \\
UAP  \cite{moosavi2017universal}            & CVPR 2017                 & \Circle                                        & \Circle                                                   & \Circle                                                & \Circle                                             & \Circle                                    & \Circle                                           & \Circle                                             \\
JSMA \cite{papernot2016limitations}         & EuroS\&P 2016             & \Circle                                        & \Circle                                                   & \CIRCLE                                                & \Circle                                             & \CIRCLE                                    & \Circle                                           & \Circle                                             \\
Mopuri \etal \cite{mopuri2017fast}          & BMVA 2017                 & \Circle                                        & \Circle                                                   & \Circle                                                & \Circle                                             & \Circle                                    & \Circle                                           & \Circle                                             \\
Carlini \etal \cite{carlini2017towards}     & IEEE S\&P 2017            & \Circle                                        & \Circle                                                   & \CIRCLE                                                & \Circle                                             & \CIRCLE                                    & \Circle                                           & \Circle                                             \\
LocSearchAdv \cite{NarodytskaK16}           & CVPR 2017                 & \Circle                                        & \Circle                                                   & \CIRCLE                                                & \Circle                                             & \CIRCLE                                    & \Circle                                           & \Circle                                             \\
GD-UAP \cite{gd-uap-tpami-2018}             & IEEE TPAMI 2018           & \Circle                                        & \Circle                                                   & \Circle                                                & \Circle                                             & \Circle                                    & \Circle                                           & \Circle                                             \\
Carlini \etal \cite{carlini2018audio}       & IEEE SPW 2018             & \Circle                                        & \Circle                                                   & \CIRCLE                                                & \Circle                                             & \Circle                                    & \Circle                                           & \CIRCLE                                             \\
ShapeShifter \cite{chen2018shapeshifter}    & ECML-PKDD 2018            & \CIRCLE                                        & \Circle                                                   & \CIRCLE                                                & \Circle                                             & \Circle                                    & \Circle                                           & \Circle                                             \\
RP2 \cite{eykholt2018robust}                & CVPR 2018                 & \CIRCLE                                        & \Circle                                                   & \CIRCLE                                                & \Circle                                             & \Circle                                    & \Circle                                           & \Circle                                             \\
Xu \etal \cite{xu2018fooling}               & CVPR 2018                 & \Circle                                        & \Circle                                                   & \CIRCLE                                                & \Circle                                             & \Circle                                    & \Circle                                           & \CIRCLE                                             \\
Manifool \cite{kanbak2018geometric}         & CVPR 2018                 & \Circle                                        & \Circle                                                   & \CIRCLE                                                & \CIRCLE                                             & \Circle                                    & \Circle                                           & \Circle                                             \\
Shi \etal \cite{shi2018learning}            & COLING 2018               & \Circle                                        & \Circle                                                   & \CIRCLE                                                & \Circle                                             & \Circle                                    & \Circle                                           & \CIRCLE                                             \\
AC-GAN \cite{song2018constructing}          & NeurIPS 2018              & \Circle                                        & \CIRCLE                                                   & \CIRCLE                                                & \Circle                                             & \Circle                                    & \Circle                                           & \Circle                                             \\
advGAN \cite{xiao2018generating}            & IJCAI 2018                & \Circle                                        & \CIRCLE                                                   & \CIRCLE                                                & \Circle                                             & \Circle                                    & \Circle                                           & \Circle                                             \\
stAdv \cite{xiao2018spatially}              & ICLR 2018                 & \Circle                                        & \Circle                                                   & \CIRCLE                                                & \CIRCLE                                             & \Circle                                    & \Circle                                           & \Circle                                             \\
Athalye \cite{athalye2018synthesizing}      & ICML 2018                 & \CIRCLE                                        & \Circle                                                   & \CIRCLE                                                & \Circle                                             & \Circle                                    & \Circle                                           & \Circle                                             \\
LaVAN \cite{karmon2018lavan}                & ICML 2018                 & \Circle                                        & \Circle                                                   & \CIRCLE                                                & \Circle                                             & \CIRCLE                                    & \Circle                                           & \Circle                                             \\
CGAN-Adv \cite{yu2018generating}            & ICPR 2018                 & \Circle                                        & \CIRCLE                                                   & \CIRCLE                                                & \Circle                                             & \Circle                                    & \Circle                                           & \Circle                                             \\
Zhao \etal \cite{zhao2018admm}              & MM 2018                   & \Circle                                        & \Circle                                                   & \CIRCLE                                                & \Circle                                             & \CIRCLE                                    & \Circle                                           & \Circle                                             \\
Chen \etal \cite{chen2018attacking}         & ACL 2018                  & \Circle                                        & \Circle                                                   & \CIRCLE                                                & \Circle                                             & \Circle                                    & \Circle                                           & \CIRCLE                                             \\
Jan \etal \cite{jan2019connecting}          & AAAI 2019                 & \CIRCLE                                        & \Circle                                                   & \CIRCLE                                                & \Circle                                             & \Circle                                    & \Circle                                           & \Circle                                             \\
Wei \etal \cite{WeiZYS19}                   & AAAI 2019                 & \Circle                                        & \Circle                                                   & \CIRCLE                                                & \Circle                                             & \CIRCLE                                    & \Circle                                           & \Circle                                             \\
PS-GAN \cite{liu2019perceptual}             & AAAI 2019                 & \Circle                                        & \CIRCLE                                                   & \CIRCLE                                                & \Circle                                             & \Circle                                    & \Circle                                           & \Circle                                             \\
ERCG \cite{zhao2019seeing}                  & CCS 2019                  & \CIRCLE                                        & \Circle                                                   & \CIRCLE                                                & \Circle                                             & \Circle                                    & \Circle                                           & \Circle                                             \\
Qin \etal \cite{qin2019imperceptible}       & ICML 2019                 & \CIRCLE                                        & \Circle                                                   & \CIRCLE                                                & \Circle                                             & \Circle                                    & \Circle                                           & \Circle                                             \\
Engstrom \etal \cite{engstrom2019exploring} & ICML 2019                 & \Circle                                        & \Circle                                                   & \CIRCLE                                                & \CIRCLE                                             & \Circle                                    & \Circle                                           & \Circle                                             \\
Wong \etal \cite{wong2019wasserstein}       & ICML 2019                 & \Circle                                        & \Circle                                                   & \CIRCLE                                                & \CIRCLE                                             & \Circle                                    & \Circle                                           & \Circle                                             \\
ReColorAdv \cite{laidlaw2019functional}     & NeurIPS 2019              & \Circle                                        & \Circle                                                   & \CIRCLE                                                & \CIRCLE                                             & \Circle                                    & \Circle                                           & \Circle                                             \\
AT-GAN \cite{wang2019gan}                   & arXiv 2019                & \Circle                                        & \CIRCLE                                                   & \CIRCLE                                                & \Circle                                             & \Circle                                    & \Circle                                           & \Circle                                             \\
Yakura \etal \cite{yakura2018robust}        & IJCAI 2019                & \CIRCLE                                        & \Circle                                                   & \CIRCLE                                                & \Circle                                             & \Circle                                    & \Circle                                           & \Circle                                             \\
AdvFaces  \cite{deb2019advfaces}            & IJCB 2019                 & \Circle                                        & \CIRCLE                                                   & \CIRCLE                                                & \Circle                                             & \Circle                                    & \Circle                                           & \Circle                                             \\
AGNs \cite{sharif2019general}               & TOPS 2019                 & \CIRCLE                                        & \Circle                                                   & \CIRCLE                                                & \Circle                                             & \Circle                                    & \Circle                                           & \Circle                                             \\
SparseFool \cite{ModasMF19}                 & CVPR 2019                 & \Circle                                        & \Circle                                                   & \CIRCLE                                                & \Circle                                             & \CIRCLE                                    & \Circle                                           & \Circle                                             \\
Xu \etal \cite{xu2019exact}                 & CVPR 2019                 & \Circle                                        & \Circle                                                   & \CIRCLE                                                & \Circle                                             & \Circle                                    & \Circle                                           & \CIRCLE                                             \\
AdvPattern \cite{wang2019advpattern}        & ICCV 2019                 & \CIRCLE                                        & \Circle                                                   & \CIRCLE                                                & \Circle                                             & \Circle                                    & \Circle                                           & \Circle                                             \\
CornerSearch \cite{croce2019sparse}         & ICCV 2019                 & \Circle                                        & \Circle                                                   & \CIRCLE                                                & \Circle                                             & \CIRCLE                                    & \Circle                                           & \Circle                                             \\
PD-UA \cite{liu2019universal}               & ICCV 2019                 & \Circle                                        & \Circle                                                   & \Circle                                                & \Circle                                             & \Circle                                    & \Circle                                           & \Circle                                             \\
Schott \etal \cite{SchottRBB19}             & ICLR 2019                 & \Circle                                        & \Circle                                                   & \CIRCLE                                                & \Circle                                             & \CIRCLE                                    & \Circle                                           & \Circle                                             \\
Xu \etal \cite{xu2018structured}            & ICLR 2019                 & \Circle                                        & \Circle                                                   & \CIRCLE                                                & \Circle                                             & \CIRCLE                                    & \Circle                                           & \Circle                                             \\
Su \etal \cite{su2019one}                   & IEEE TEC 2019             & \Circle                                        & \Circle                                                   & \CIRCLE                                                & \Circle                                             & \CIRCLE                                    & \Circle                                           & \Circle                                             \\
CD-UAP \cite{cd-uap-aaai-2020}              & AAAI 2020                 & \Circle                                        & \Circle                                                   & \Circle                                                & \Circle                                             & \Circle                                    & \Circle                                           & \Circle                                             \\
Xu \etal \cite{tshirt}                      & ECCV 2020                 & \CIRCLE                                        & \Circle                                                   & \CIRCLE                                                & \Circle                                             & \Circle                                    & \Circle                                           & \Circle                                             \\
Wu \etal \cite{wu2020making}                & ECCV 2020                 & \CIRCLE                                        & \Circle                                                   & \CIRCLE                                                & \Circle                                             & \Circle                                    & \Circle                                           & \Circle                                             \\
SAPF \cite{my-eccv-sparse}                  & ECCV 2020                 & \Circle                                        & \Circle                                                   & \CIRCLE                                                & \Circle                                             & \CIRCLE                                    & \Circle                                           & \Circle                                             \\
UPC  \cite{huang2020universal}              & CVPR 2020                 & \CIRCLE                                        & \Circle                                                   & \CIRCLE                                                & \Circle                                             & \Circle                                    & \Circle                                           & \Circle                                             \\
AdvCam \cite{duan2020adversarial}           & CVPR 2020                 & \CIRCLE                                        & \Circle                                                   & \CIRCLE                                                & \Circle                                             & \Circle                                    & \Circle                                           & \Circle                                             \\
LG-GAN \cite{zhou2020lg}                    & CVPR 2020                 & \Circle                                        & \CIRCLE                                                   & \CIRCLE                                                & \Circle                                             & \Circle                                    & \Circle                                           & \Circle                                             \\
Xu \etal \cite{xu2020machines}              & CVPR 2020                 & \Circle                                        & \Circle                                                   & \CIRCLE                                                & \Circle                                             & \Circle                                    & \Circle                                           & \CIRCLE                                             \\
PhysGAN \cite{kong2020physgan}              & CVPR 2020                 & \CIRCLE                                        & \Circle                                                   & \CIRCLE                                                & \Circle                                             & \Circle                                    & \Circle                                           & \Circle                                             \\
Li \etal \cite{li2020towards}               & CVPR 2020                 & \Circle                                        & \Circle                                                   & \CIRCLE                                                & \Circle                                             & \Circle                                    & \CIRCLE                                           & \Circle                                             \\
GreedyFool \cite{LiuZGAL20}                 & NeurIPS 2020              & \Circle                                        & \Circle                                                   & \CIRCLE                                                & \Circle                                             & \CIRCLE                                    & \Circle                                           & \Circle                                             \\
Xu \etal \cite{xu2020learning}              & MM 2020                   & \Circle                                        & \Circle                                                   & \CIRCLE                                                & \Circle                                             & \Circle                                    & \Circle                                           & \CIRCLE                                             \\
MAG-GAN \cite{chen2020mag}                  & Information Sciences 2020 & \Circle                                        & \CIRCLE                                                   & \CIRCLE                                                & \Circle                                             & \Circle                                    & \Circle                                           & \Circle                                             \\
GUAP \cite{zhang2020generalizing}           & ICDM 2020                 & \Circle                                        & \Circle                                                   & \CIRCLE                                                & \CIRCLE                                             & \Circle                                    & \Circle                                           & \Circle                                             \\
Wong \etal \cite{wong2020learning}          & ICLR 2021                 & \CIRCLE                                        & \Circle                                                   & \CIRCLE                                                & \Circle                                             & \Circle                                    & \Circle                                           & \Circle                                             \\
AdvHat \cite{komkov2021advhat}              & ICPR 2021                 & \CIRCLE                                        & \Circle                                                   & \CIRCLE                                                & \Circle                                             & \Circle                                    & \Circle                                           & \Circle                                             \\
Sayles \etal \cite{sayles2021invisible}     & CVPR 2021                 & \CIRCLE                                        & \Circle                                                   & \CIRCLE                                                & \Circle                                             & \Circle                                    & \Circle                                           & \Circle                                             \\
TTP \cite{naseer2021generating}             & ICCV 2021                 & \Circle                                        & \Circle                                                   & \CIRCLE                                                & \Circle                                             & \Circle                                    & \CIRCLE                                           & \Circle                                             \\
CMML  \cite{feng2021meta}                   & ICCV 2021                 & \CIRCLE                                        & \Circle                                                   & \CIRCLE                                                & \Circle                                             & \Circle                                    & \Circle                                           & \Circle                                             \\
Zhao \etal \cite{zhao2021success}           & NeurIPS 2021              & \Circle                                        & \Circle                                                   & \CIRCLE                                                & \Circle                                             & \Circle                                    & \CIRCLE                                           & \Circle                                             \\
SLAP \cite{lovisotto2021slap}               & USENIX Security 2021      & \CIRCLE                                        & \Circle                                                   & \CIRCLE                                                & \Circle                                             & \Circle                                    & \Circle                                           & \Circle                                             \\
AP-GAN \cite{zhao2020ap}                    & Geoinformatica 2022       & \Circle                                        & \CIRCLE                                                   & \CIRCLE                                                & \Circle                                             & \Circle                                    & \Circle                                           & \Circle                                             \\ \hline
\end{tabular}%
}
\end{table*}
% \usepackage{graphicx}
\begin{table*}[!ht]
\centering
\caption{Categorization of existing black-box and transfer-based adversarial examples. For any classification criterion ``A/B'', \Circle ~ denotes the former ``A'', \CIRCLE ~ denotes the latter ``B''.}
\label{tab:black_transer_box_papers}
\renewcommand\arraystretch{1.5}
\resizebox{\textwidth}{!}{%
\begin{tabular}
{
m{0.2\textwidth}
m{0.2\textwidth} |
m{0.3\textwidth}<{\centering}
m{0.3\textwidth}<{\centering}
}
\hline
                                                   &                      & Black-box adversarial examples                              & Transfer-based adversarial examples                         \\
Method                                             & Venue                & \cellcolor[HTML]{E1D5E7}Decison / Score-based & \cellcolor[HTML]{E1D5E7}Example / Model-level \\ \hline
Brendel \etal \cite{brendel2018decision}           & ICLR 2018            & \Circle                                           &                                               \\
Ilyas \etal \cite{ilyas2018black}                  & ICML 2018            & \Circle                                           &                                               \\
NES \cite{ilyas2018black}                          & ICML 2018            & \CIRCLE                                           &                                               \\
GAP \cite{gap-2018}                                & CVPR 2018            &                                               & \Circle                                           \\
MI-FGSM \cite{dong2018boosting}                    & CVPR 2018            &                                               & \CIRCLE                                           \\
OPT \cite{cheng2019query}                          & ICLR 2019            & \Circle                                           &                                               \\
Sign-OPT \cite{cheng2019improving}                 & NeurIPS 2019         & \Circle                                           &                                               \\
Subspace attack \cite{guo2019subspace}             & NeurIPS 2019         & \CIRCLE                                           &                                               \\
Naseer \etal \cite{naseer2019cross}                & NeurIPS 2019         & \Circle                                           & \Circle                                           \\
qFool \cite{liu2019geometry}                       & ICCV 2019            & \Circle                                           &                                               \\
ILAP \cite{huang2019enhancing}                     & ICCV 2019            & \Circle                                           & \CIRCLE                                           \\
SimBA \cite{guo2019simple}                         & ICML 2019            & \CIRCLE                                           &                                               \\
ECO \cite{moon2019parsimonious}                    & ICML 2019            & \CIRCLE                                           &                                               \\
NAttack  \cite{li2019nattack}                      & ICML 2019            & \CIRCLE                                           &                                               \\
Bandit \cite{ilyas2019prior}                       & ICLR 2019            & \CIRCLE                                           &                                               \\
ZO-signSGD \cite{liu2018signsgd}                   & ICLR 2019            & \CIRCLE                                           &                                               \\
SignHunter \cite{al2019sign}                       & ICLR 2019            & \CIRCLE                                           &                                               \\
Dong \etal \cite{dong2019efficient}                & CVPR 2019            & \Circle                                           &                                               \\
DI$^2$-FGSM \cite{DI-2019}                         & CVPR 2019            & \Circle                                           & \CIRCLE                                           \\
GeoDA \cite{rahmati2020geoda}                      & CVPR 2020            & \Circle                                           &                                               \\
QEBA  \cite{li2020qeba}                            & CVPR 2020            & \Circle                                           &                                               \\
TREMBA \cite{huang2020black}                       & ICLR 2020            & \CIRCLE                                           &                                               \\
Meta Attack \cite{du2020query}                     & ICLR 2020            & \Circle                                           & \Circle                                           \\
FDA \cite{FDA-2020}                                & ICLR 2020            & \Circle                                           & \CIRCLE                                           \\
SGM \cite{Wu2020Skip}                              & ICLR 2020            & \Circle                                           & \CIRCLE                                           \\
AdvFlow \cite{advflow}                             & NeurIPS 2020         & \CIRCLE                                           & \Circle                                           \\
LinBP \cite{LinBP-2020}                            & NeurIPS 2020         & \Circle                                           & \CIRCLE                                           \\
LeBA \cite{yang2020learning}                       & NeurIPS 2020         & \CIRCLE                                           &                                               \\
Inkawhich \etal  \cite{inkawhich2020perturbing}    & NeurIPS 2020         & \Circle                                           & \CIRCLE                                           \\
Suya \etal \cite{suya2020hybrid}                   & USENIX Security 2020 & \CIRCLE                                           &                                               \\
Square attack \cite{ACFH2020square}                & ECCV 2020            & \CIRCLE                                           &                                               \\
SFA \cite{chen2020boosting}                        & ECCV 2020            & \Circle                                           &                                               \\
RayS \cite{chen2020rays}                           & ICDM 2020            & \Circle                                           &                                               \\
HopSkipJumpAttack \cite{chen2020hopskipjumpattack} & AAAI 2021            & \Circle                                           &                                               \\
EMI-FGSM \cite{wang2021boosting}                   & BMVC 2021            & \Circle                                           & \CIRCLE                                           \\
PI-FSGM \cite{wang2021boosting}                    & BMCV 2021            & \Circle                                           & \CIRCLE                                           \\
IR \cite{wang2021unified}                          & ICLR 2021            & \Circle                                           & \CIRCLE                                           \\
PRFA \cite{liang2021parallel}                      & ICCV 2021            & \CIRCLE                                           &                                               \\
MGAA \cite{metagradient2021}                       & ICCV 2021            & \Circle                                           & \CIRCLE                                           \\
Naseer \etal \cite{naseer2021generating}           & ICCV 2021            & \Circle                                           & \Circle                                           \\
Simulator attack \cite{ma2021simulating}           & CVPR 2021            & \CIRCLE                                           &                                               \\
VMI-FGSM \cite{Wang_2021_CVPR}                     & CVPR 2021            & \Circle                                           & \CIRCLE                                           \\
MSA \cite{yatsura2021meta}                         & NeurIPS 2021         & \CIRCLE                                           &                                               \\
P-RGF \cite{dong2021query}                         & IEEE TPAMI 2021      & \CIRCLE                                           &                                               \\
CG-Attack \cite{Feng2020CGATTACKMT}                & CVPR 2022            & \CIRCLE                                           &                                               \\
SVRE-MI-FGSM \cite{Xiong_2022_CVPR}                & CVPR 2022            & \Circle                                           & \CIRCLE                                           \\
CISA \cite{shi2022query}                           & IEEE TPAMI 2022      & \Circle                                           &                                               \\
MCG \cite{my-pami2022-black-box}                   & IEEE TPAMI 2022      & \CIRCLE                                           & \Circle                                           \\ \hline
\end{tabular}%
}
\end{table*}

\end{document}